\documentclass[a4paper]{IEEEtran}
\IEEEoverridecommandlockouts

\usepackage{graphicx}
\usepackage{multirow}
\usepackage{amsmath,amssymb,amsfonts}
\usepackage{amsthm}
\usepackage{mathrsfs}
\usepackage{xcolor}
\usepackage{textcomp}
\usepackage{manyfoot}

\usepackage{algorithm}
\usepackage{algorithmicx}
\usepackage{algpseudocode}
\usepackage{listings}

\usepackage{booktabs}
\usepackage{cite}
\usepackage{color}
\usepackage{soul}
\usepackage{hyperref} 
\usepackage{caption} 
\usepackage{subcaption} 
\usepackage{rotating} 
\usepackage{multirow} 
\usepackage{nicefrac} 
\usepackage{tikz} 
\usepackage{mathabx} 
\usepackage[normalem]{ulem} 
\usepackage{siunitx}
\usepackage{comment}
\usepackage{orcidlink}
\usepackage{adjustbox}  
\usepackage{cleveref}

\usetikzlibrary{shapes.geometric, arrows} 
\tikzstyle{boox} = [rectangle,rounded corners,minimum width=1.6cm,minimum height=0.55cm,text centered,draw=black,fill=black!10]

\begin{document}

\title{
    SYNOSIS: Image synthesis pipeline for machine vision in metal surface inspection
}

\author{
    \IEEEauthorblockN{\orcidlink{0000-0003-1858-5978} Juraj Fulir*},
    \IEEEauthorblockA{
        \textit{Fraunhofer ITWM},
        Kaiserslautern, Germany \\
    }
    \and
    \IEEEauthorblockN{Natascha Jeziorski},
    \IEEEauthorblockA{
        \textit{RPTU Kaiserslautern-Landau},
        Kaiserslautern, Germany \\
    }
    \and
    \IEEEauthorblockN{\orcidlink{0000-0002-0521-6615} Lovro Bosnar},
    \IEEEauthorblockA{
        \textit{Fraunhofer ITWM},
        Kaiserslautern, Germany \\
    }
    \and[\hfill\mbox{}\par\mbox{}\hfill]
    \IEEEauthorblockN{\orcidlink{0000-0001-5626-963X} Hans Hagen},
    \IEEEauthorblockA{
        \textit{RPTU Kaiserslautern-Landau},
        Kaiserslautern, Germany \\
    }
    \and
    \IEEEauthorblockN{\orcidlink{0000-0002-8030-069X} Claudia Redenbach},
    \IEEEauthorblockA{
        \textit{RPTU Kaiserslautern-Landau},
        Kaiserslautern, Germany \\
    }
    \and
    \IEEEauthorblockN{\orcidlink{0000-0002-2102-2618} Petra Gospodneti{ć}},
    \IEEEauthorblockA{
        \textit{Fraunhofer ITWM},
        Kaiserslautern, Germany \\
    }
    \and[\hfill\mbox{}\par\mbox{}\hfill]
    \IEEEauthorblockN{Tobias Herrfurth},
    \IEEEauthorblockA{
        \textit{Fraunhofer IOF},
        Jena, Germany \\
    }
    \and
    \IEEEauthorblockN{Marcus Trost},
    \IEEEauthorblockA{
        \textit{Fraunhofer IOF},
        Jena, Germany \\
    }
    \and
    \IEEEauthorblockN{Thomas Gischkat},
    \IEEEauthorblockA{
        \textit{Fraunhofer IOF},
        Jena, Germany \\
    }
}

\maketitle

\begin{abstract}
    The use of machine learning (ML) methods for development of robust and flexible visual inspection system has shown promising.
    However their performance is highly dependent on the amount and diversity of training data.
    This is often restricted not only due to costs but also due to a wide variety of defects and product surfaces which occur with varying frequency.
    As such, one can not guarantee that the acquired dataset contains enough defect and product surface occurrences which are needed to develop a robust model.
    Using parametric synthetic dataset generation, it is possible to avoid these issues.
    In this work, we introduce a complete pipeline which describes in detail how to approach image synthesis for surface inspection - from first acquisition, to texture and defect modeling, data generation, comparison to real data and finally use of the synthetic data to train a defect segmentation model.
    The pipeline is in detail evaluated for milled and sandblasted aluminum surfaces.
    In addition to providing an in-depth view into each step, discussion of chosen methods, and presentation of ML results, we provide a comprehensive dual dataset containing both real and synthetic images.
\end{abstract}


\begin{figure*}
    \centering
    \includegraphics[width=0.99\linewidth]{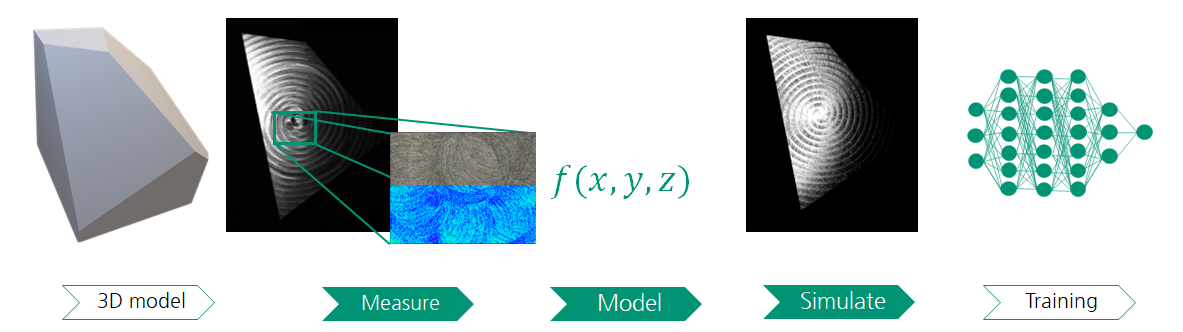}
    \caption{SYNOSIS synthesis pipeline. Based on a \textbf{3D model} of an object, spatial properties of the surface texture are \textbf{measured}. Surface topography and manufacturing parameters are used to develop a \textbf{mathematical model} capable of reproducing the texture as a normal map. Multiple realizations of the texture are generated by varying the parameters and are applied onto the 3D model of the object which has been altered to include surface defects of varying types and sizes. During the \textbf{simulation} step, the final image is computed based on the interaction between the acquisition environment (light, camera), object geometry and texture normal map. This process is repeated an arbitrary amount of times. Parameters defining texture and surface defects are varied to generate a \textbf{training dataset} which is sufficient in terms of both image quantity and content variance.}
    \label{fig:overview_pipeline}
\end{figure*}

\section{Introduction}
\label{sec:introduction}
Machine vision is used to automate repetitive processes which rely on visual information. As such, it has many uses in industrial applications.
Processes such as visual quality inspection would otherwise be performed either by humans or might not even be performed at all in cases such as extreme working environments (temperature, noise, chemicals, etc.).
Albeit being very flexible in terms of variety of possible applications, every machine vision solution on its own is highly specialized, integrates expert knowledge and cannot be easily adapted if inspection requirements or production environment change.

Automated visual inspection systems (further referred to as inspection systems) are currently developed in a manner which requires the rigidness to ensure reliable performance \cite{Gospodnetic2020}.
This further reduces the flexibility and customization capabilities of production lines, which are ever more required.
As a mean to reduce the rigidity, the integrators of inspection systems are turning towards machine learning (ML) techniques for machine vision.
While the ML solutions have showed promising, their reliability and flexibility over longer periods of time is yet to be proven, making the industry skeptic towards their integration into production lines.
Additionally, machine learning requires a significant amount of training data, which has shown to be challenging for industrial inspection systems.
The inspection tasks are typically very specific and there are little to no publicly available datasets \cite{Wagenstetter2024}.
Therefore, a dataset must be created for each new system.
This is not only costly and time consuming, but also poses a significant challenge to generate a well balanced dataset.

The use of synthetic data for ML is rapidly gaining popularity \cite{nikolenko2021synthetic}.
When it comes to its use in inspection systems, isolated studies have been performed providing evidence of its benefit.
However, to date there is a lack of commonly agreed upon best practice or a more comprehensive evaluation of different steps required to generate a photorealistic synthetic dataset for visual inspection.
Furthermore, inspection datasets containing both real and synthetic images of the same task are scarce.
Such datasets are needed to help establish benchmark performance and support further research and result comparison on the use of synthetic data for visual inspection.

In this work we address the above-mentioned gap by presenting the \textbf{SYNOSIS\footnote{Synthetic, optically realistic images for ML-based inspection systems} pipeline}. As illustrated in \cref{fig:overview_pipeline}, the pipeline includes 1) surface measurement and parameterization, 2) generation of physically correct synthetic images through parameterization of surface, defects and acquisition environment, 3) evaluation of synthetic image quality, and 4) their use for surface inspection ML.
For the purpose of this work, the chosen inspection use-case focused on milled and sandblasted metal surfaces without coating.
However, the pipeline is developed to be applicable for a wide variety of materials and defects, and, for the first time, introduces \textbf{decomposition of different geometry scales} in inspection context.
Finally, we \textbf{publish a dual dataset\footnote{Dataset is publicly available at: \url{http://dx.doi.org/10.24406/fordatis/370}}} for defect recognition containing real and synthetic image pairs acquired and simulated for corresponding parameters.

\subsection{Paper Structure}
The content is grouped into 8 sections and contains bolded keywords to make the reading easier.
The generation process needed to obtain synthetic images for surface inspection is covered in \cref{sec:method:synth_datagen}. There we provide an overview of the related work, introduce inspection relevant decomposition of scales, and discuss the most important elements of the generation process and their relation to different scales.
Sections \ref{sec:method:test_body_design} and \ref{sec:method:material_measurement} introduce methods chosen to produce test object surfaces in a controlled manner and measure surface topography.
Sections \ref{sec:texture_modeling} and \ref{sec:defect_modeling} focus on texture and defect modeling as the most relevant aspects which need to be controlled when creating synthetic data for surface inspection.
Section \ref{sec:texture_modeling} discusses different approaches which can be taken to represent surface texture and their benefits or shortcomings. Further, the section introduces stochastic geometry modeling approaches used to the specific surfaces found in our specific use-case.
Section \ref{sec:defect_modeling} discusses different approaches to defect generation in the related work, and provides an overview of the defect modeling workflow used for our specific use-case.
Section \ref{sec:dual_dataset} introduces how was both real and synthetic data obtained. For synthetic data it explains how was the synthetic image generation process (introduced in \cref{sec:method:synth_datagen}) used and with which parameters to create the synthetic dataset.
Finally, \cref{sec:pipeline_evaluation} provides a discussion of different means which can be used to evaluate the quality of the created synthetic dataset and their results when applied to our use-case.

\begin{figure*}
    \centering
     \includegraphics[width=0.3\textwidth]{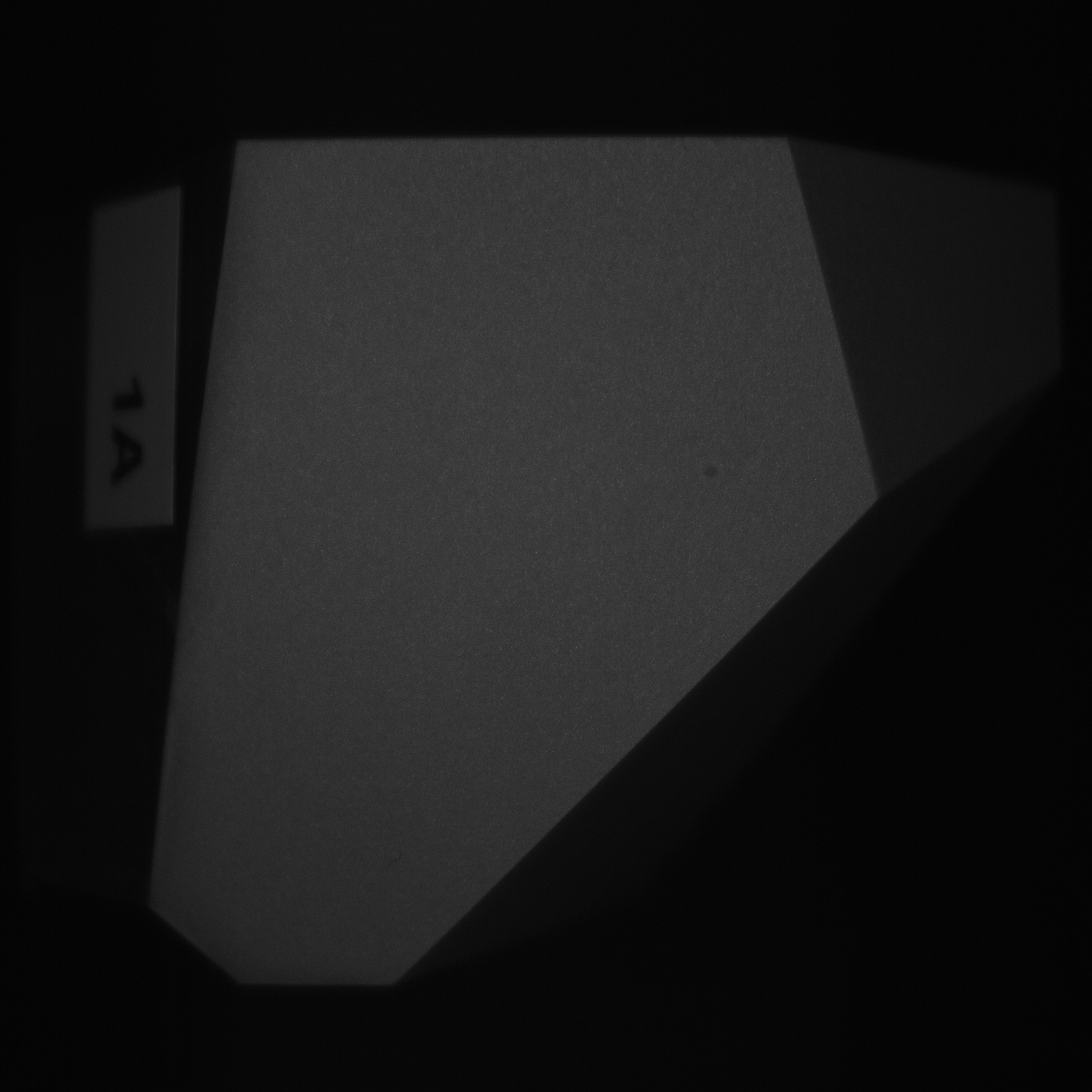}
     \includegraphics[width=0.3\textwidth]{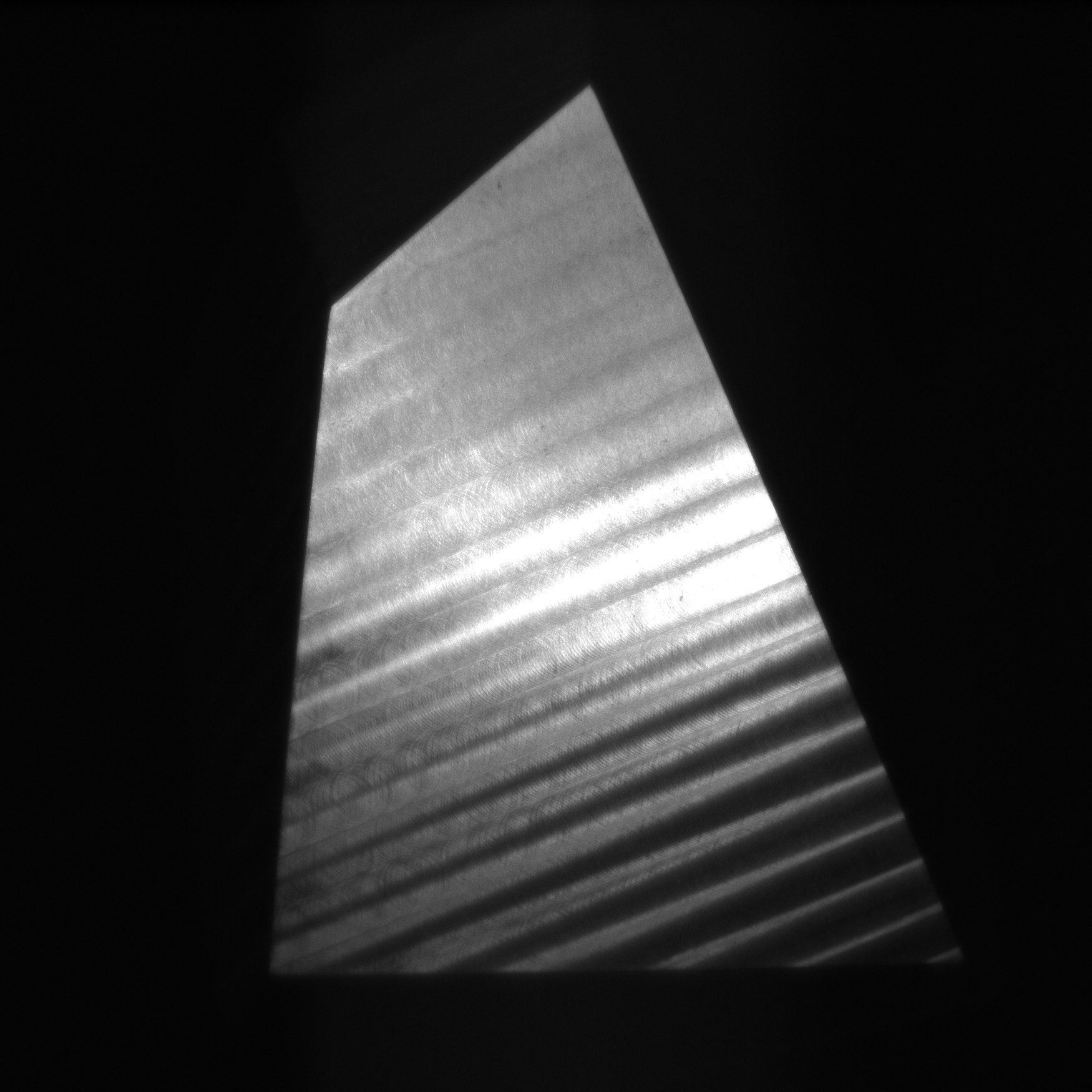}
     \includegraphics[width=0.3\textwidth]{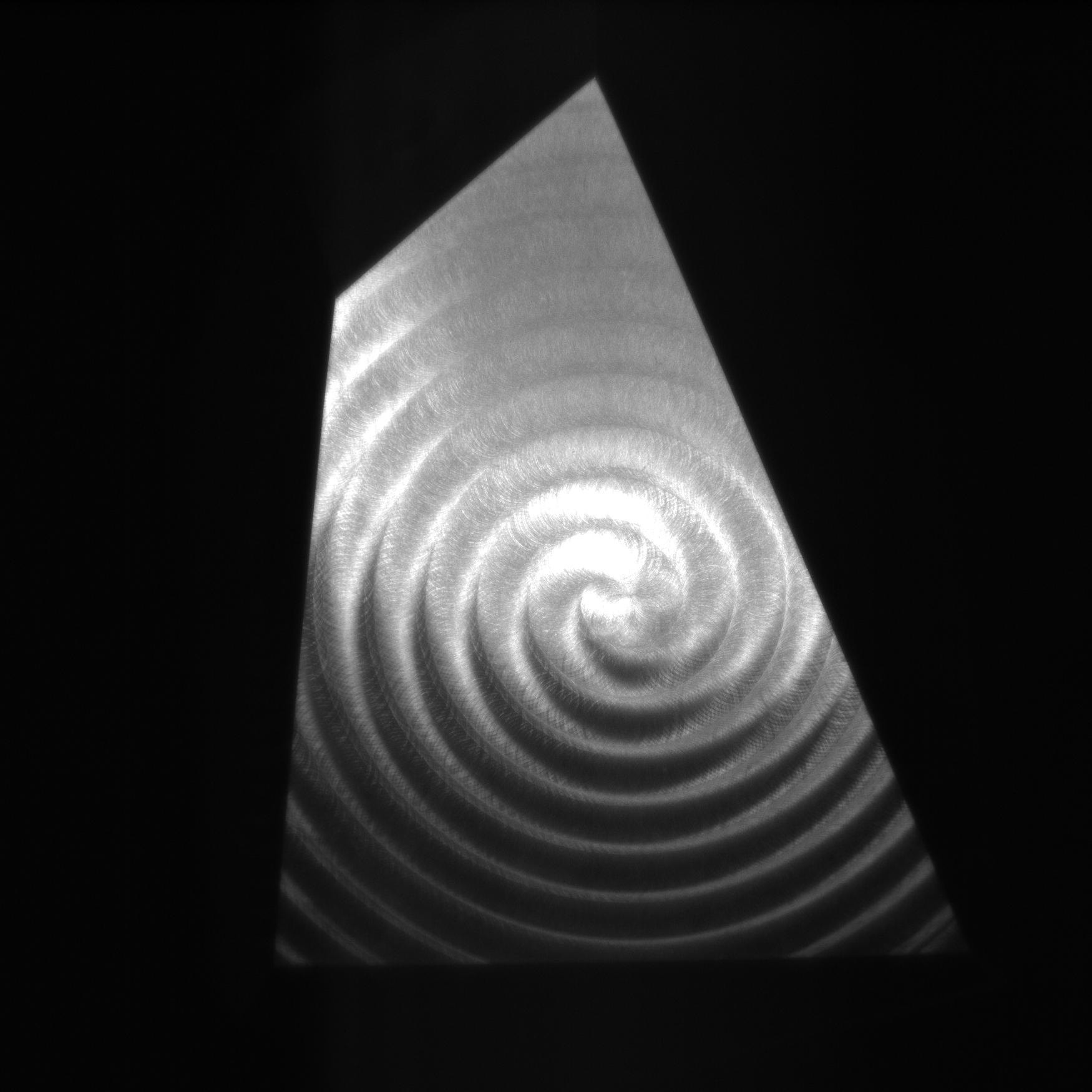}
    \caption{Examples of the three textures considered in this paper: surface finishing by sandblasting, parallel milling, and spiral milling.}
    \label{fig:image_synthesis:TexturesConsidered}
\end{figure*}

\section{Synthetic Dataset Generation}
\label{sec:method:synth_datagen}
\subsection{Related work}
\label{sec:image_synthesis:related_work}
The goal of \textbf{synthetic image data generation} is to generate image datasets where image characteristics are ideally the same as those of images acquired in reality.
The characteristics include both appearance and content distribution.
When there is a discrepancy between the synthetic and real images in terms of their characteristics, it is referred to as \textit{domain gap}.

Early, straightforward, approach to extending training datasets was to add defect textures to images of defect-free surfaces \cite{Haselmann2019defect_segmentation}.
The defects are 2D image patches which cannot correctly represent light interaction and have inconsistent appearance with respect to the rest of the surface.
Due to the nature of the approach it is very difficult to enhance it in a way which would minimize the resulting domain gap.
Therefore, further synthetic data efforts were aimed at \textbf{generative} methods, based on predominantly deep learning, and \textbf{rule-based} approaches, using computer graphics simulation based on a well-defined set of rules which describe the imaging context \cite{Schladitz2023}.

The high degree of complexity when generating photorealistic surface features caused many of the authors to adopt the \textbf{generative approach}, focusing on defect generation. Jain et al. \cite{Jain2020Steel} compare the performance of three different GAN architectures for generating defects in hot-rolled steel strips.
Defect-GAN \cite{Zhang2021DefectGAN} was introduced for automated surface defect synthesis by mimicking the defacement and restoration process and capturing stochastic variation of defects.
Defect-Transfer GAN \cite{Wang2022DTGAN} is trained over multiple products with similar defects in order to enable introduction of semantically new information and make up for the limited availability of all possible object attributes in the real dataset.
Wei et al. \cite{Wei2023GAN} generate defect patches and blend them into a clean image, thus being able to obtain approximate segmentation masks.
These approaches are more suitable for generating large defects which visibly differ from the surrounding surface and do not suffer if image is resized to lower resolution for inference performance reasons.
As such they can be expected to struggle when confronted with surface containing texture pattern whose appearance has prominent local variation, as is the case for milled surfaces.
Schmedemann \cite{schmedemann2023adapting} explored a hybrid approach by applying style-transfer on images generated by a rule-based approach to obtain more control and reduce the \textit{domain gap}.
It has shown that this approach can lead to a loss of features making it not reliable without a human operator review of the generated images.

Generative deep learning approaches may result in images which look realistic but also incorporate errors on a level which is beyond our comprehension, further causing misalignment when that data is used for training.
Also, the generative methods do not know anything about the context or what has caused an image to appear in a particular manner. Hence, it is difficult to control the dataset content and the distribution of features within.
That said, they cannot introduce edge-case scenarios or guarantee correct data generation.
This problem can be partially circumvented only by retraining on new data containing the desired features.

\textbf{Rule-based image synthesis} relies on computer graphics for modeling and rendering \textbf{virtual environments} which have recently gained popularity beyond the entertainment industry (e.g., movies or gaming).
It represents a controllable, versatile and reliable approach for generating arbitrary amounts of data with customized features and variations.
As such, it has proven useful across various machine vision tasks as 
discussed by Dahmen et al. \cite{Dahmen2019digital_reality_survey}.
Synthetic data are currently predominantly used in the human-oriented domains, autonomous driving and robotics, see \cite{Tsirikoglou2020Survey}.

Human-oriented synthetic data is used in tasks which include body tracking and pose estimation \cite{varol2017learning,park2015articulated,fabbri2018learning} or face recognition \cite{kortylewski2018training}.
Synthetic data for autonomous driving relies on virtual environments of large-scale urban and traffic scenes \cite{dosovitskiy2017carla} for recognition tasks \cite{shafaei2016play} and object detection \cite{hurl2019precise}.
The robotics domain requires virtual environments \cite{tremblay2018deep}, \cite{nvidiaIsaacSim} for recognition tasks such as semantic segmentation \cite{khan2019procsy} and object detection \cite{buls2019generation}.
In all three domains, the focus is on large scale geometric features while simulation of small scale details and textures is not necessary.

Surface inspection belongs to industry automation, which is closely linked to robotics, but includes tasks beyond those needed for movement, localization and mapping. 
In particular, building virtual environments for inspection planning and the development of defect recognition algorithms require a high level of realism and small-scale surface details. This makes this domain highly different from the domains discussed above.

A general-purpose dataset generator is introduced by Greff et al. \cite{greff2022kubric}. It promises realism (using physically-based ray tracing), scalability and reproducibility which are also covered by our pipeline. However, being a general framework, it requires a user to define a 3D scene for the particular task using scripting. Assets for the 3D scene must be either created by the user or imported from general-purpose asset libraries. As such, data generation for specialized problems, such as quality assurance, is challenging due to  unavailable assets, assets with unsatisfactory features or limited modeling capabilities of a user.

Moonen et al. \cite{moonen2023cad2render} presented a toolkit for synthetic image data generation in manufacturing.
Synthesizing datasets for industrial inspection using a rule-based approach was further discussed in \cite{wieler2007weakly,bao2023miad,boikov2021synthetic,roovere2022dimo,hoog2023cad2x}.
Synthetic data was further used for various quality assurance tasks such as inspection of industrial components \cite{saiz2021synthetic}, metal surface inspection \cite{yang2021mask2defect}, industrial visual inspection \cite{abubakr2021learning}, scaffolding quality inspection \cite{kim2022synthetic} or viewpoint estimation \cite{SauerKI2022}.
Raymond et al. \cite{raymond2016multi} introduced models of scratched metal surface where 
they stacked layers containing different scratch distributions.
However, their model does not represent curving 
scratches nor dents or scratches with varying geometrical depth. 
None of the mentioned works provide the level of control introduced by Bosnar et al. \cite{Bosnar2020SynthesisPipeline} with parameters designed to separately control image acquisition context, texture appearance and defect characteristics.

The work presented in this paper builds on the virtual inspection planning research done by Gospodneti{ć} \cite{GospodneticThesis2021} and Bosnar et al. \cite{Bosnar2020SynthesisPipeline, Bosnar2022TextureSynthesis, Bosnar2022DefectModelling}.
Bosnar et al. \cite{Bosnar2020SynthesisPipeline} identified core computer graphics components needed to perform image synthesis for surface inspection and stressed the importance of texture and defect parameterization to minimize the need for artistic approach.
However it focused mainly on the concepts and not challenges which arise when applying it to a realistic scenario.
We extend the aforementioned pipeline and apply it to a realistic scenario.
Procedural textures and defects for generating synthetic data for defect recognition in visual surface inspection were introduced in \cite{Bosnar2022TextureSynthesis,Bosnar2022DefectModelling}. 
Their defect modeling approach was also adopted by Schmedemann et al. \cite{schmedemann2022procedural} for the rule based generated images before application of style transfer.
Fulir et al. \cite{fulir2023syntheticclutch} also built on top of  \cite{Bosnar2020SynthesisPipeline,Bosnar2022TextureSynthesis,Bosnar2022DefectModelling} to produce a synthetic dataset for defect segmentation on metal surfaces but discussed mainly ML results.
\subsection{Decomposition of Scales}
\label{sec:image_synthesis:decomposition_of_scales}

A 3D scene contains \textbf{3D objects} which are traditionally decoupled into geometry and material.
The geometry specifies the size, position and shape of an object (high-level appearance), while the material influences how the light interacts with its surface (detailed appearance).
Alternatively, a 3D object can be considered at three different scales: macro, meso and micro scale.
The \textbf{macro scale} refers to object geometry (e.g., represented by a mesh) and large geometrical features such as defects.
\textbf{Meso scale} refers to the surface structure on a much smaller scale than the shape of the object, but larger than the wavelength of light (i.e., surface texture), as discussed by Dorsey et al. \cite[ch. 2]{dorsey2010digital}. 
\textbf{Micro scale} refers to the surface structure and properties which are not separately distinguishable by the imaging sensor, but are contributing to light reflection.

It is difficult to make a strict distinction between the scales using units, because the scales are relative to the imaging sensor resolution and viewing distance.
Therefore, we propose the following guidelines. The macro scale should be attributed to geometrical shapes which take larger portions of the image. The micro scale can be determined using the Nyquist-Shannon sampling theorem \cite{Shannon1949} stating that \textit{the sampling rate must be at least twice the bandwidth of the signal}. In other words, every feature which cannot be sampled by more than one pixel falls under the micro scale class.
Finally, the meso scale constitutes features which occupy a smaller but visible portion of the image, measured in a handful of pixels.

To give some examples, assume we were to observe an outdoor scene of a park containing people and trees. Then the shapes of people and trees would make the macro scale, whereas the eyes or leaves would already fall under the meso scale, and the skin or the leaf veins would be considered micro scale.
In the case of visual surface inspection, we are looking at specific parts of an object with high resolution. 
Therefore, the inspected object is in the macro scale, its defects and surface texture caused by the manufacturing process are in the meso scale, while roughness and smallest surface features are part of the micro scale. 

\subsection{Rule based image synthesis}
\label{sec:methods:image_synthesis:3d_scene_modeling}
\begin{figure}
    \centering
    \includegraphics[width=\columnwidth]{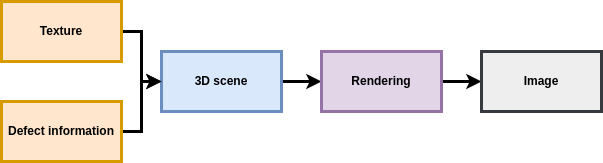}
    \caption{Image synthesis overview. Texture and defect information is joined with the 3D scene to perform rendering and generate an image. The defect information is varied to perform photo-realistic image synthesis of both defected and defect-free object instances. In case of defected instances, pixel-precise defect annotations are automatically created.} \label{fig:methods_image_synthesis:image_synthesis_overview}
\end{figure}

Rule based image synthesis (\cref{fig:methods_image_synthesis:image_synthesis_overview}) can be divided into two main steps:
3D scene modeling (i.e., modeling of the virtual environment) and rendering procedures as discussed by Greenberg et al. \cite{greenberg1997framework} and Tsirikoglou \cite{Tsirikoglou2020Survey}. 
During rendering, the 3D scene is used to generate 2D images.
The crucial element for achieving realism of those images is physically-based light transport, using path-tracing (Kajiya \cite{kajiya1986rendering}, Pharr et al. \cite{pharr2016physically}).
In this work, we use appleseed \cite{francois_beaune_2019_3456967} rendering engine which implements these principles.

\subsection{3D scene modeling}
The \textbf{3D scene} represents the real world context within which the images are acquired.
As such, the scene to be simulated must faithfully represent the environment. For surface inspection this includes 3D objects, lights and cameras.
Note that we do not necessary include environment illumination, since inspection systems often have a strictly controlled acquisition environment and no uncontrolled illumination.
While all the components contribute to the image realism, in this work we focus on the realism of the 3d object, influenced by the geometry, material and surface topography (i.e. texture).

The \textbf{geometry} of the 3D models representing the inspected object (macro scale) must correspond to the geometry of the real inspected product.
In this work we use the 3D model of \textit{test bodies} (\cref{sec:method:test_body_design}) and create defected product instances by augmenting defects (meso scale) directly into the product geometry \cite{Bosnar2022DefectModelling}.
This is done because the defects are larger than the surface texture, and their complex shape causes intricate light scattering that can not be captured by normal maps alone, as described in \cref{sec:defect_modeling}.

\textbf{Material} properties describe the light-surface interaction and are modeled the bidirectional reflectance distribution function (BRDF) as discussed by Dorsey et al. \cite{dorsey2010digital}.
For this work, rough metal surface BRDF is used (Walter et al. \cite{walter2007microfacet}, Kulla et al. \cite{kulla2017revisiting} and Turquin et al. \cite{turquin2019practical}).
The BRDF is evaluated in every surface point based on the local surface normal.
Unrealistically smooth surface is avoided through variation of BRDF parameters and normals using texture.

\textbf{Texture} is a term for which the exact definition differs across domains.
Here we refer to the computer graphics concept which defines it as a procedure to introduce the surface topography by influencing (perturbing) the surface normal and BRDF parameters for every surface point at rendering time (Mikkelsen et al. \cite{mikkelsen2010bump}).
As such, the texture describes the meso scale surface patterns, which would otherwise be very inefficient to be represented by the geometry.
The \cref{sec:texture_modeling} offers a more complete overview of the texture modeling approaches.
Traditionally, texture modeling is aimed to be used by artists to approximately recreate photorealistic visual appearance and enable their creative expression (Guerrero et al. \cite{guerrero2022matformer}, Adobe Substance \cite{substanceAdobe}, Ebert et al. \cite{ebert2003texturing}).
Therefore, the surface models are based more on artist oriented parameters, rather then the real-world parameters, making them visually appealing but not necessarily correct.
Dong et al. \cite{dong2015predicting} approached realism by fitting a gaussian random field to surface measurements and using it as texture.
It gave impressive results for a single instance of measurements, but offered no further control over the pattern properties.

In contrast to \cite{dong2015predicting}, we recreate the surface measurements using mathematical models which can be controlled based on real-world parameters (see \cref{sec:texture_modeling}).
The textures are generated as 2D images encoding the normal variation, which are then mapped onto the surface of the object as explained in \cref{sec:metods:image_synthesis:texture_application}.
Given the high level of details produced by the models, we keep the BRDF roughness parameter fixed to a constant small values (e.g., within a range (0.0-0.1)).

\begin{figure}
    \centering
    \includegraphics[width=0.3\textwidth]{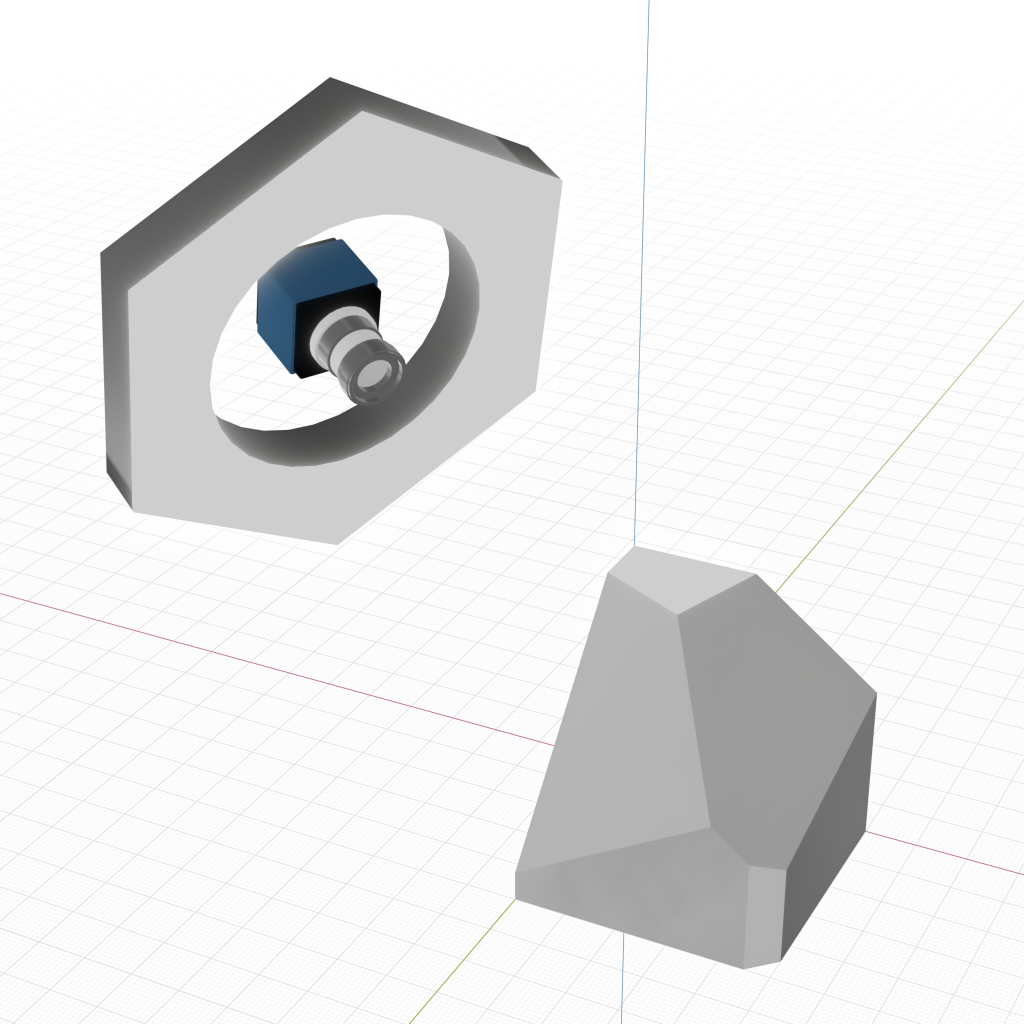}
    \caption{Simulated 3D scene representing the real inspection environment.} \label{fig:methods_image_synthesis:3d_scene}
\end{figure}

\textbf{Light} in the 3D scene represents the illumination devices used in the surface inspection environment. 
We use a model of the ring light geometry used in the real inspection system which has a torus-like shape (\cref{fig:methods_image_synthesis:3d_scene}) with diffuse emission distribution describing the light emission in each point of the shape. As in the real setup, the camera is placed in the center of the ring with both objects having the same orientation.

The \textbf{camera} is using a pinhole camera model defined by its resolution, pixel size, focal length, position and orientation parameters.
Simulated camera parameters are set to resemble the real camera parameters used in the surface inspection environment. 
The pinhole model does not account for the depth of field or optical aberrations such as distortions.
For the purpose of this work, this is sufficient since the inspected surface is expected to be in focus and effects such as image distortion or global blurring can be introduced in post processing or during training in form of image augmentations.

\subsection{Texture mapping}
\label{sec:metods:image_synthesis:texture_application}

The textures used in this work are represented as normal map images where each pixel encodes a normal vector.
While \cite{bosnar2021texture} suggests that the implicit procedural generation is ideal, it requires an additional level of restrictions and complexity when developing the new models.
In particular, it is the limited spatial information which makes describing spatially-varying patterns with regular global structures such as the milling patterns in the physical test bodies (\cref{fig:image_synthesis:TexturesConsidered}) challenging.
While the implicit implementation is possible, it is outside the scope of this work in order to allow focusing on physically realistic texture modeling.

In order to use the texture images together with BRDF evaluation during rendering, it is required to map them onto the 3D object.
Typically either it is done by using projection directly \cite{pharr2016physically} (e.g., planar, cube, spherical) or by unwrapping the object \cite{hormann2007mesh} (i.e., mapping the object faces to a plane).
Applying the texture projection directly onto the whole object will cause artifacts such as texture stretching for complex geometries.
This is not acceptable for generating synthetic training data.
Therefore, in this work we first separate the geometry into main planar surfaces and then use planar projection separately per surface.
To avoid the occurrence of texture tiling (visible repetition in pattern), the texture images are generate large enough to cover the entire target surface.

\begin{figure*}[ht]
    \centering
    \begin{subfigure}[t]{0.24\textwidth}
        \centering
        \includegraphics[width=0.99\textwidth]{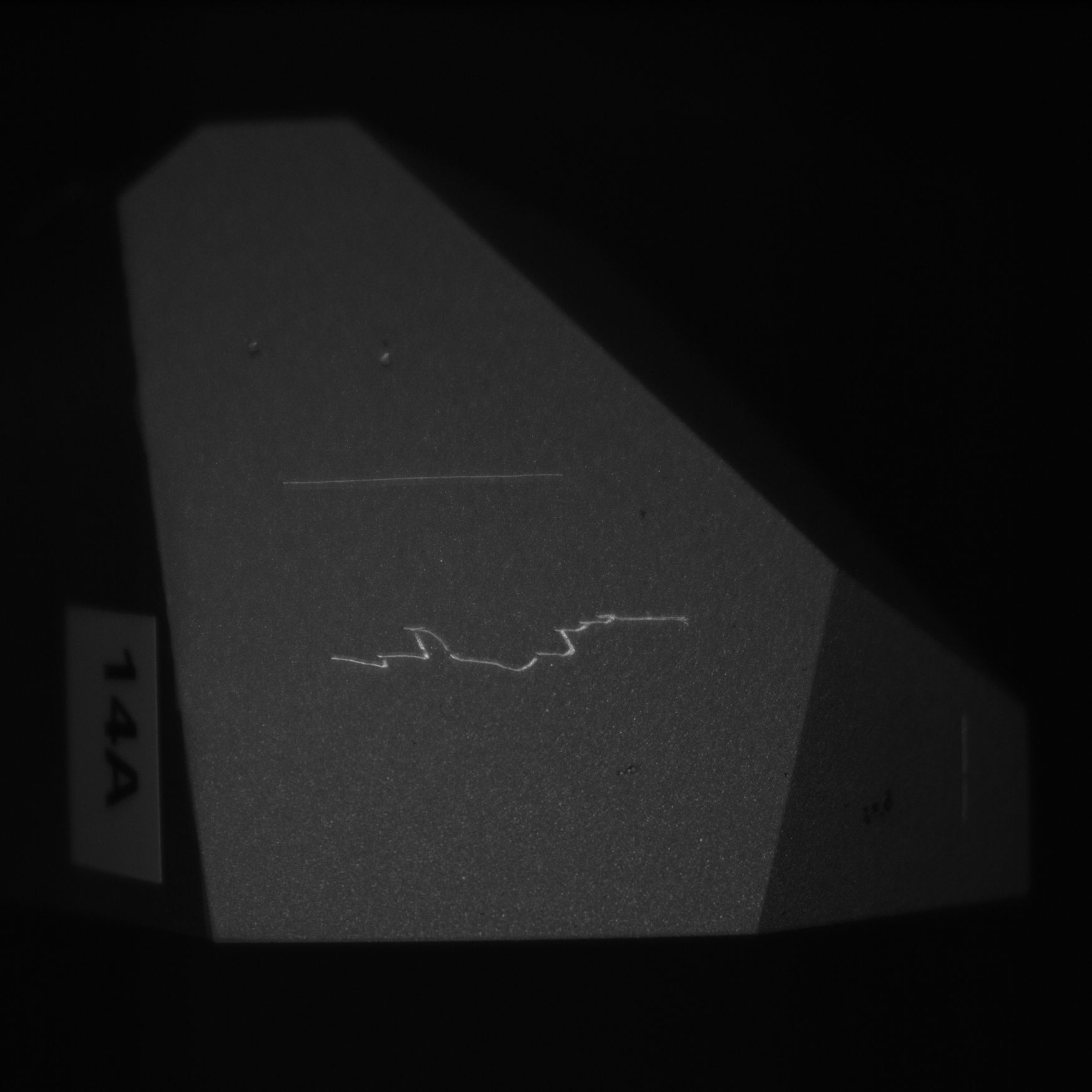}
        \vspace{-1.2\baselineskip}
        \label{fig:image_synthesis:real_defected_1}
    \end{subfigure}
    \begin{subfigure}[t]{0.24\textwidth}
        \centering
        \includegraphics[width=0.99\textwidth]{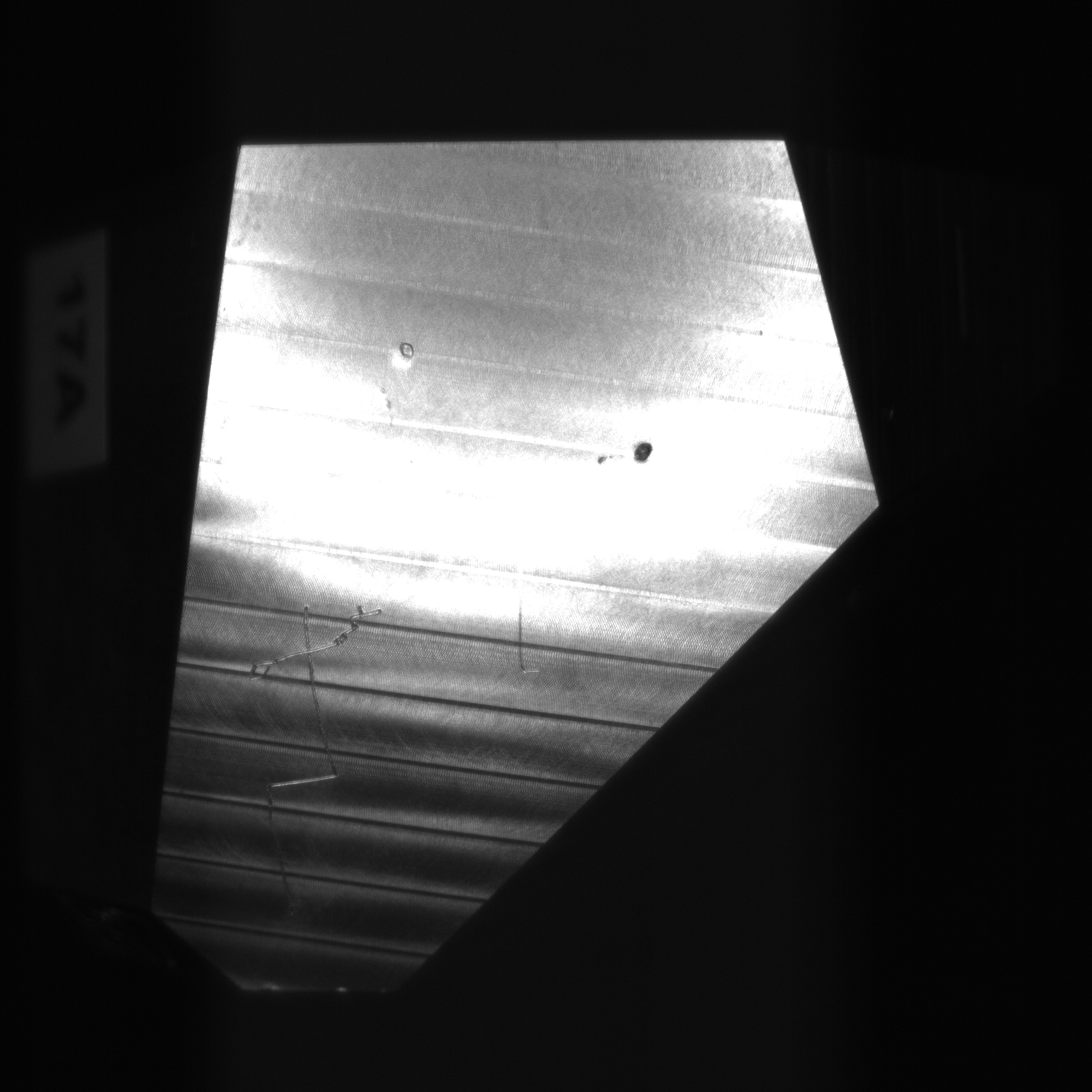}
        \vspace{-1.2\baselineskip}
        \label{fig:image_synthesis:real_defected_2}
    \end{subfigure}
    \begin{subfigure}[t]{0.24\textwidth}
        \centering
        \includegraphics[width=0.99\textwidth]{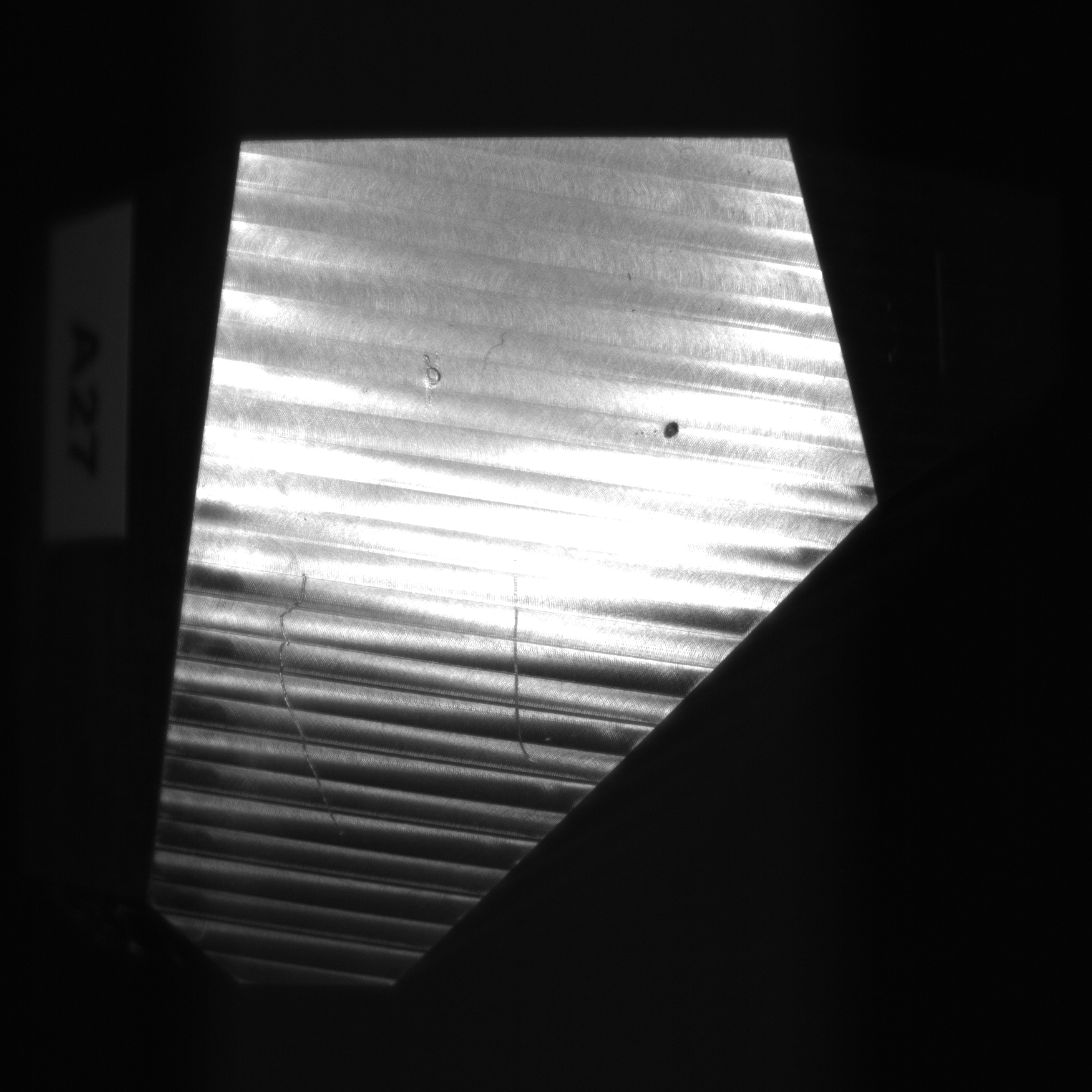}
        \vspace{-1.2\baselineskip}
        \label{fig:image_synthesis:real_defected_3_1}
    \end{subfigure}
    \begin{subfigure}[t]{0.24\textwidth}
        \centering
        \includegraphics[width=0.99\textwidth]{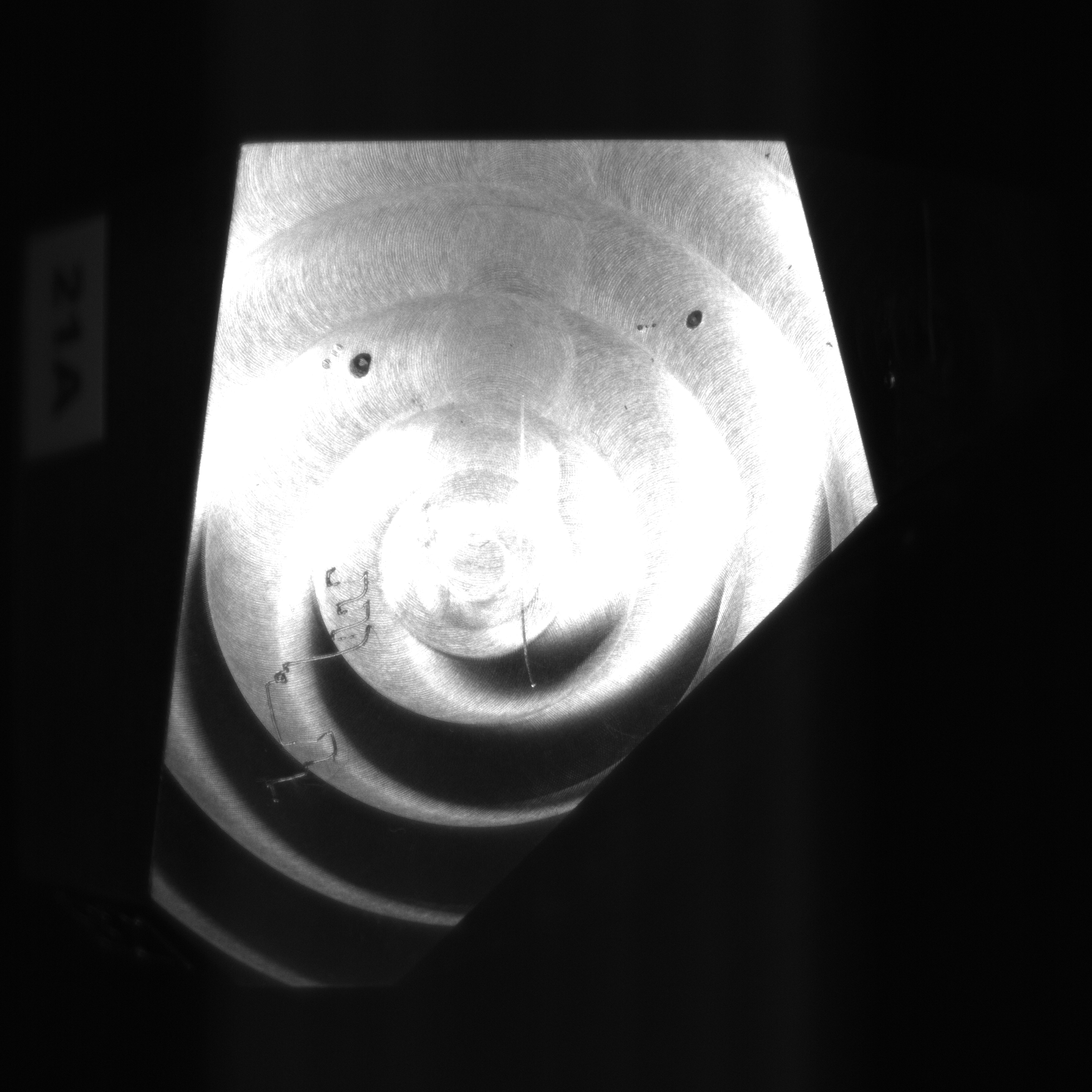}
        \vspace{-1.2\baselineskip}
        \label{fig:image_synthesis:real_defected_3}
    \end{subfigure} \\

    \vspace{1pt}
    
    \begin{subfigure}[t]{0.24\textwidth}
        \centering
        \includegraphics[width=0.99\textwidth]{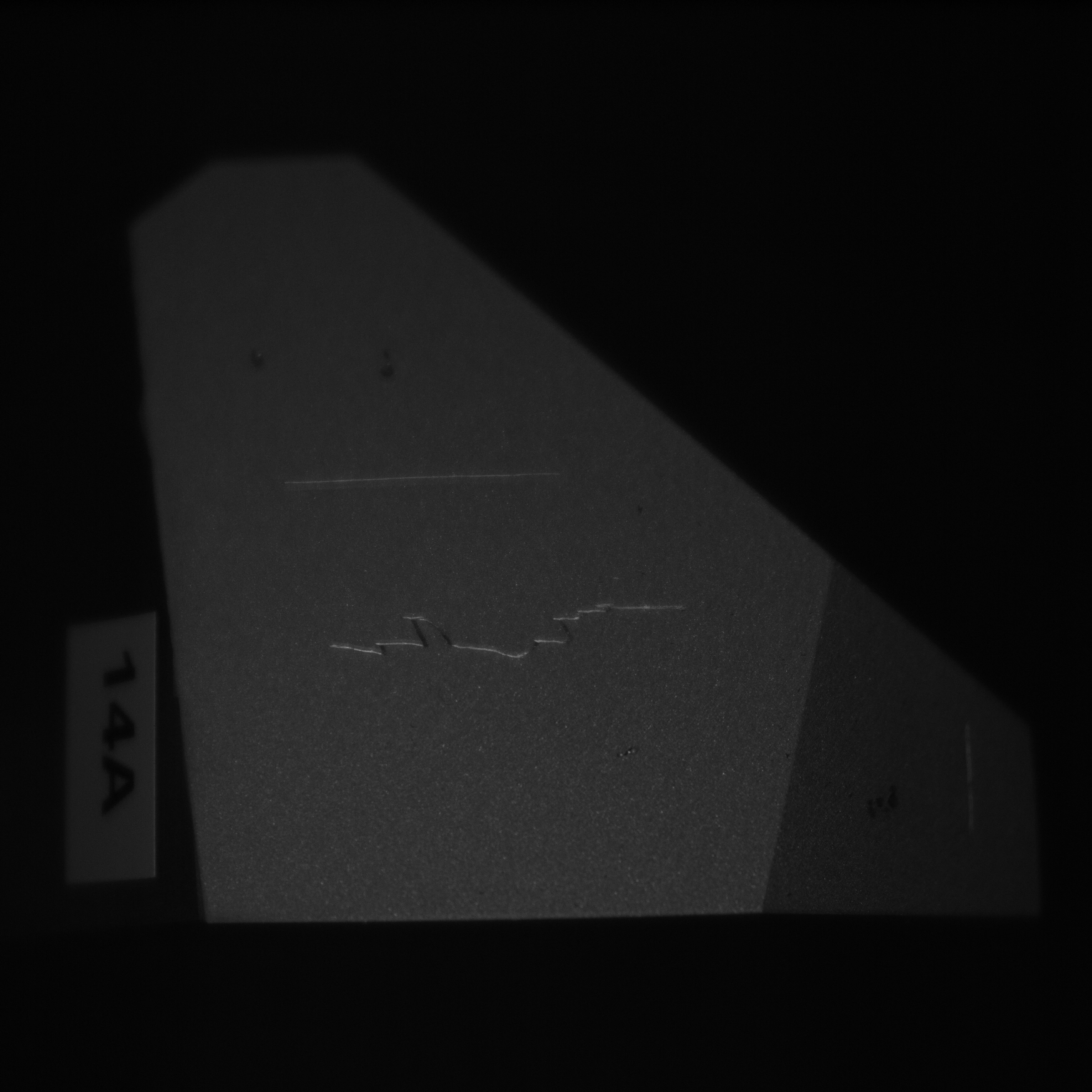}
        \vspace{-1.2\baselineskip}
        \label{fig:image_synthesis:real_defected_4}
    \end{subfigure}
    \begin{subfigure}[t]{0.24\textwidth}
        \centering
        \includegraphics[width=0.99\textwidth]{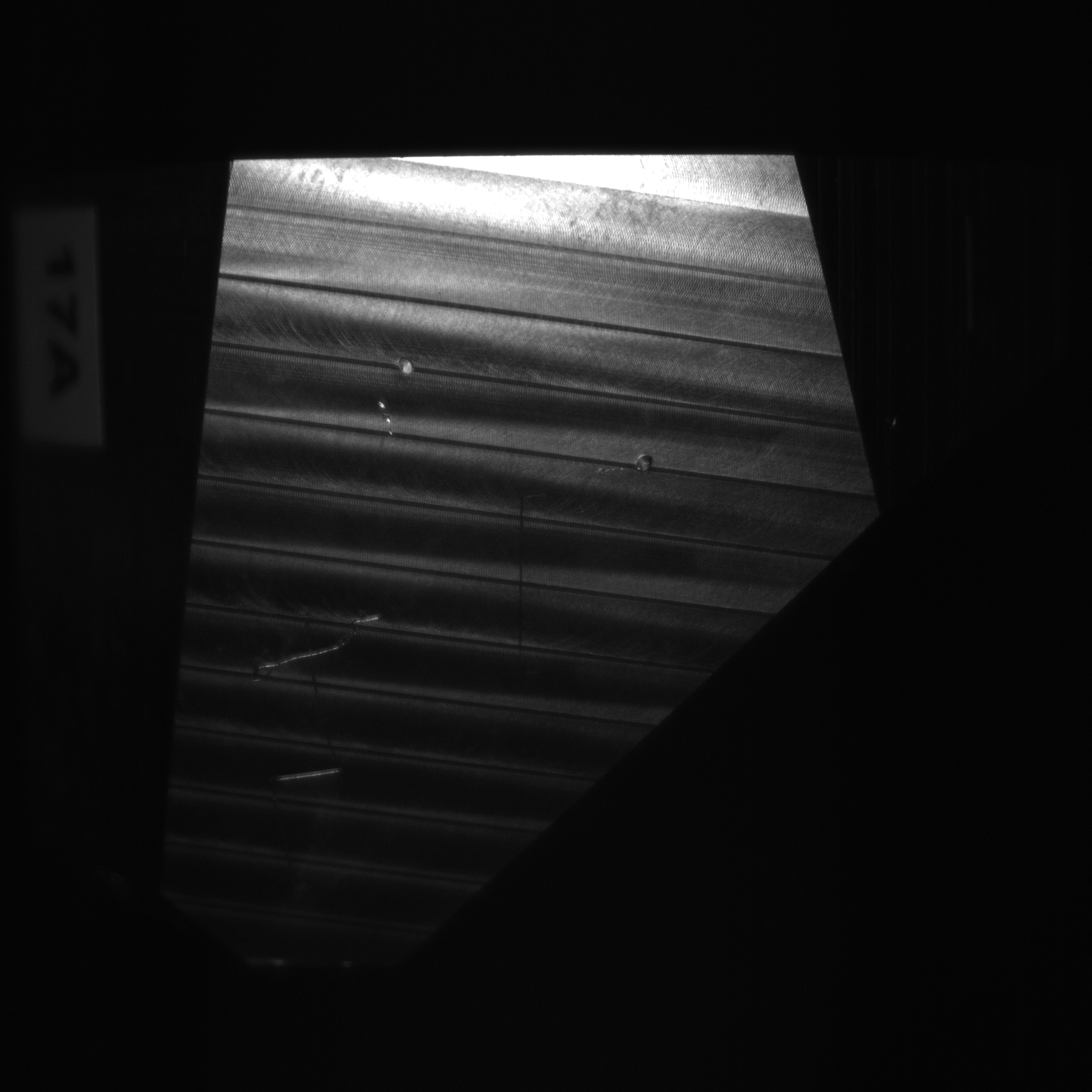}
        \vspace{-1.2\baselineskip}
        \label{fig:image_synthesis:real_defected_5}
    \end{subfigure}
    \begin{subfigure}[t]{0.24\textwidth}
        \centering
        \includegraphics[width=0.99\textwidth]{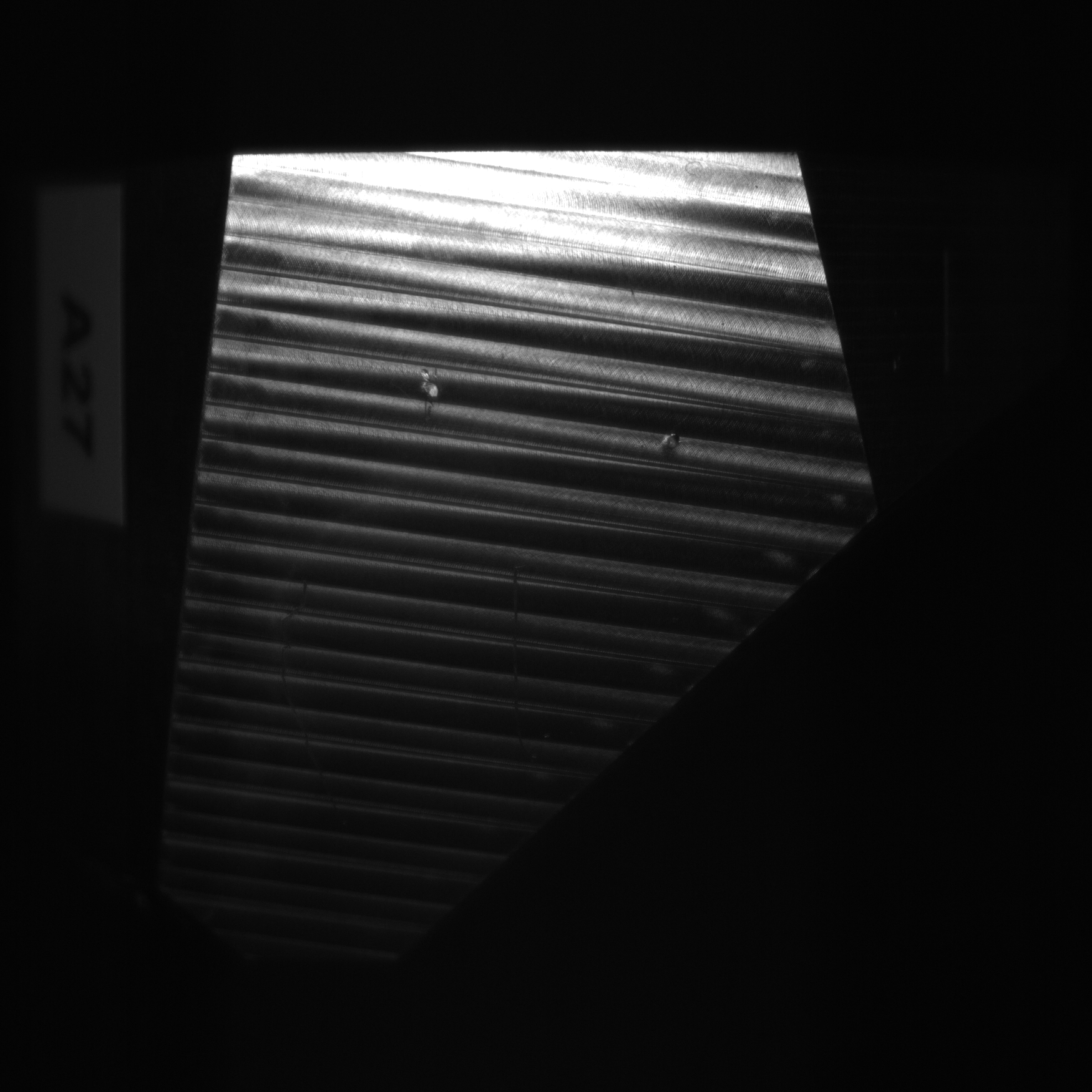}
        \vspace{-1.2\baselineskip}
        \label{fig:image_synthesis:real_defected_7}
    \end{subfigure}
    \begin{subfigure}[t]{0.24\textwidth}
        \centering
        \includegraphics[width=0.99\textwidth]{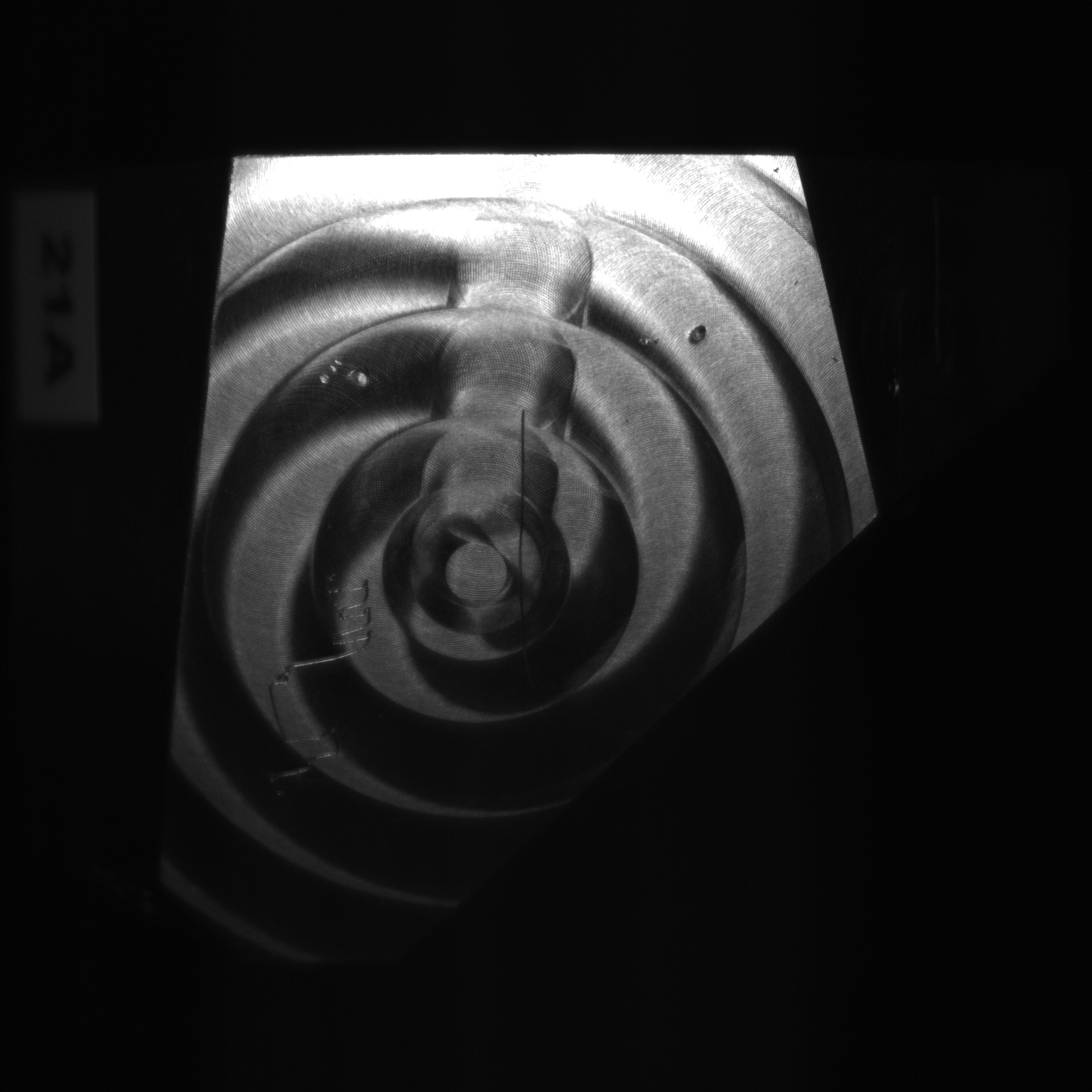}
        \vspace{-1.2\baselineskip}
        \label{fig:image_synthesis:real_defected_6}
    \end{subfigure}

    \caption{Real images of object surface with defects. Top row: 20 degrees angle from perpendicular view. Bottom row: 40 degrees angle from perpendicular view.}
    \label{fig:image_synthesis:real_defected}
\end{figure*}

\subsection{Rendering}

\label{sec:methods:image_synthesis:rendering}
The 3D scene description is used to render the synthetic images from all given camera positions.
The main rendering parameters influencing the quality of synthesized images are the number of samples per pixel (SPP) and the number of light ray bounces.

In surface inspection each pixel covers rich and complex surface structure.
Higher SPP ensures that the computed pixel color is a good approximation of the mean light response value of the covered area.
As such, increasing the number of SPPs will reduce noise and aliasing in the rendered image.

In addition to texture, our 3D objects contain defects in form of small, complex, geometrical features.
Increasing the number of light bounces (i.e., times the ray is allowed to bounce off the intersection point) enables to represent 
the complex light phenomena such as shadowing, masking or interreflections.
This phenomena can not be correctly represented using normal maps alone.

\subsection{Defect annotations generation}
\label{sec:methods:image_synthesis:object_shape_defecting}

\begin{figure*}
    \centering
    \includegraphics[width=0.3\textwidth]{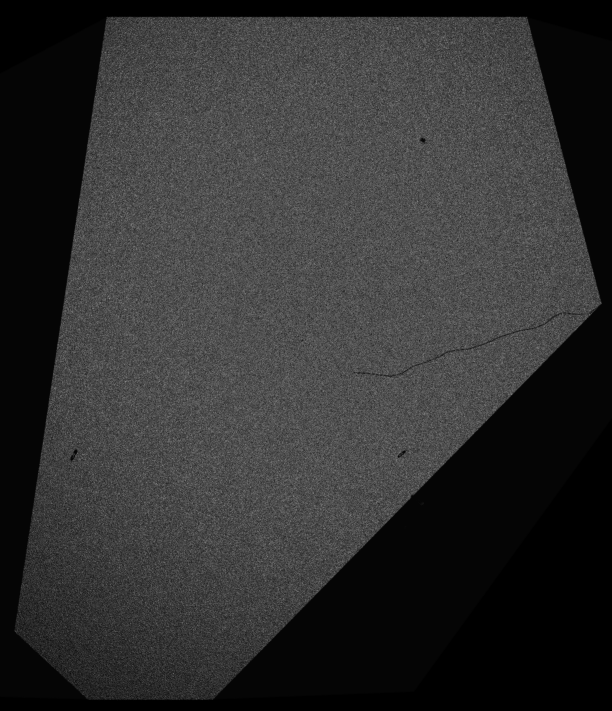}
    \includegraphics[width=0.3\textwidth]{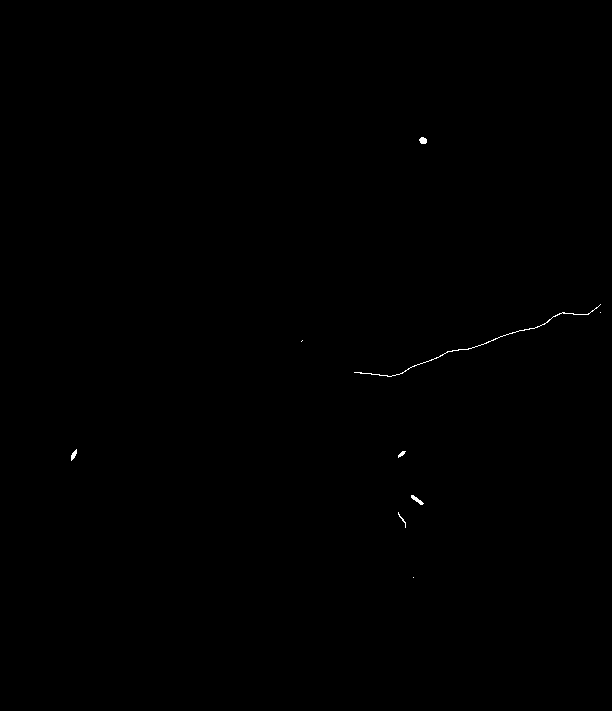}
    \includegraphics[width=0.3\textwidth]{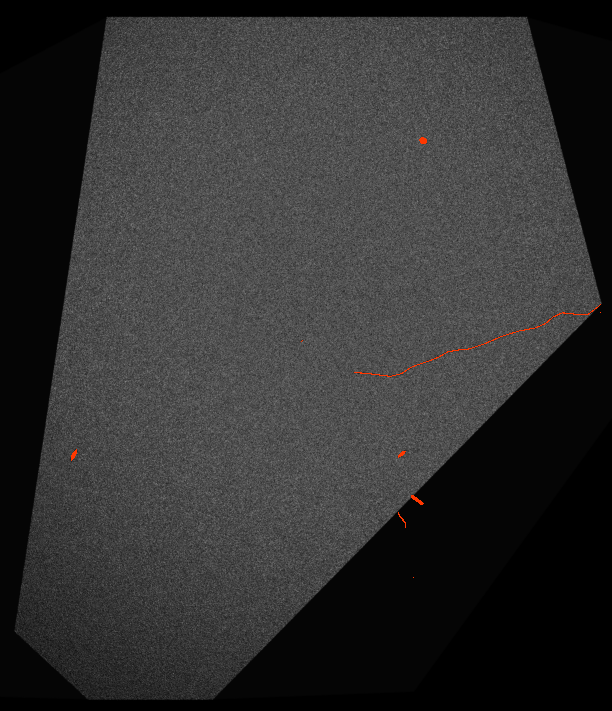}
    \caption{Left: defected synthetic image. Middle: defect annotations. Right: defected synthetic image with annotations overlay.}
    \label{fig:image_synthesis:synth_defected_annotated}
\end{figure*}

The object geometry used in the 3D scene is defected using methods based on work by Bosnar et al. \cite{Bosnar2022DefectModelling}.
Each imprinted defect geometry has the corresponding geometrical defect mask \cref{sec:defect_modeling}.
Black (non-emissive) material is applied on the 3D object, emissive material is set to the geometrical masks and all light sources are disabled.
Finally, the annotations are generated using the same camera parameters as for the corresponding photorealistic image, which ensures pixel-precise labels of the defect areas visible from the camera position and not occluded by the 3D object.
This approach is introduced by Bosnar et al. \cite{Bosnar2022DefectModelling} and will always generate defect annotations regardless of their light response, as can be seen in \cref{fig:image_synthesis:synth_defected_annotated}.

\section{Test body design}
\label{sec:method:test_body_design}
Injection of the domain knowledge is crucial for developing photorealistic models.
Therefore, we use custom designed and manufactured test objects, giving us complete insight into and control over the processing chain. This allows us to focus on the image synthesis, surface modeling, dataset generation and training challenges.

\begin{figure}
    \centering
    \begin{subfigure}[b]{\linewidth}
        \centering
        \includegraphics[width=0.5\linewidth]{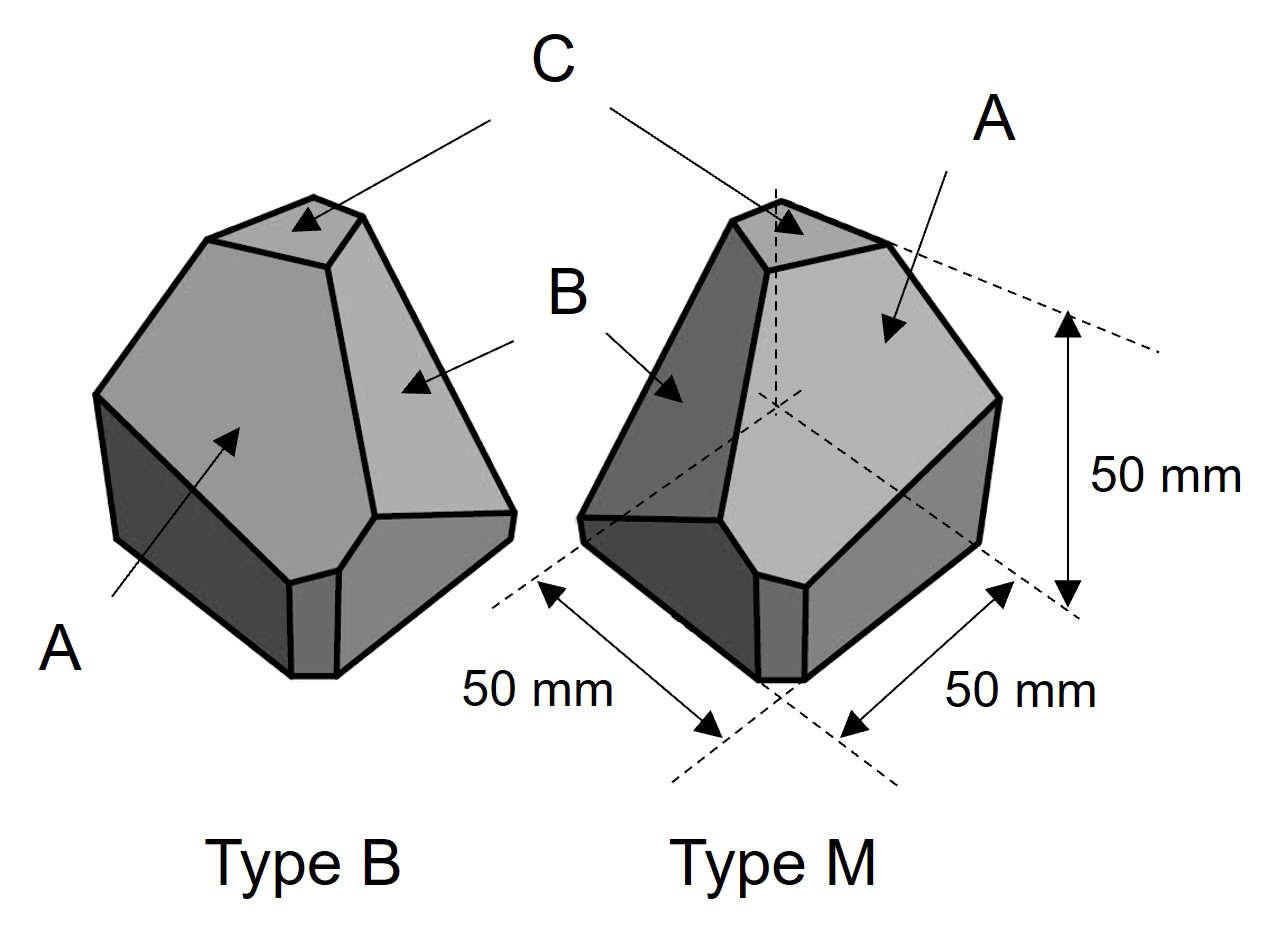}
        \caption{Drawing of the base test bodies.}
        \label{fig:manufacturing:polygons}
    \end{subfigure}
    \begin{subfigure}[b]{\linewidth}
        \centering
        \includegraphics[width=0.7\linewidth]{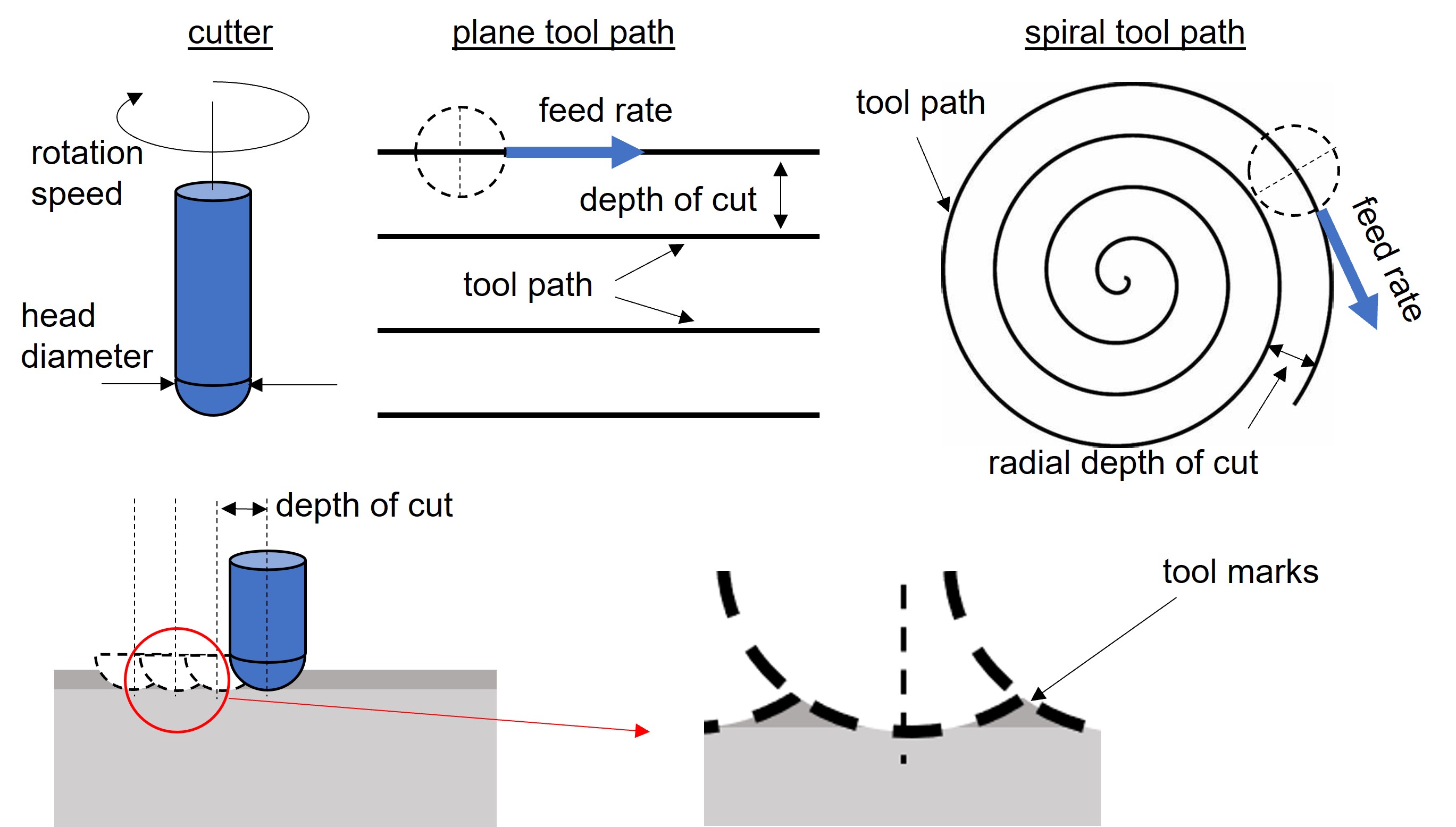}
        \caption{Sketch of the milling processes and its main parameters.}
        \label{fig:manufacturing:details}
    \end{subfigure}
    \begin{subfigure}[b]{\linewidth}
        \centering
        \includegraphics[width=0.7\linewidth]{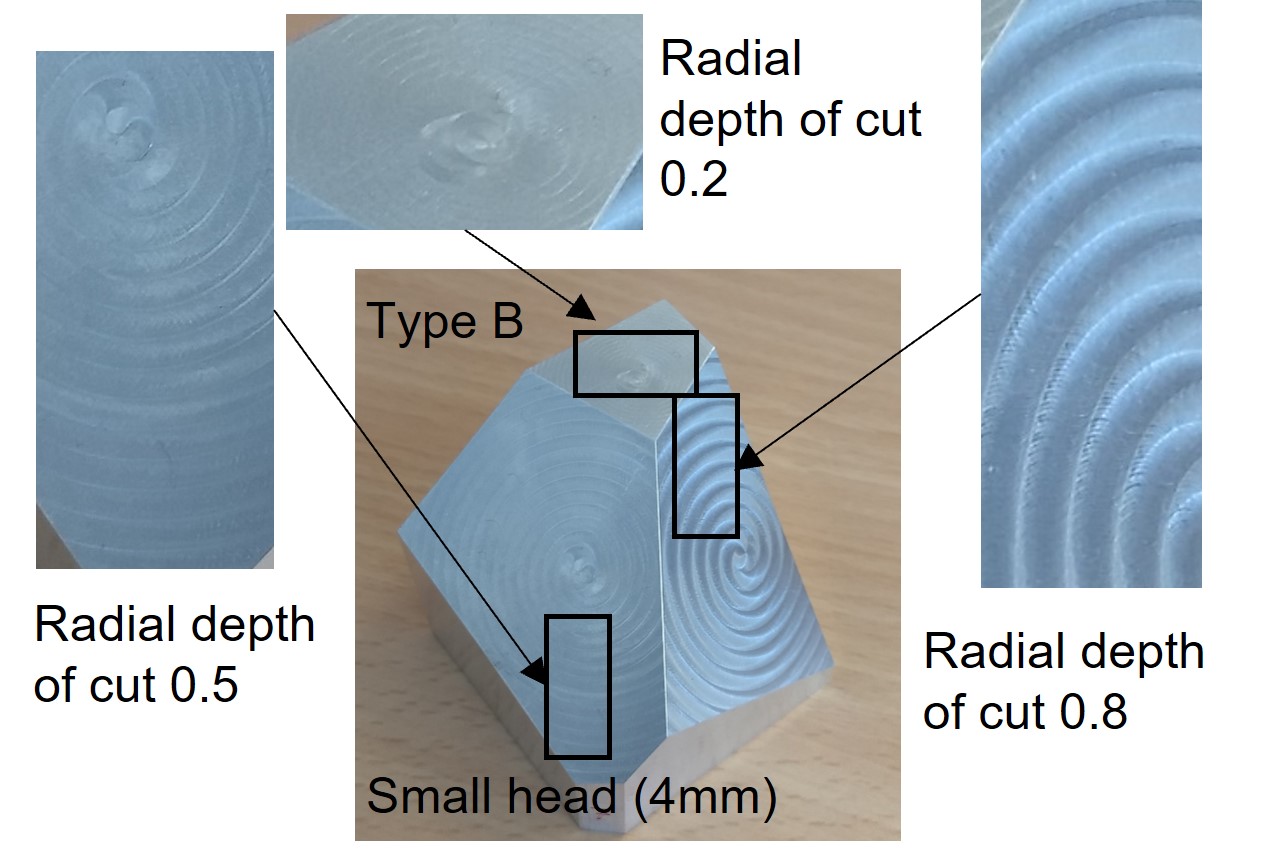}
        \caption{Picture of the manufactured surfaces of type B spiral milled with a small head (4 mm) and various radial cutting depths.}
        \label{fig:manufacturing:polygon_detail}
    \end{subfigure}
    \caption{Illustration of the test object used in the project, the milling processes and a test object with milled surfaces.}
    \label{fig:manufacturing}
\end{figure}

In total, three identical sets of test objects were fabricated, each containing 10 polyhedral test bodies.
When designing the test bodies, the aim was to generate bodies with multiple faces showing different realistic surface texture patterns. 

The base bodies (type B and type M) are polyhedrons made from aluminum, see \cref{fig:manufacturing:polygons}.
Each polyhedron has size 5\,cm $\times$ 5\,cm $\times$ 5\,cm with planar faces A, B and C of approximately size 2\,cm $\times$ 2\,cm which are measured and modeled subsequently.
The remaining faces are left “unfinished” (rough milling) and are not further considered.
The difference between type B and type M is the mirrored arrangement of the faces of the base body.

First, the surfaces are milled using different processing parameters (\cref{fig:manufacturing:details}).
A rotating cutter with a specific head diameter is used to remove material from the surfaces by performing many separate, small cuts along the tool path with a defined feed rate.
We consider face-milling, which means that the tool moves perpendicularly to the object surface.
The distance between the neighboring tool paths depends on the radial depth of the cut and influences the resulting tool marks, see \cref{fig:manufacturing:polygon_detail}.
Parallel linear and spiral tool paths are used.
Artifacts in form of exit lines can occur for spiral milling.

After milling, some of the faces were sandblasted, a process in which a hand-held nozzle of a sandblasting tool blasts the surface with a mixture of sand and air at high pressure.
This abrasive method can smoothen or roughen the surface. The resulting surface topography depends thus on the pressure of the air jet, the grain size distribution, shape and material of the sand.

The parameter values of both processing techniques (see \cref{tab:processing_params}) were chosen to result in realistic patterns expected to emerge from industrial processing. 

\begin{table}
\centering
\begin{tabular}{llc}

Technique & Parameter & Values \\ 
\hline
\multirow{3}{*}{milling}                 & milling head diameter & 4 mm, 8 mm \\
\cline{2-3}
                        & radial depth of cut & 0.2, 0.5, 0.8 \\ 
\cline{2-3}
                        & path      & parallel, spiral \\ 
\hline
sandblasting            & pressure  & 2.5 bar, 6 bar \\ 
\end{tabular}
\caption{\label{tab:processing_params} Used parameter settings for processing.}
\end{table}

\begin{figure}
    \centering
    \centering
    \includegraphics[width=0.6\linewidth]{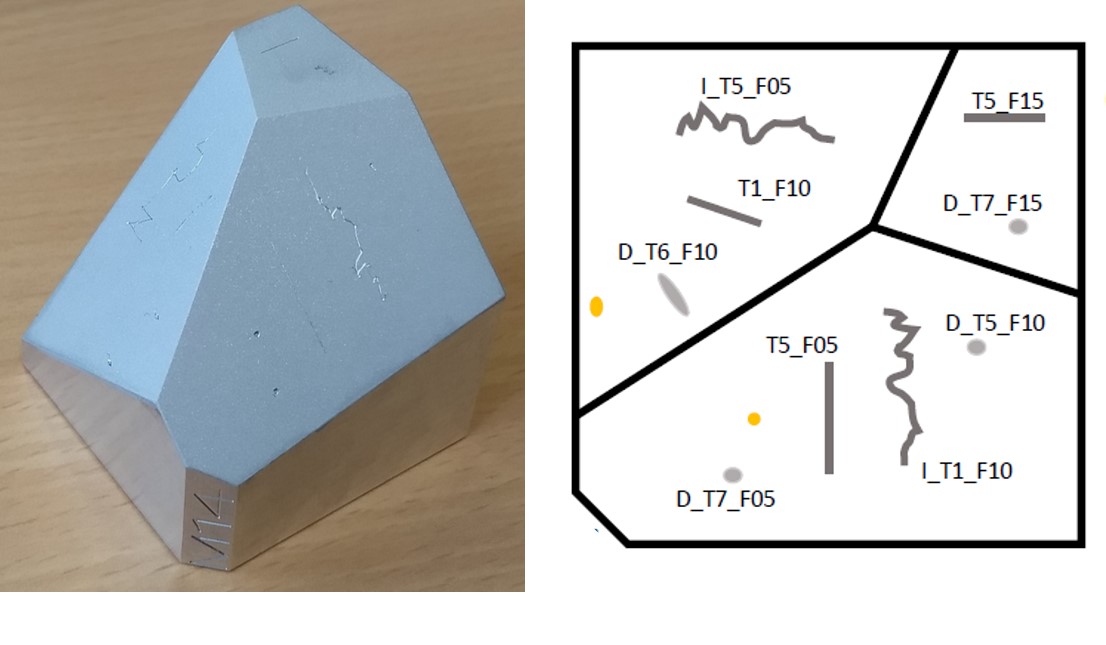}
    \caption{Test object type M after sandblasting and subsequently introduced defects with different types and sizes.}
    \label{fig:manufacturing:defects}
\end{figure}

Objects in set 1 contain no surface defects, while sets 2 and 3 contain more than 200 scratches and localized point defects, created precisely on the faces.
For this purpose, a custom “indenter” and a scratch test tool were used where the tips can be loaded with different masses.
As tips we used a diamond tip, a steel needle and a screw.
The load of the tips were 500 g, 1000 g and 1500 g.
Using the tools, the scratches (straight and free-form lines) and digs (dents) are realized on the surfaces as typical surface defects in accordance to ISO8785.
The generated defect sizes were in the range of sub-millimeter to millimeter.
See \cref{fig:manufacturing:defects} for an example of defect specification.

\section{Material measurements}
\label{sec:method:material_measurement}
In order to characterize the machined surfaces, optical 3d and topography measurements are performed by means of focus-variation microscopy (Bruker alicona InfiniteFocus G4). In the high precision machining industry measurement systems based on the focus-variation technique are standard inspection tools to characterize 3D mechanical parts. The advantage is the high precision and contact-free measurement of roughness, surface structure, micro-geometry and form using one optical sensor.  
The measurement system is based on a precise optical lens system with shallow depth sharpness.
By changing the working distance between the measured surface and the microscope lens, different depths of the surface come into focus and are projected sharply onto the sensor.
The proprietary software analyzes the distance for each point and measures the sharpness, which is used for the calculation of the surface depth profile (topography).

\begin{figure}
    \centering
    \begin{subfigure}[b]{\linewidth}
        \centering
        \includegraphics[height=2.5cm]{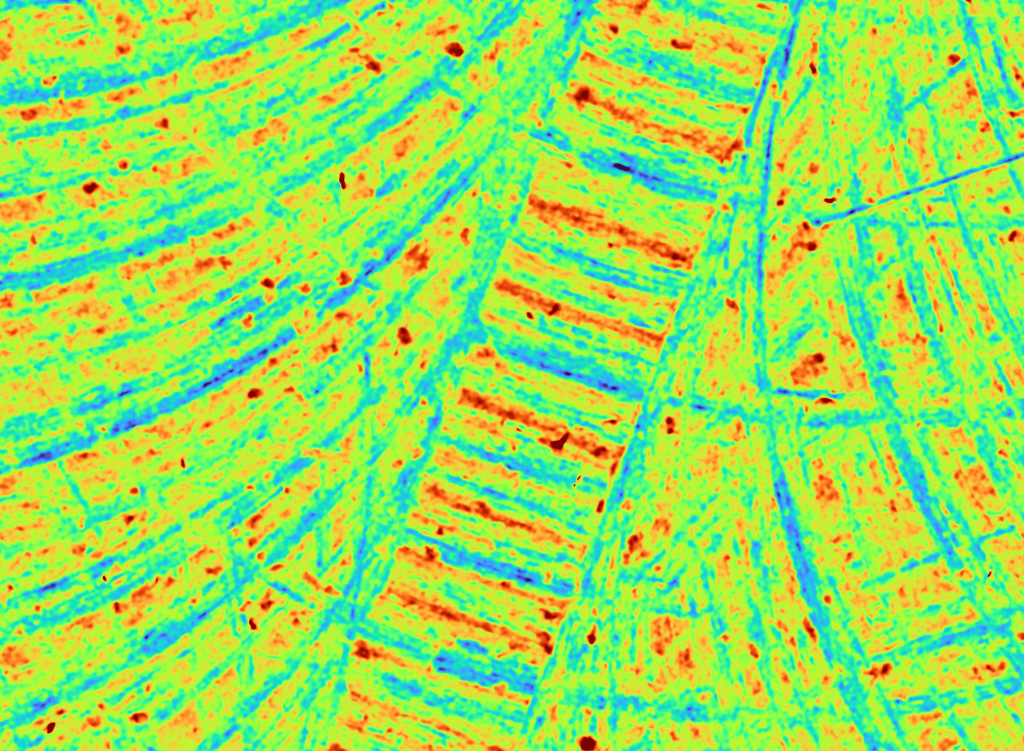}
        \includegraphics[height=2.5cm]{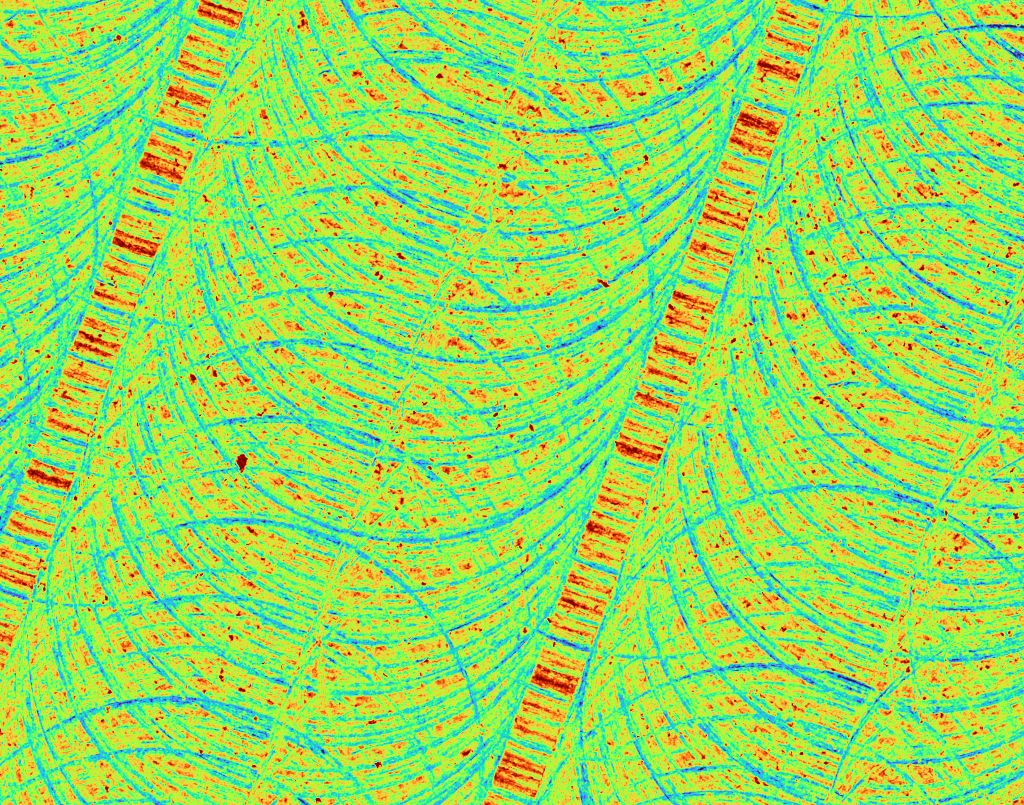}
        \includegraphics[height=2.5cm]{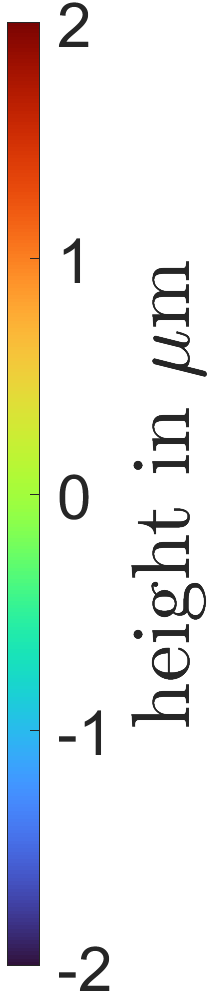}
        \caption{Parallel milled surface using milling head diameter $4$ mm and radial depth of cut 0.5. Small imaged region (left) is $2\text{ mm}\times1.4\text{ mm}$ with pixel size $0.44\, \mu$m and large imaged region (right) is $7.5\text{ mm}\times5.9\text{ mm}$ with pixel size $1.75\,\mu$m. }
        \label{fig:Overview_measurements:mill}
    \end{subfigure}\\[8pt]
    \begin{subfigure}[b]{\linewidth}
        \centering
        \includegraphics[height=2.5cm]{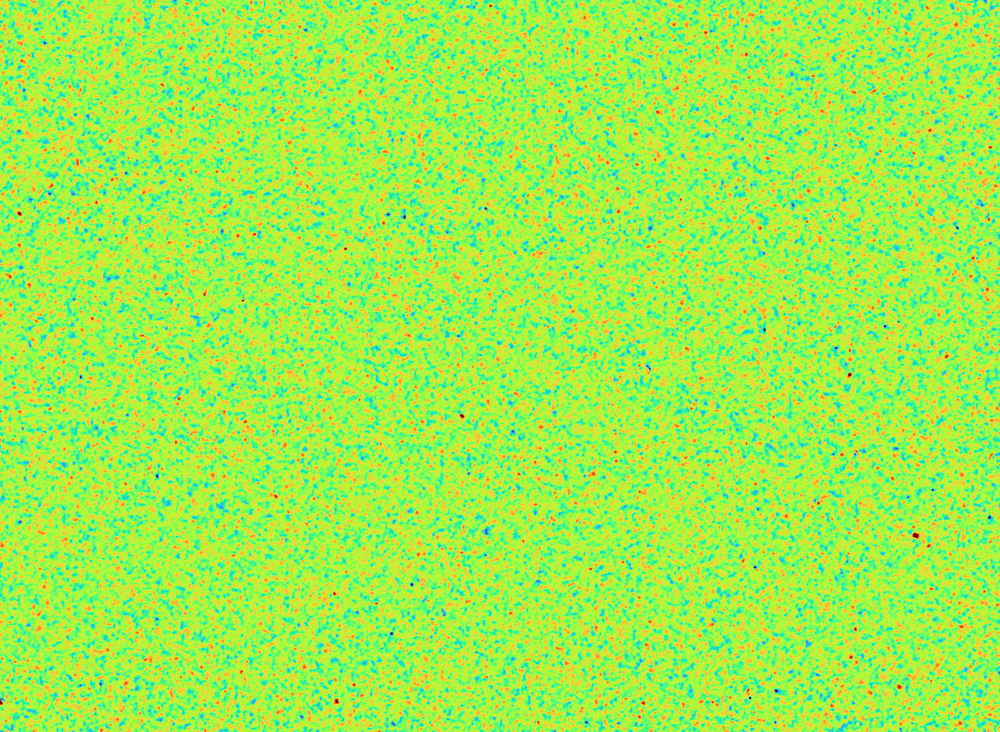}
        \includegraphics[height=2.5cm]{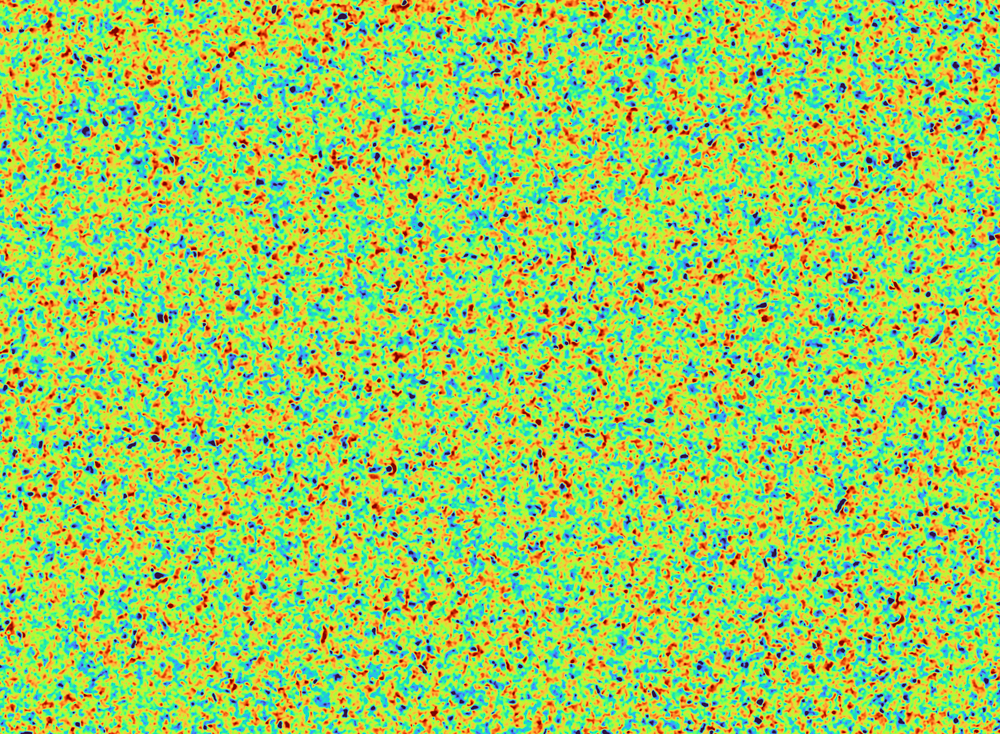}
        \includegraphics[height=2.5cm]{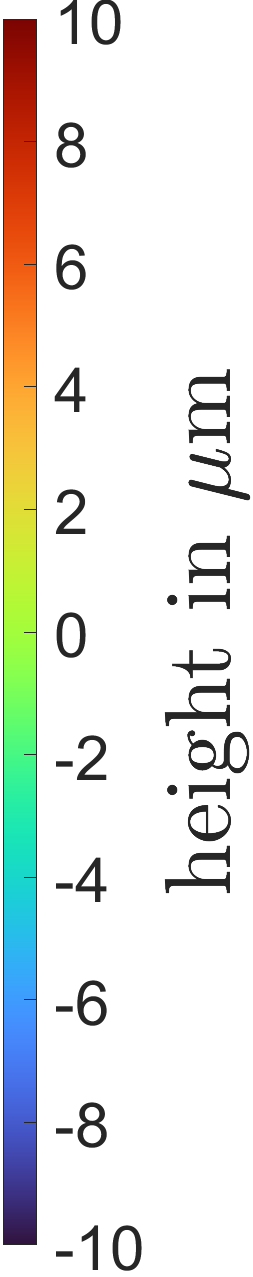}
        \caption{Sandblasted surface using pressure $2.5$ bar (left) and $6$ bar (right). Imaged region is $7.7\text{ mm}\times5.7\text{ mm}$ with pixel size $1.75\,\mu$m.}
        \label{fig:Overview_measurements:sand}
    \end{subfigure}
    \caption{Topography measurements of sandblasted and milled surfaces.}
    \label{fig:Overview_measurements}
\end{figure}

Depending on the magnification of the optical lens system, this process allows a vertical resolution down to 10 nm (magnification equivalent of 100x, which is a physical magnification limit).
For the purpose of this work, an optical lens system with a magnification of 5x and a nominal vertical resolution of 410 nm is used.
The vertical resolution was limited during data acquisition by the measurement software.
The defect free samples of objects in test set 1 are measured with different lateral resolutions, depending on the size of the scanned surface area, as it is shown in \cref{fig:Overview_measurements}.
The topography measurements are converted into ASCII xyz-files in order to create readable files used for surface modeling.
This results in height images, where each pixel is assigned a height value.

\begin{figure}
    \centering
        \begin{subfigure}[b]{0.48\linewidth}
            \centering
            \includegraphics[width=\linewidth]{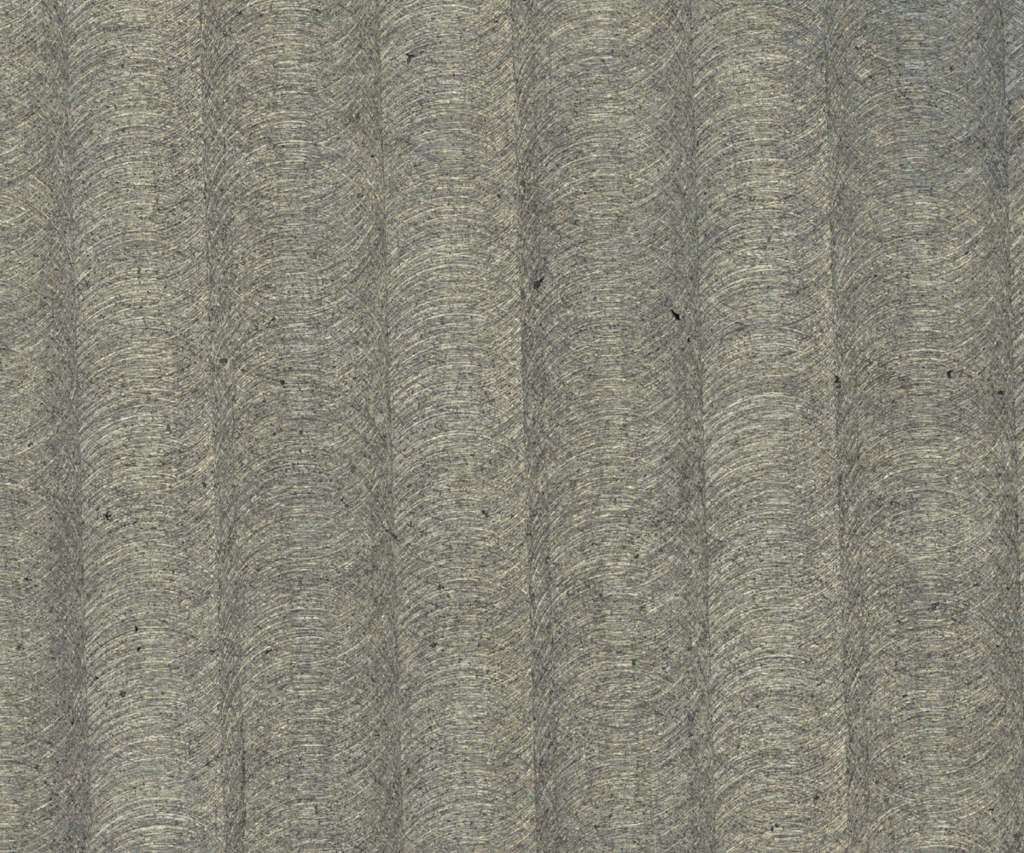}
            \caption*{Parallel milling, head diameter $4$ mm, radial depth of cut 0.8.}
        \end{subfigure}
        \begin{subfigure}[b]{0.48\linewidth}
            \centering
            \includegraphics[width=\linewidth]{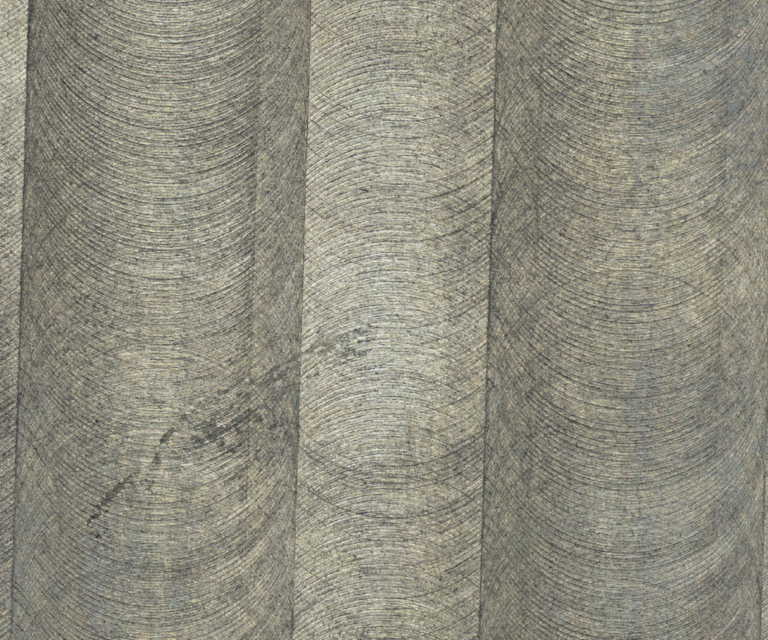}
            \caption*{Parallel milling, head diameter $8$ mm, radial depth of cut 0.8.}
        \end{subfigure}\\[8pt]
        \begin{subfigure}[b]{0.48\linewidth}
            \centering
            \includegraphics[width=\linewidth]{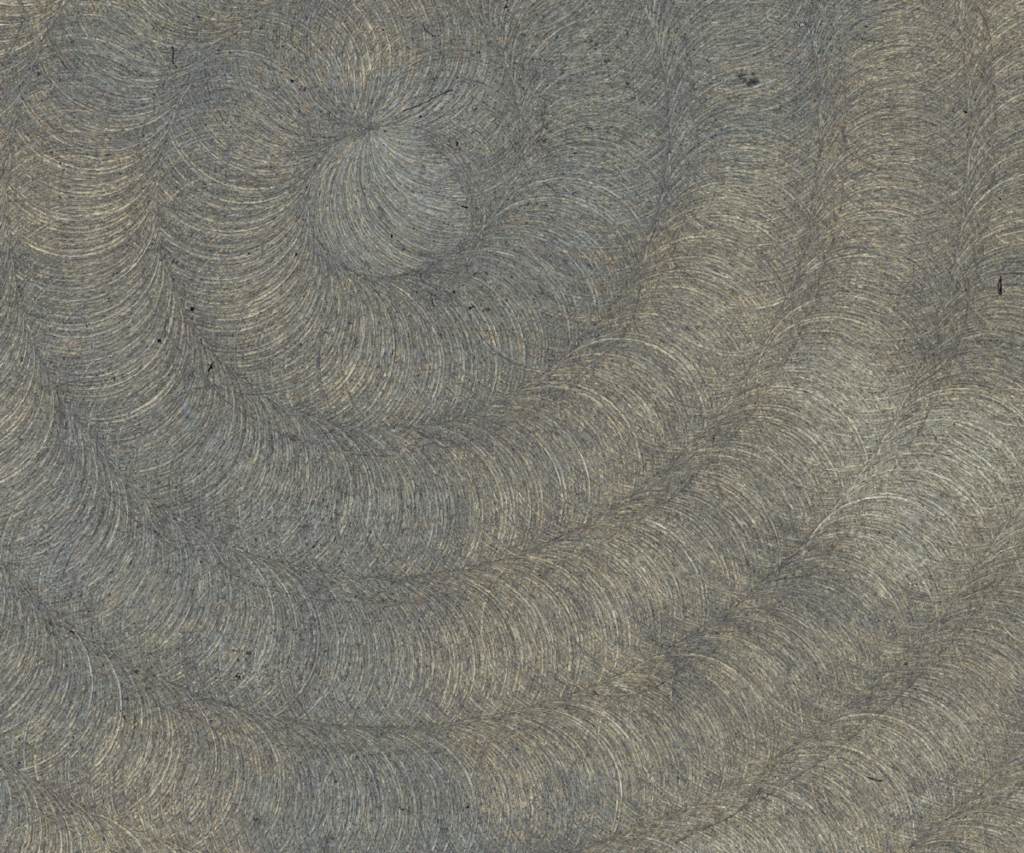}
            \caption*{Spiral milling, head diameter $4$ mm, radial depth of cut 0.5.}
        \end{subfigure}
        \begin{subfigure}[b]{0.48\linewidth}
            \centering
            \includegraphics[width=\linewidth]{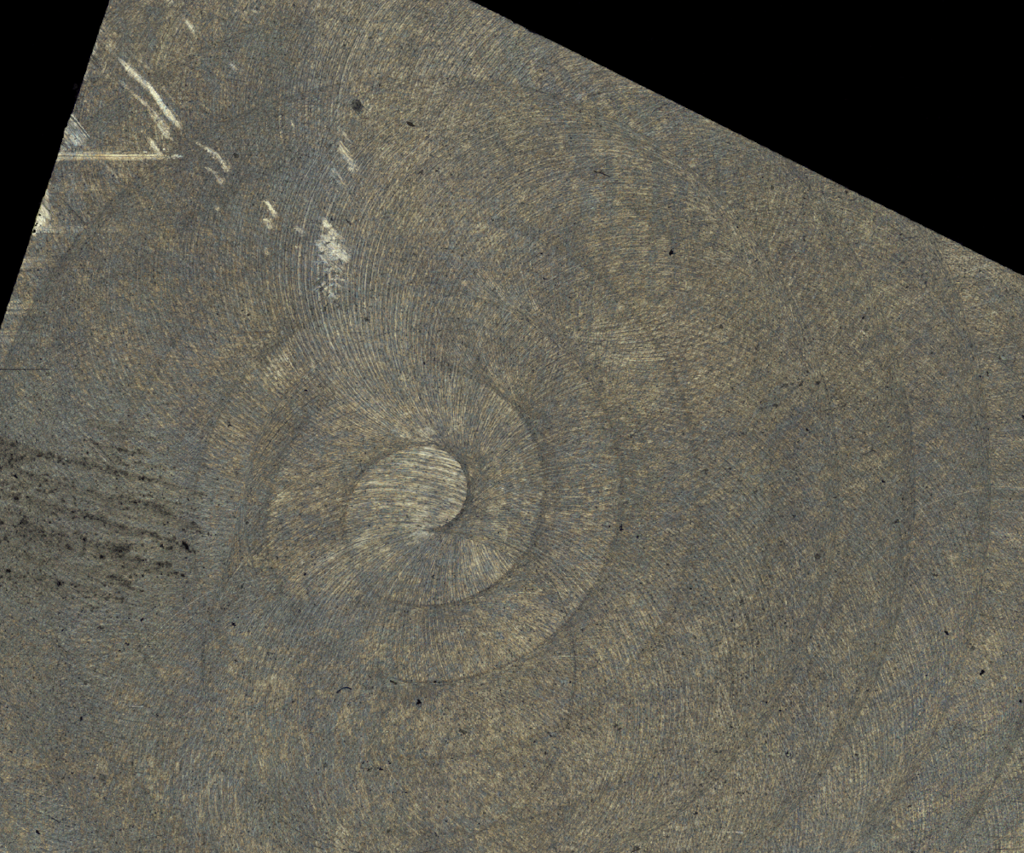}
            \caption*{Spiral milling, head diameter $8$ mm, radial depth of cut 0.2.}
        \end{subfigure}
    \caption{Optical 3d measurements of milled surfaces using different parameter settings. Imaged region is $21\text{ mm}\times17.5\text{ mm}$ with pixel size $7\,\mu$m.}
    \label{fig:MillingPatternOverview}
\end{figure}

Depending on the process parameters, the tool marks created by the milling process form a unique pattern which can be observed by the naked eye (\cref{fig:manufacturing:polygon_detail}) and influences the subsequent surface light response, playing the key role in surface inspection.
Measurements of surfaces processed using different parameter settings are given in \cref{fig:MillingPatternOverview}.

When sandblasting, the previous milling does not affect the final surface pattern, see \cref{fig:Overview_measurements:sand}.
The process results in a homogeneous surface texture without long scale spatial relations.

\begin{figure}
    \centering
    \includegraphics[width=0.99\linewidth]{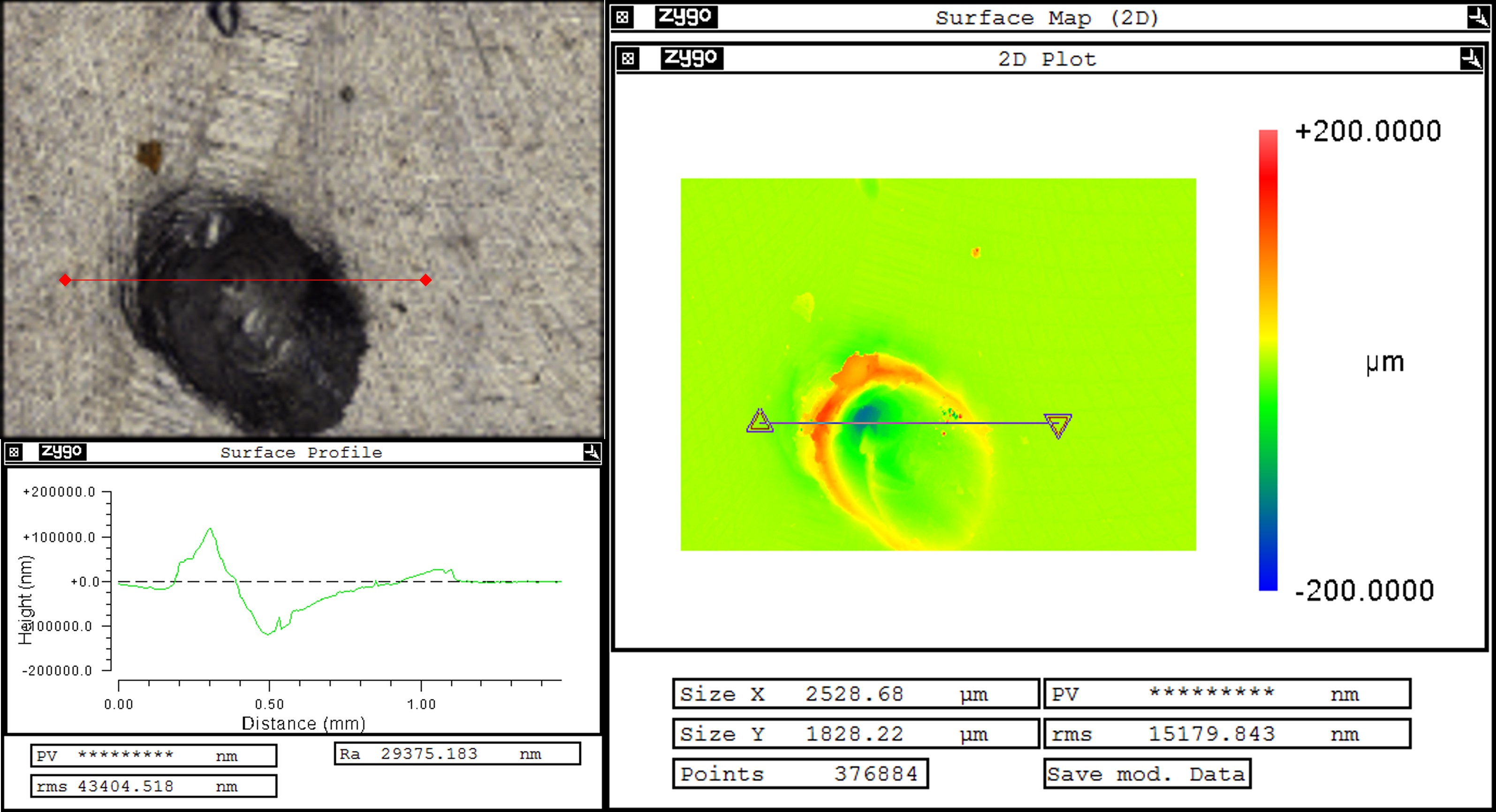}
    \caption{Optical 3d (top left) and topography (right) measurement with 2d intersection of the height profile (bottom left) marked in the topography measurement of a defect created using the indentor with load of 500g.}
    \label{fig:defect}
\end{figure}

In order to characterize the generated defects we used an optical lens system with a magnification of 20x and manually adjusted the vertical resolution during data acquisition through the software.
Depth quality filter value was set between \num{2e-6} and \num{8e-6}, depending on the defect type and size (e.g., \cref{fig:defect}).
The shape of the dent resembles a dig with a peak and a valley of about 150 $\mu$m in height and depth with respect to the undamaged surface level. 
 
\section{Texture Modeling}
\label{sec:texture_modeling}
Due to the very different surface topographies (\cref{fig:Overview_measurements}), separate models are required for sandblasted and milled surfaces. Models are selected based on the corresponding topography measurements $\mathcal{M}$, and used to simulate the meso scale of the object (texture). 

Measurements of the test body surfaces are provided as 2D height images $\mathcal{M}: \{1,\dots,M_\mathcal{M}\}\times\{1,\dots,N_\mathcal{M}\} \rightarrow\mathbb{R}$ of size $M_\mathcal{M}\times N_\mathcal{M}$ for $M_\mathcal{M},N_\mathcal{M}\in\mathbb{N}$ and pixel spacing $\nu_\mathcal{M}\in\mathbb{R}_{>0}$. 
The output of the models is a texture image $\mathcal{T}$ of arbitrary size $M_\mathcal{T}\times N_\mathcal{T}$ and pixel spacing $\nu_\mathcal{T}\geq\nu_\mathcal{M}$ coarser or equal to the input's pixel spacing. 
For the purpose of this work, we use $\nu_\mathcal{T}\approx6.1\,\mu$m and $M_\mathcal{T}=N_\mathcal{T}=13107$ so that a squared imaged region with edge length $80\text{mm}$ is provided.
The chosen pixel spacing is large enough to generate images of appropriate size for which the rendering process can still be done in acceptable time. 
On the other hand, pixel spacing is chosen small enough so that no visible discretization artifacts occur.
The resulting texture image can completely cover test body surfaces during the rendering process, even when the texture image is rotated.
This way, image tiling is avoided as it may produce visible edges between the tiles. 

\subsection{Related work}
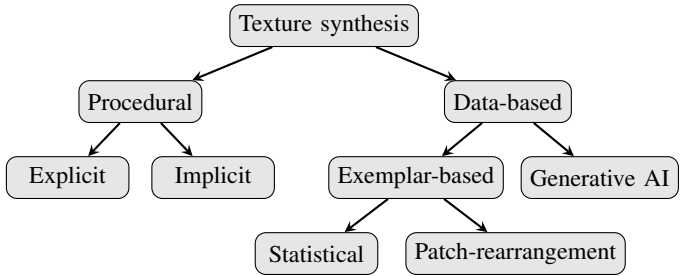
\begin{figure}
    \centering
    \begin{adjustbox}{width=\linewidth}
    \begin{tikzpicture}[node distance=10mm]
        \node (texture)    [boox] {\small Texture synthesis};
        
        \node (proc)       [boox, below of=texture, xshift=-2.4cm] {\small Procedural};
        \node (data)       [boox, below of=texture, xshift= 2.4cm] {\small Data-based};
        
        \node (ex)         [boox, below of=proc, xshift=-0.95cm] {\small Explicit};
        \node (im)         [boox, below of=proc, xshift= 0.95cm] {\small Implicit};
        \node (exemplar)   [boox, below of=data, xshift=-1.2cm] {\small Exemplar-based};
        \node (generative) [boox, below of=data, xshift= 1.2cm] {\small Generative AI};

        \node (stat)       [boox, below of=exemplar, xshift=-1.3cm] {\small Statistical};
        \node (patch)      
        [boox, below of=exemplar, xshift= 1.3cm] {\small Patch-rearrangement};

        \draw [thick,->,>=stealth] (texture) -- (proc);
        \draw [thick,->,>=stealth] (texture) -- (data);

        \draw [thick,->,>=stealth] (proc) -- (ex);
        \draw [thick,->,>=stealth] (proc) -- (im);
        \draw [thick,->,>=stealth] (data) -- (exemplar);
        \draw [thick,->,>=stealth] (data) -- (generative);
        
        \draw [thick,->,>=stealth] (exemplar) -- (stat);
        \draw [thick,->,>=stealth] (exemplar) -- (patch);
    \end{tikzpicture}
    \end{adjustbox}
    \caption{Classification of computer graphics methods for texture synthesis.}
    \label{fig:classification_textures}
\end{figure}

In computer graphics, the textures representing surface topographies can be roughly separated into procedural and data-based, see \cref{fig:classification_textures}.
In this work we used explicit procedural for the milling texture and exemplar-based approach for the sandblasted texture.

\textbf{Procedural texture modeling} is based on algorithms which completely describe the surface patterns without the need for additional data (e.g., image data).
The main concept is to combine regular patterns (e.g. sine functions, simple shapes such as lines and circles as well as other mathematical functions) discussed by Vivo et al. \cite{vivo2015book} with irregular patterns (e.g. noise functions) discussed by Dong et al. \cite{dong2020survey}.
Since procedural texture models are algorithmic descriptions of a texture, they offer parameterization and thus immense controllability which is recognized by the computer graphics community for content creation \cite{deguy2019new}.
However, developing procedural textures is often done by artists who base their work on experience and expression rather than encoding physically correct parameters.

Deguy \cite{deguy2019new} separates procedural models into implicit and explicit. 
An \textbf{explicit} procedural texture model generates a texture image which is first mapped onto the object surface and then used during rendering.
Contrary, an \textbf{implicit} procedural texture model is directly evaluated during the rendering procedure, thus generating texture in render-time and allowing render-time re-parametrization. 
While explicit procedural models are easier to develop for complex texture patterns, implicit procedural texture models can cover arbitrarily large surfaces without repetitions or seams and are not restricted by the resolution.
Bosnar et al. \cite{Bosnar2022TextureSynthesis} developed implicit procedural texture models for circular, parallel and radial textures.
Those patterns resemble processed surfaces but without taking an explicit machining method into account or basing them on real surface topography.

\textbf{Data-based texture modeling} relies on sensor data and measurements as input, as discussed by Tsirikoglou et al. \cite{Tsirikoglou2020Survey}.
For example, the input texture image may be obtained using photogrammetry \cite{dostal2018photogrammetric} or other scanning techniques \cite{ritz2019seamless}. 

The advantage of data-driven texture modeling is that it can synthesize approximations of real-world surfaces for which there are no well-defined mathematical descriptions.
However, those methods offer little control over the generated content since they depend entirely on the input data.
Data-based texture modeling can be roughly separated into generative AI and exemplar-based approaches.
\textbf{Generative AI texture synthesis} is often based on GANs as discussed by Jetchev et al. \cite{jetchev2016texture}, Bergmann et al. \cite{bergmann2017learning} and Zeltner et al. \cite{zeltner2023real}.
While the results of these models often have remarkable appearance to a person, it is not possible to ensure that the results are physically correct and is very difficult to introduce controllable parameters for different kinds of textures.
\textbf{Exemplar-based texture synthesis} uses algorithms to generate new texture images similar to a given input image. 
While stationary textures can be well reproduced, problems occur for textures with complex, large structures, e.g. milling rings.

Furthermore, since the output image relies entirely on the input image, it leaves no possibility of influencing the pattern itself.
Raad et al. \cite{raad2017OverviewExemplarBasedTextures} give an overview of common exemplar-based texture synthesis methods.
They distinguish between statistical and patch-rearrangement methods.

\textbf{Statistical} exemplar-based methods perform two steps.
First, model-dependent statistics of the input image are computed. Second, a randomly initialized image is adjusted such that it fits the observed statistics.
The asymptotic discrete spot noise (ADSN) and the random phase noise (RPN) are both parameter-free and non-iterative statistical methods \cite{galerne2011ADSN}.
A texture image simulated by the ADSN maintains the mean of the input image but has minor differences when comparing the sample auto-correlation since its Fourier modulus is the Fourier modulus of the input image multiplied by Rayleigh noise.
The RPN creates texture images that have the same Fourier modulus as the input image and thus the same auto-correlation but a random Fourier phase.
Heeger and Bergen \cite{HeegerBergen1995,Briand2014HeegerBergen} introduced a procedure for texture modeling that uses image pyramids.
A random image is adjusted iteratively by matching histograms of all images in its image pyramid to the corresponding images in the image pyramid of the reference image.
Portilla and Simoncelli \cite{PortillaSimoncelli2000,Vacher2021PortillaSimoncelli} improved this method by using statistics such as cross-correlations between pyramid levels, instead of histogram matching.
In contrast to the exemplar-based methods that generate completely new texture images, the \textbf{patch-rearrangement methods} quilt patches taken from the input image.
With the method of Efros and Freeman \cite{EfrosFreeman2001,Raad2017EfrosFreeman}, the output image grows successively by adding certain patches one after the other in a raster-scan order.
It is an extension of the method of Efros and Leung \cite{Efros1999EfrosLeung} that generates a new texture image pixel by pixel.

Guehl et al. \cite{guehl2020semi} introduce a hybrid approach combining procedural and data-based texture synthesis.
The visual structure of the generated image texture is based on a procedural parametric model, while texture details are synthesized using a data-driven approach. 
Another hybrid approach is presented by Hu et al. \cite{hu2022inverse}, in which the texture of the input image, which contains the BRDF parameters, is decomposed and proceduralized. 

\textbf{Modelling of metallic surfaces in computer graphics} has received a lot of attention.
Particularly interesting methods
are the ones describing highly specular surfaces with high-frequency details (glints) discussed by Zhu et al. \cite{zhu2022recent} and Chermain \cite{chermain2019glint}.
Glint effects appear due to complex and unstructured small-scale, high-frequency geometries such as tiny bumps, dents and scratches.
The problem with these approaches is that they do not model the actual properties of surfaces to resemble any particular and standardized surface. 
Rather, they are modeled to resemble the surface finishing appearance in an artistic way which is difficult to compare to any particular real-world counterpart.

Milling belongs to the machining processes resulting in deterministic and describable surface patterns \cite{groover2020fundamentals, childs2000metal}.
Therefore it is extremely useful to incorporate existing domain knowledge into texture models for milled surfaces. 
To estimate the surface quality of milled surfaces and thus determine the required quality, Felho et al. \cite{Felho2015SurfaceRoughnessFaceMilling,Felho2018FaceMillingRoughnessErrors} and Kundrak et al. \cite{Kundrak2022RoughnessFaceMilling} developed models for parallel face-milled surfaces.
Hadad et al. \cite{Hadad2016FaceMillingImages} developed a more detailed model that takes other typical milling parameters into account, e.g. tilting of the tool to avoid re-cutting.
All models create milled surfaces using CAD software and the patterns are thus geometrically imprinted into the surface. 
Oranli et al. \cite{2023OranliSandParticles} provide a similar approach to simulate the physical process of sandblasting and the resulting surface deformation in high precision using the Abaqus software.
Using such approach in for data synthesis would result in unfeasibly long computation times.
Furthermore, the surfaces generated in this way look too perfect, as possible irregularities, e.g. slight deviations in the cutting geometry due to irregular material and machine behavior, are not included in the models.

\subsection{Sandblasted surface}

The surface topography shows a different range of height values and degree of roughness depending on the pressure of the air-jet (\cref{fig:Overview_measurements:sand}).
The resulting surface topography is homogeneous, resembling a stationary Gaussian random field.
However, fitting parametric Gaussian random fields did not provide satisfactory texture images since the structure sizes were not reproduced correctly.
In addition, due to the small number of process parameters, it is difficult to develop a model that depends exclusively on them. 
Therefore, we used a combination of exemplar-based texture synthesis methods that only receive the measurement as input.
We select measurements with a size between $4430\times3248$ and $4440\times 3288$ and pixel spacing $\nu_\mathcal{M}\approx1.75\,\mu$m that correspond to an area of about $7.8\text{ mm}\times5.8\text{ mm}$.

To generate texture images for sandblasted surfaces, the asymptotic discrete spot noise (ADSN) introduced by Galerne at al. \cite{galerne2011ADSN} is used.
The size and pixel spacing of the input image are maintained.
To obtain a new texture image by simulating the ADSN, sample from the multivariate normal distribution $\mathcal{N}(\hat{\mu}_\mathcal{M},\widehat{C}_\mathcal{M})$ with $\hat{\mu}_\mathcal{M}$ the sample mean and $\widehat{C}_\mathcal{M}$ the sample covariance matrix of the input image.

\begin{figure}
    \centering
    \begin{tabular}{rcl}
        \includegraphics[height=2cm]{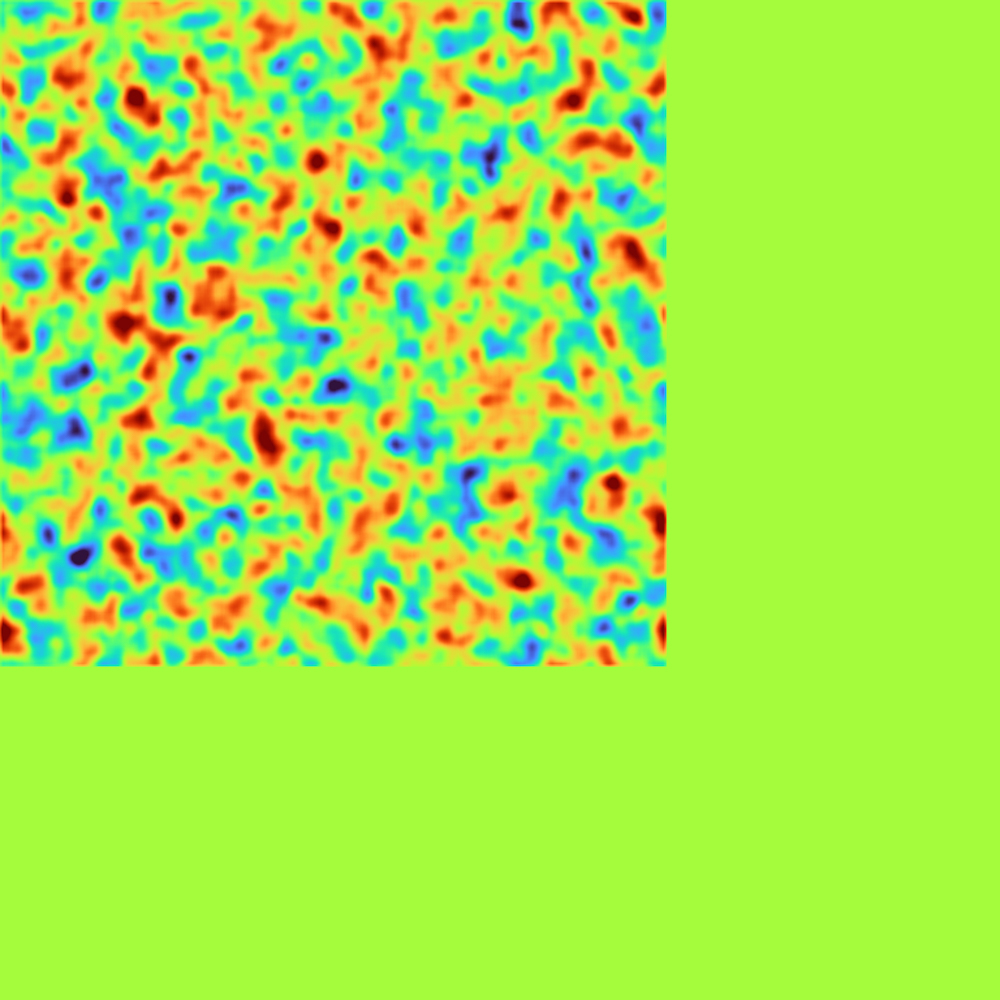} & \includegraphics[height=2cm]{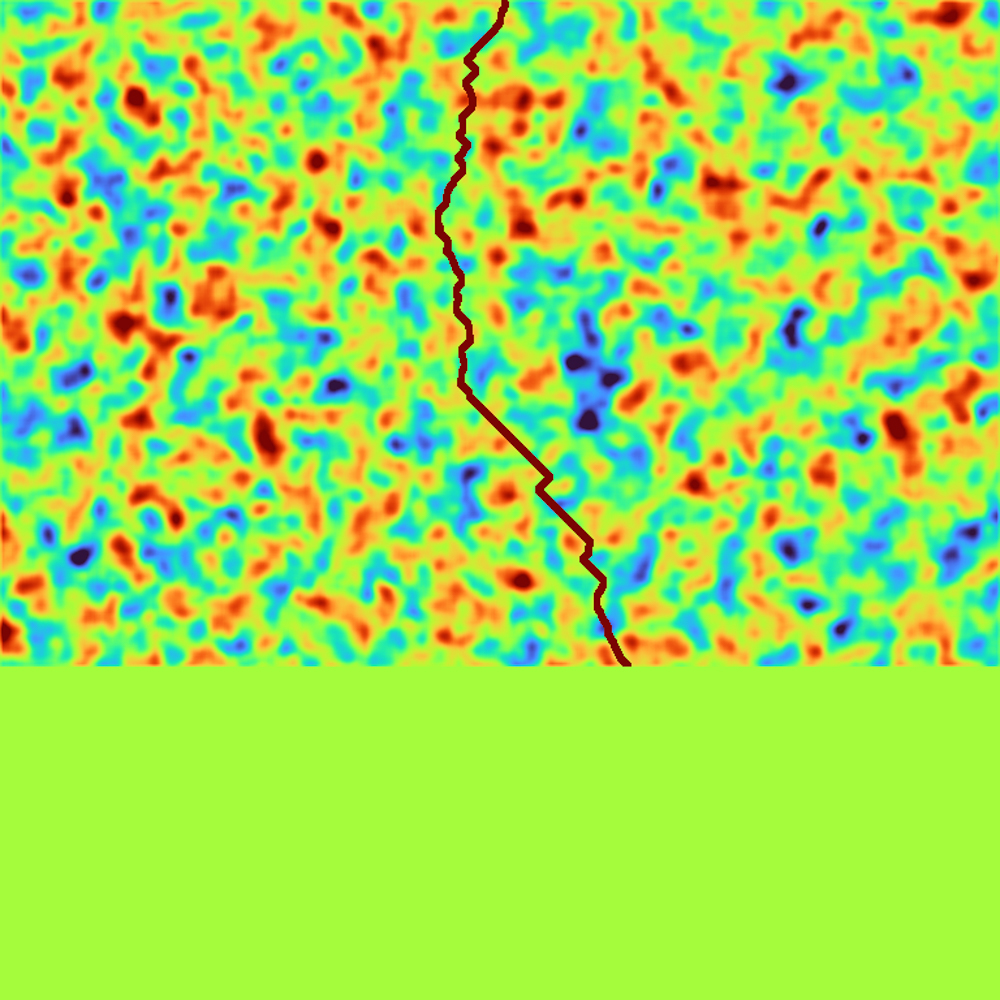} &  \\
        \includegraphics[height=2cm]{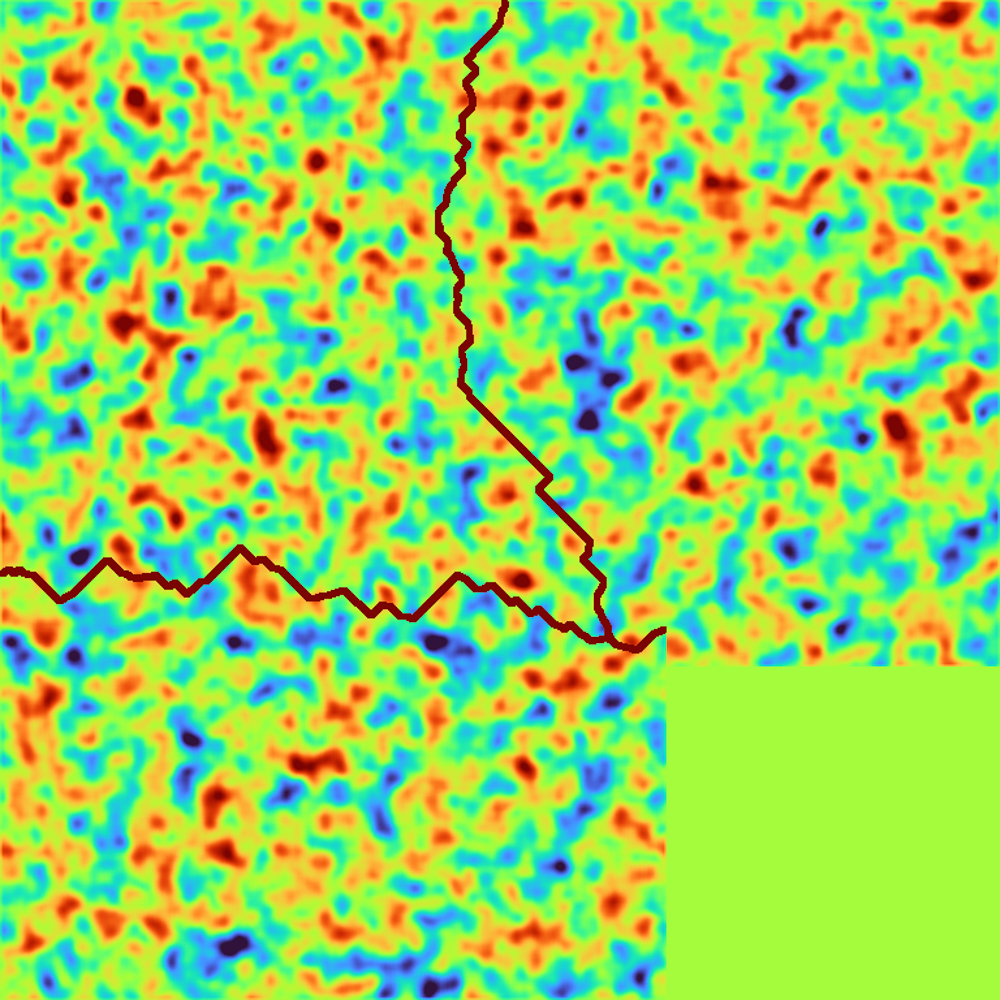} & \includegraphics[height=2cm]{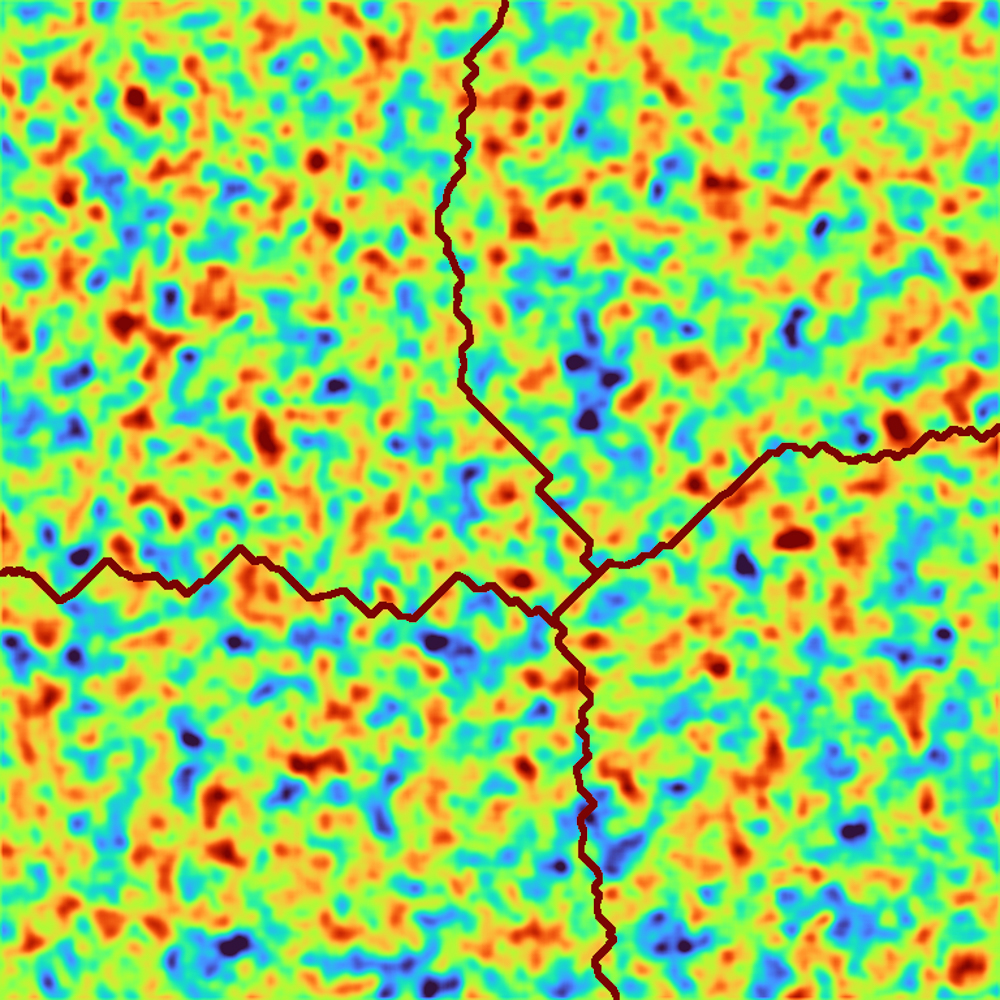} & \includegraphics[height=2cm]{images/Fig9bc.png}
    \end{tabular}
    \caption{Illustration of EF-stitching marking the minimal path in each step. Patches have the size of $512\times 512$ and the overlap region width is $256$ pixels. The imaged region is approximately $1.34\text{ mm}\times 1.34\text{ mm}$.}
    \label{Fig:EFstitching}
\end{figure}

Next, the desired size $M_\mathcal{T}\times N_\mathcal{T}$ and pixel spacing $\nu_\mathcal{T}$ of the texture image have to be met.
The pixel spacing can be adjusted by downsampling the input image using nearest neighbor interpolation.
ADSN can generate texture images larger than the input image, while maintaining the same pixel spacing \cite{galerne2011ADSN}.
Therefore, the normalized input image 
$\sqrt{
    \nicefrac{
        M_\mathcal{T}N_\mathcal{T}}{M_\mathcal{M}N_\mathcal{M}}}\left(\mathcal{M}-\hat{\mu}_\mathcal{M}\right)+\hat{\mu}_\mathcal{M}$ is enlarged using padding with mean value $\hat{\mu}_\mathcal{M}$ before applying the ADSN.
However, periodical repetitions are visible in the resulting texture. 
To avoid this, multiple texture image patches are generated and stitched using EF-stitching \cite{EfrosFreeman2001,Raad2017EfrosFreeman}.
To stitch two neighboring patches within an overlap region, both are cut apart along an intersection edge first and put together afterwards.
The intersection edge is determined by finding the path within the overlap region where both images are most similar.
Depending on the alignment of two neighboring patches we distinguish between vertical, horizontal and L-shaped stitching (\cref{Fig:EFstitching}), see \cite{jeziorski2024stochastic} for details.

\subsection{Milled surface}
\begin{figure}
	\centering
    \begin{subfigure}[b]{0.4\columnwidth}
        \centering
        \includegraphics[width=\columnwidth]{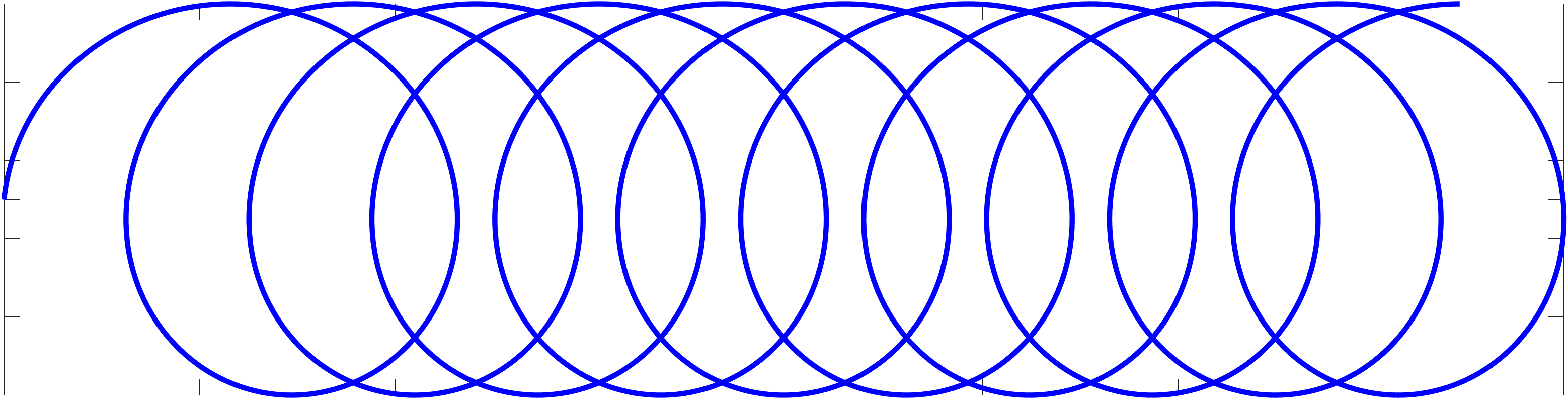}\\[4pt]
        \includegraphics[width=\columnwidth]{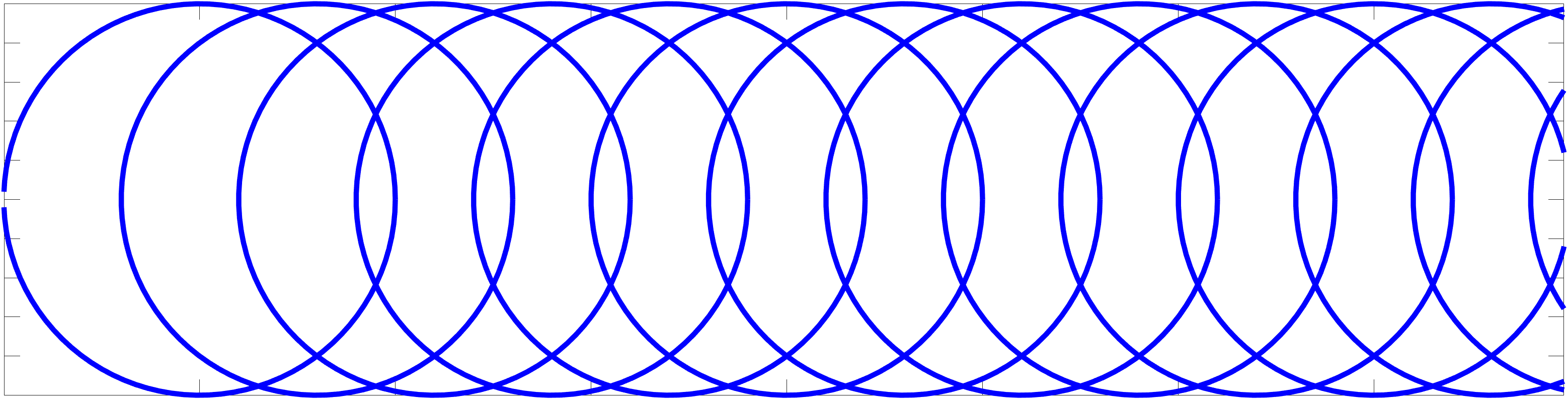}
        \caption{Edge path (top) and its approximation (bottom).}
        \label{Fig:Sub:Helix}
    \end{subfigure}
    \begin{subfigure}[b]{0.54\columnwidth}
        \centering
        \includegraphics[width=0.43\columnwidth]{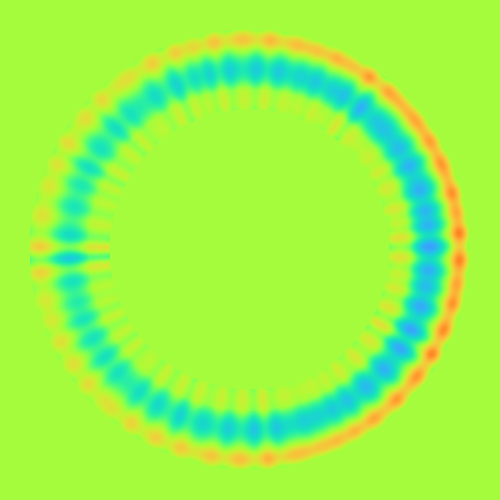}
        \includegraphics[width=0.43\columnwidth]{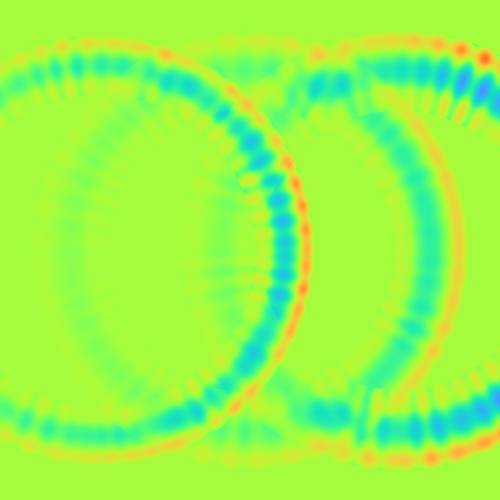}
        \includegraphics[width=0.073\columnwidth]{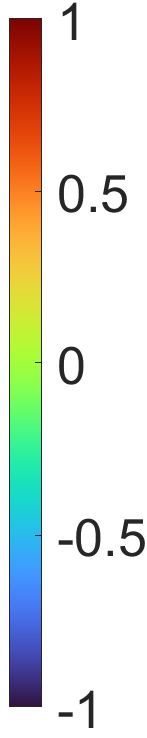}
        \caption{Sub-models for one ring (left) and interaction of multiple rings (right).}
        \label{Fig:Sub:ModelMilling}
    \end{subfigure}
    \caption{Overview of model generating milled surfaces.}
    \label{Fig:ModelMilling}
\end{figure}

The model generating textures of milled surfaces is given as a function in a continuous domain $\mathbb{R}^2$.
Thus, it is procedural but implemented explicitly for the sake of simplicity.
The discrete texture image is obtained by evaluating the function at the image grid-points $\left(\nu_\mathcal{T} x,\nu_\mathcal{T} y\right)$ for $x=1,\dots,M_\mathcal{T}$ and $y=1,\dots,N_\mathcal{T}$.
The milling pattern appearance depends on a number of parameters associated with production parameters.
The most prominent structures are cycloidal edge paths caused by the rotation of the milling head.
They are approximated by $n\in\mathbb{N}$ rings $R_1,\dots,R_n$ (\cref{Fig:Sub:Helix}).
The model is further divided into the following sub-models for the
\begin{enumerate}
    \item tool path providing the arrangement and order of the rings by defining their center points $c_k$
    \item appearance of an individual ring $R_k$ (\cref{Fig:Sub:ModelMilling})
    \item interaction between neighboring rings (\cref{Fig:Sub:ModelMilling})
\end{enumerate}
for $k=1,\dots,n$.

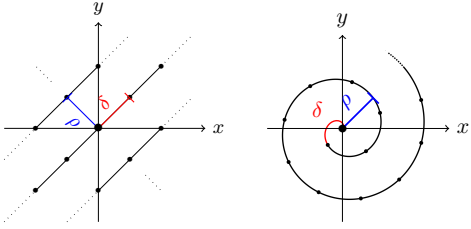
\begin{figure}
    \centering
    \begin{subfigure}[b]{0.35\columnwidth}
    \centering
    \resizebox{\columnwidth}{!}{
    \begin{tikzpicture}
        \draw[->,line width = 0.3pt] (-1.5,0) -- (1.7,0) node[right] {\small $x$};
    	\draw[->,line width = 0.3pt] (0,-1.5) -- (0,1.7) node[above] {\small $y$};
    		
    	\draw[dotted,line width = 0.3pt] (-0.5,-1.5) -- ( 0,-1);
    	\draw[dotted,line width = 0.3pt] (-1.5,-1.5) -- (-1,-1);
    	\draw[dotted,line width = 0.3pt] (-1.5,-0.5) -- (-1, 0);
    	\draw[line width = 0.3pt] ( 0,-1) -- (1,0);
    	\draw[line width = 0.3pt] (-1,-1) -- (1,1);
        \draw[line width = 0.3pt] (-1, 0) -- (0,1);
    	\draw[dotted,line width = 0.3pt] (1,0) -- (1.5,0.5);
    	\draw[dotted,line width = 0.3pt] (1,1) -- (1.5,1.5);
    	\draw[dotted,line width = 0.3pt] (0,1) -- (0.5,1.5);
    		
    	\draw[dotted,line width = 0.3pt] ( 0.75,-0.75) -- ( 1,-1);
    	\draw[dotted,line width = 0.3pt] (-0.75, 0.75) -- (-1, 1);
    		
    	\node at (-1  , 0  ) {\tiny $\bullet$};
    	\node at (-0.5, 0.5) {\tiny $\bullet$};
    	\node at ( 0  , 1  ) {\tiny $\bullet$};
    	\node at (-1  ,-1  ) {\tiny $\bullet$};
    	\node at (-0.5,-0.5) {\tiny $\bullet$};
    	\node at ( 0  , 0  ) {\tiny $\bullet$};
    	\node at ( 0.5, 0.5) {\tiny $\bullet$};
    	\node at ( 1  , 1  ) {\tiny $\bullet$};
    	\node at ( 0   ,-1 ) {\tiny $\bullet$};
    	\node at ( 0.5,-0.5) {\tiny $\bullet$};
    	\node at ( 1  , 0  ) {\tiny $\bullet$};
    		
    	\draw[-|,color=blue, line width = 0.5pt] (0,0) -- (-0.5,0.5) node[midway,below,sloped] {\small $\rho$};
    	\draw[-|,color=red,  line width = 0.5pt] (0,0) -- ( 0.5,0.5) node[midway,above,sloped] {\small $\delta$};
    	\node at (0, 0) {$\bullet$};	
    \end{tikzpicture}}
    \end{subfigure}
    \begin{subfigure}[b]{0.35\columnwidth}
    \centering
    \resizebox{\columnwidth}{!}{
        \begin{tikzpicture}
        	\node at (0,0) {\includegraphics[height=2.75cm]{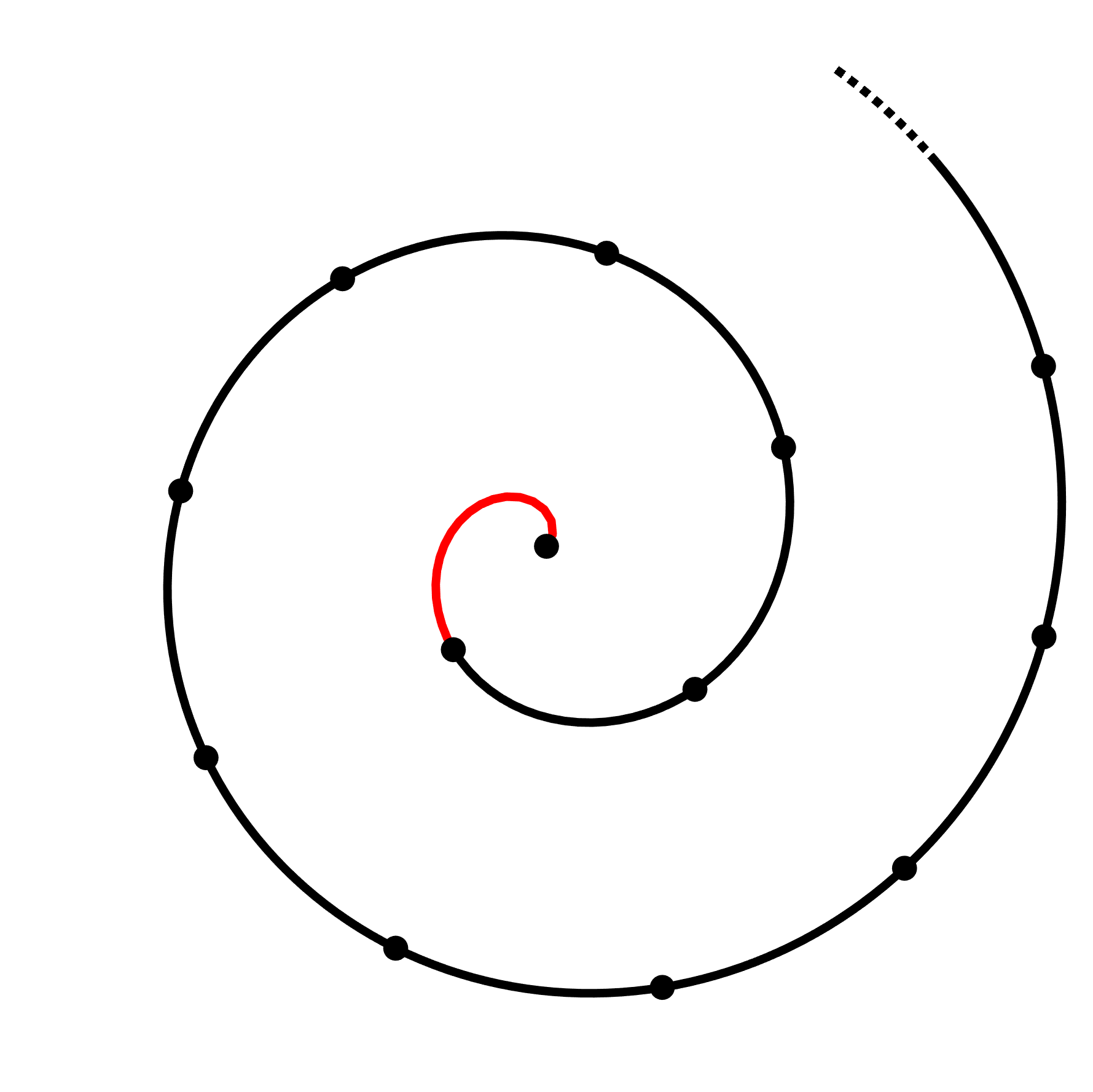}};
        	\draw[->,line width = 0.3pt] (-1.2,0) -- (1.7,0) node[right] {$x$};
        	\draw[->,line width = 0.3pt] (0,-1.5) -- (0,1.5) node[above] {$y$};
            \draw[dotted,line width = 0.5pt] (0.686,1.25) -- (0.685,1.251);
        	\draw[-|,line width=0.8pt,color=blue] (0,0) -- (0.5,0.5) node[midway,above,sloped] {$\rho$};	
        	\node at (0.005,-0.005) {$\bullet$};
        	\node at (-0.4,0.3) {\textcolor{red}{$\delta$}};
        \end{tikzpicture}}
    \end{subfigure}
    \caption{Tool paths of parallel (left) and spiral milling (right) with center points.}
    \label{Fig:Pattern}
\end{figure}

The first sub-model generates the ring center points $c_k\in\mathbb{R}^2$ along the tool path.
Within this study, either parallel or spiral milling (\cref{Fig:Pattern}) is considered.
First, the respective tool path has to be modeled, together with the distance $\rho\in(0,d)$ between two neighboring sub paths.
The distance is controlled by the amount of ring overlap $\alpha\in(0,1)$ and the diameter $d\in\mathbb{R}_{>0}$ of the rings, as $\rho=(1-\alpha)\cdot d$.
The diameter of the milling head is crucial for defining the ring diameter and the ring overlap depends on the radial cutting depth $a_e$ by $\alpha=1-a_e$.
The radial cutting depth defines the distance the tool steps over in relation to the ring diameter.
As the blades do not reach the edge of the tool, $\alpha$ is reduced here by taking $\alpha-\gamma$ with $\gamma\in[0,\alpha)$.
The ring center points $c_k$ are then modeled by random displacements of points $\tilde{c}_k$ which are computed along the tool path. That is, we sample from a two-dimensional normal distribution $c_k\sim\mathcal{N}\left(\tilde{c}_k,\Sigma_c\right)$.
The location of $\tilde{c}_k$ is specified by the distance $\delta\in\mathbb{R}_{>0}$ between two center points along the tool path.
The distance $\delta$ depends on the tool's feed rate and rotational speed.
It is important to keep the chronological order of the rings in order to keep the tool path visible.
Finally, while being very precise, the milling process is not perfect and every once in a while there is an occurrence of a ring which is more visible than the others.
To introduce such irregularities in visibility of rings, the order of a certain amount $\epsilon\in[0,1]$ of rings is changed.

\begin{figure}
    \centering
    \begin{subfigure}[b]{\columnwidth}
        \centering
        \includegraphics[width=0.22\columnwidth]{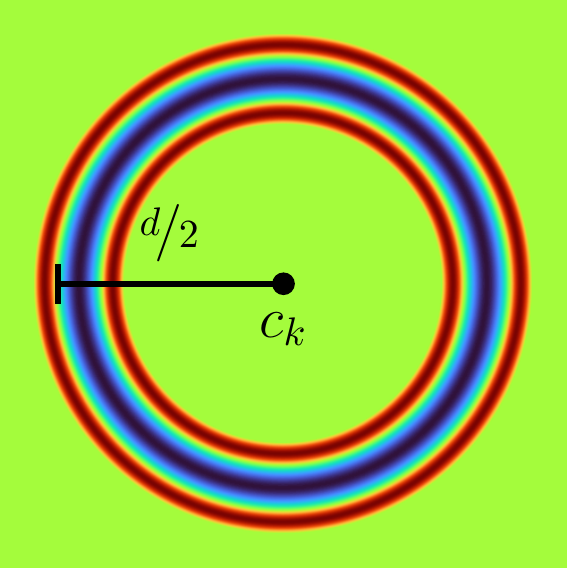}
        \includegraphics[width=0.22\columnwidth]{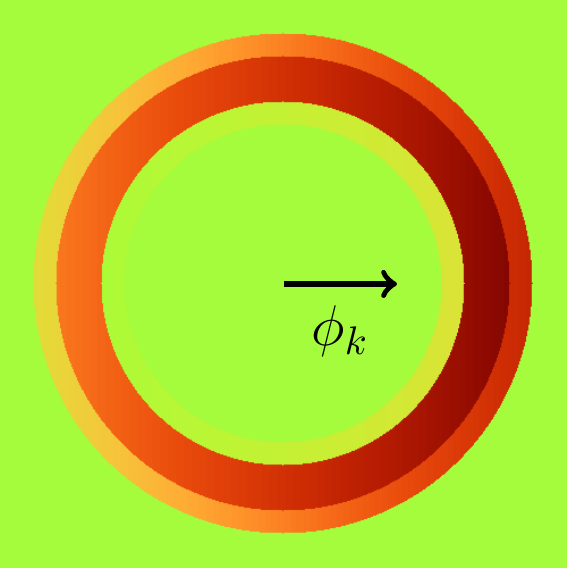}
        \includegraphics[width=0.22\columnwidth]{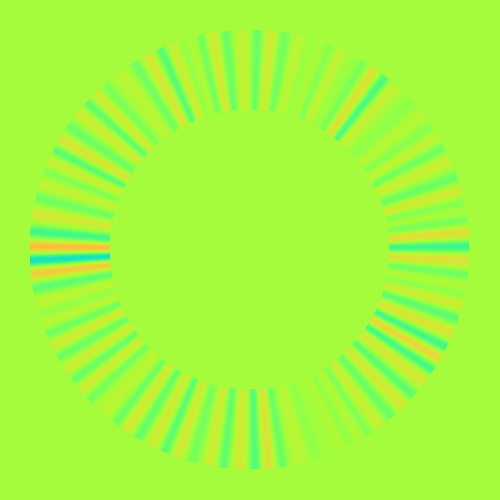}
        \includegraphics[width=0.22\columnwidth]{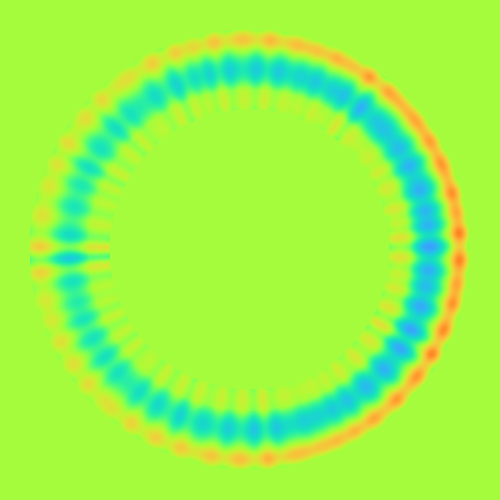}
        \includegraphics[width=0.04\columnwidth]{images/Fig14bc.png}
        \caption{Sub-model for ring appearance using $\phi_k=0$, from left to right: $S_k$, $T_k$, $N_k$, $R_k$.}
        \label{Fig:Sub:Appreance}
    \end{subfigure}
    
    \begin{subfigure}[b]{\columnwidth}
        \centering
        \includegraphics[width=\columnwidth]{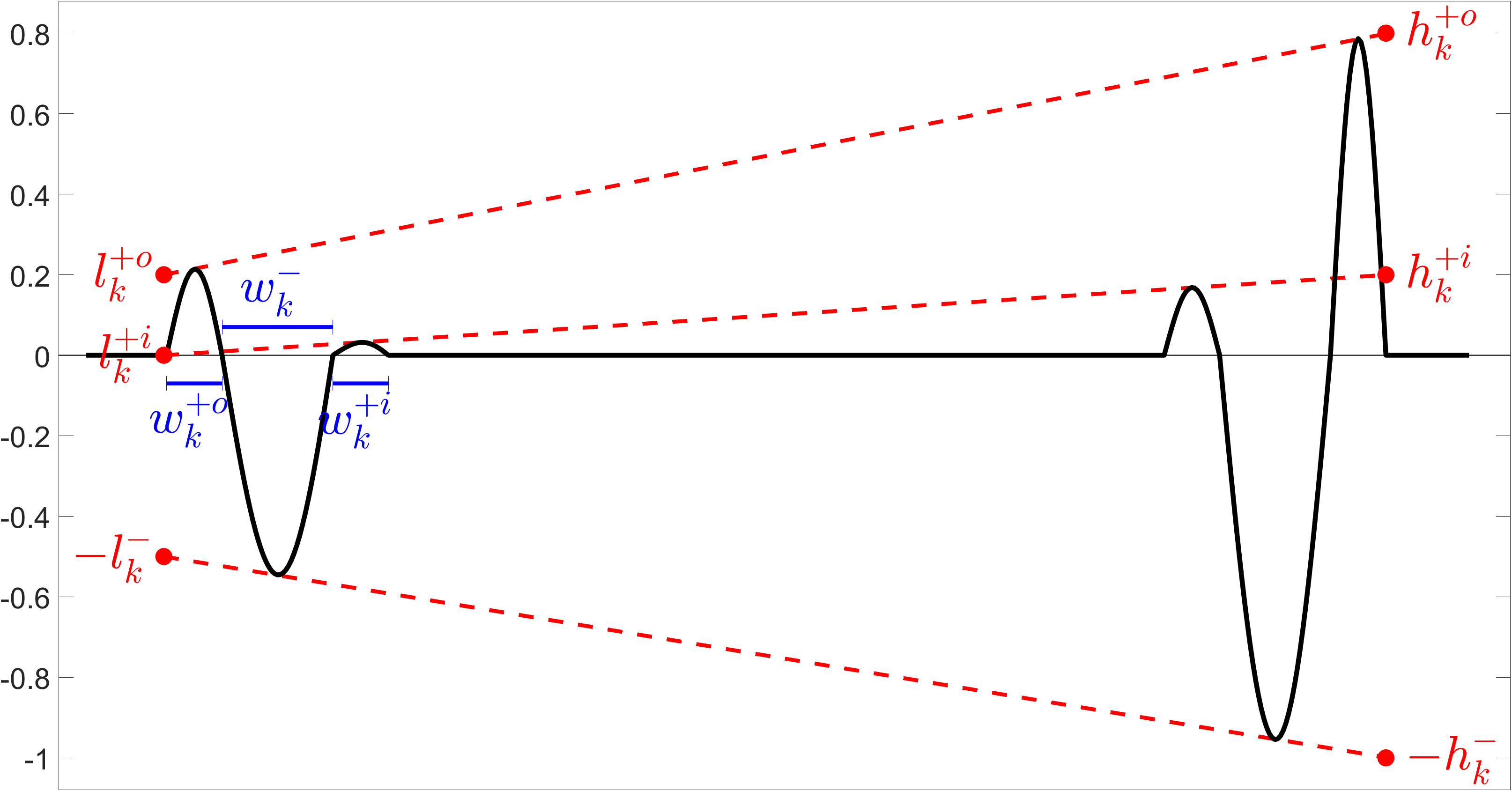}
        \caption{1d intersection of $S_k\cdot T_k$.}
        \label{Fig:Sub:Shape}
    \end{subfigure}
    \caption{Illustration of sub-model for ring appearance with explanation of its parameters.}
    \label{Fig:Appearance}
\end{figure}

The second sub-model controls the appearance of an individual ring $R_k$, produced by a single rotation of the milling head cutting edge.
Therefore, the width of the cutting edge is crucial for the indentation width $w^-_k\sim\mathcal{N}\left(\mu_{w^-},\sigma_{w^-}\right)$.
To get height values, the shape of the indentation is modeled by the cosine function $-\cos(x)$ for $x\in[-\nicefrac{\pi}{2},\nicefrac{\pi}{2}]$ scaled to the width of the ring.
Material depositions are modeled by accumulations beside the indentation in form of rings having positive values.
The width for the inner and outer rings are $w^{+i}_k\sim\mathcal{N}\left(\mu_{w^{+i}},\sigma_{w^{+i}}\right)$ and $w^{+o}_k\sim\mathcal{N}\left(\mu_{w^{+o}},\sigma_{w^{+o}}\right)$ and the height component is again modeled using the cosine function.
All functions are combined to form the shape $S_k$ of ring $R_k$ (\cref{Fig:Appearance}).

So far, perpendicularity between the milling head and the surface has been assumed. 
However, the milling head is often tilted forwards in direction of the tool movement to prevent re-cutting the surface.
This direction is given by the angle $\phi_k\in(-\pi,\pi]$ in the $x$-$y$-plane between the tool path and the $x$-axis at point $\tilde{c}_k$ 
Tilting introduces a slope within the rings defined by the minimal $l^\bullet_k\sim\mathcal{N}\left(\mu_{l^\bullet},\sigma_{l^\bullet}\right)$ and maximal $h^\bullet_k\sim\mathcal{N}\left(\mu_{h^\bullet},\sigma_{h^\bullet}\right)$ indentation, inner and outer accumulation at the ring's front respectively back for $\bullet\in\left\{-,+i,+o\right\}$ (\cref{Fig:Appearance}). 
Since the tool's presence prevents pushing material to the inside, $l^{+i}$ and $h^{+i}$ are chosen small.
Tilting $T_k$ is then applied by multiplying with the shape $S_k$.

Finally, noise $N_k$ is added, simulating irregularities caused by resistances in the material or vibrations during the process.
It is modeled by a combination of $\lambda_k\sim\mathcal{P}(\lambda)$ sine curves with varying frequencies $\tau_{k_j}\sim\mathcal{P}(\tau)$ and random shifts $\xi_{k_j}\sim\mathcal{U}\left(-\pi,\pi\right)$ for $j=1,\dots,\lambda_k$. Here, $\mathcal{P}$ denotes a Poisson and $\mathcal{U}$ a uniform distribution.
Thus, $R_k=S_k\cdot T_k+N_k$ defines the sub-model for the appearance of one individual ring (\cref{Fig:Appearance}).

\begin{figure}
    \centering
    \includegraphics[width=0.22\columnwidth]{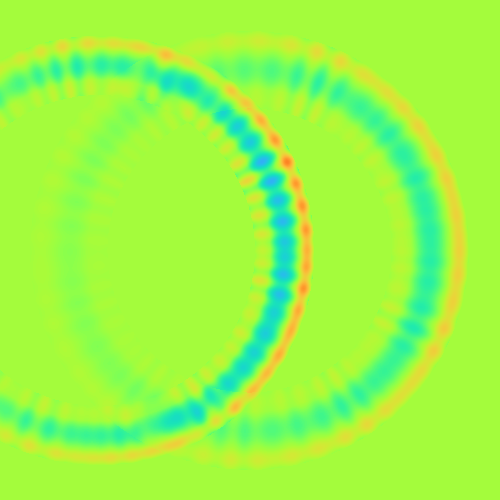}
    \includegraphics[width=0.22\columnwidth]{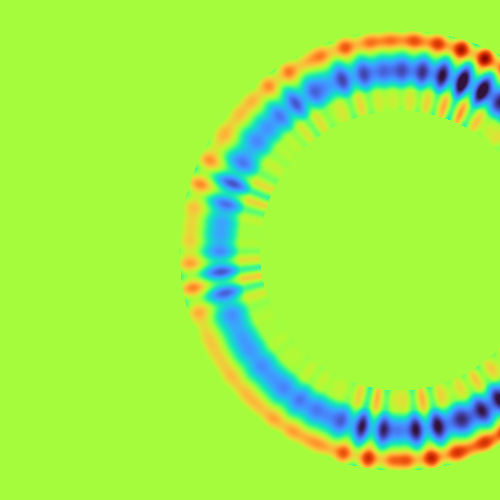}
    \includegraphics[width=0.22\columnwidth]{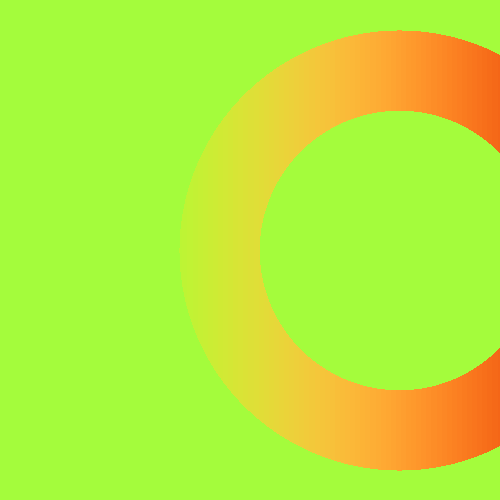}
    \includegraphics[width=0.22\columnwidth]{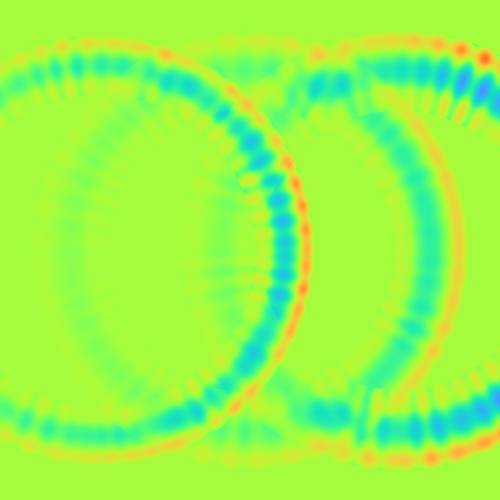}
    \includegraphics[width=0.04\columnwidth]{images/Fig14bc.png}
    \caption{Sub-model for rings' interaction using $\phi_k=0$ (no tilting), from left to right: $f(R_1,R_2)$, $R_3$, $L_3$, $f(R_1,R_2,R_3)$.}
    \label{Fig:Interaction}
\end{figure}

The third sub-model deals with the interaction of the neighboring rings.
First, the rings are successively added into the model, according to their order by using a recursive function $f\left(R_1,\dots,R_{k}\right)=L_k\cdot R_k + (1-L_k)\cdot f\left(R_1,\dots,R_{k-1}\right)$ for $k=1,\dots,n$ and $R_0=0$ (\cref{Fig:Interaction}).
Since every part of the surface is processed several times, all associated rings must be taken into account with most of the weight on the temporally last ones.
This is achieved by a convex combination.
Weighting parameters are chosen as linear function $L_k:\mathbb{R}^2\rightarrow[0,1]$ which gives more weight to the front part of the ring than to the back. 
It is zero for points outside the ring.
Values at the front and back of the rings are given by $a_k\sim\mathcal{U}(a^{\text{min}},a^{\text{max}})$ and $b_k\sim\mathcal{U}(b^{\text{min}},b^{\text{max}})$.

\begin{figure}
    \centering
    \begin{subfigure}[t]{0.44\columnwidth}
        \centering
        \includegraphics[width=\columnwidth]{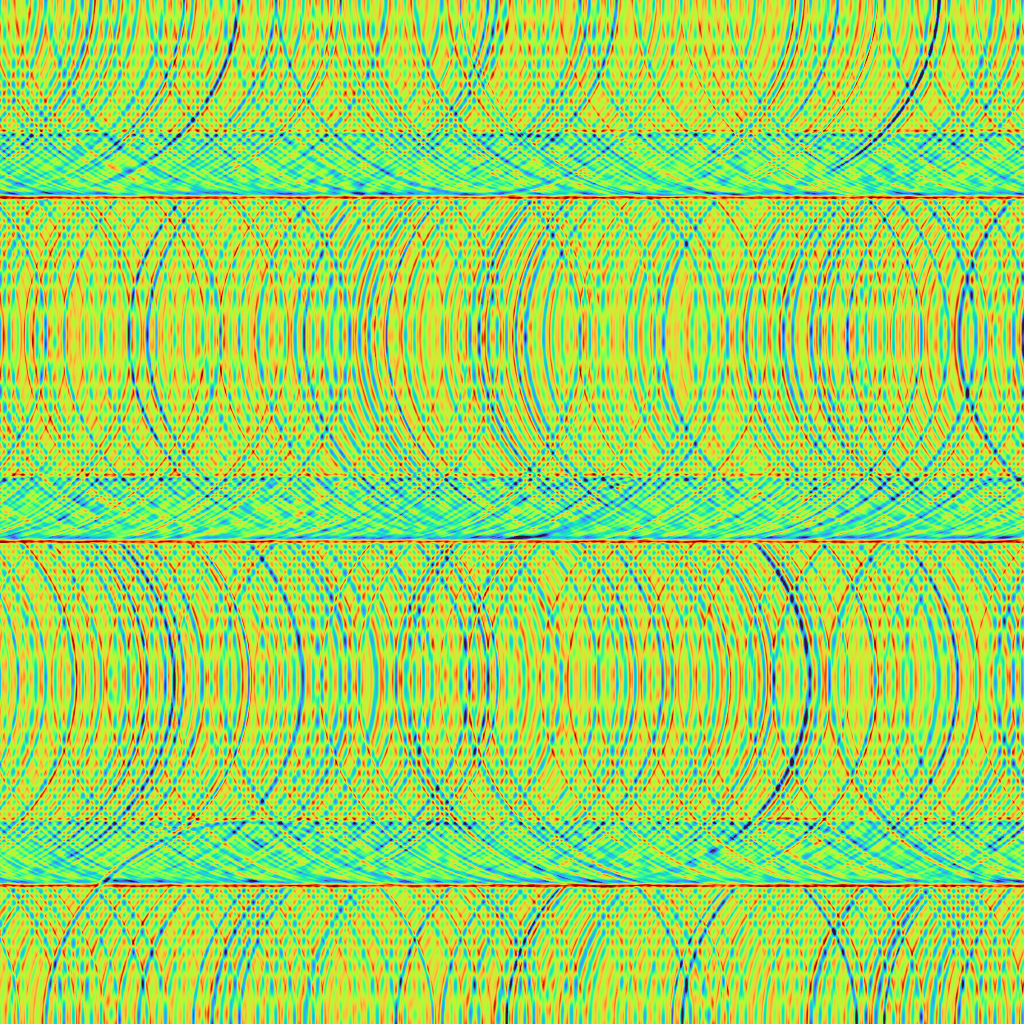}
        \caption*{Parallel milling using $d=4$mm, $\alpha=0.2$, $\delta=0.09$mm.}
    \end{subfigure}
    \begin{subfigure}[t]{0.44\columnwidth}
        \centering
        \includegraphics[width=\columnwidth]{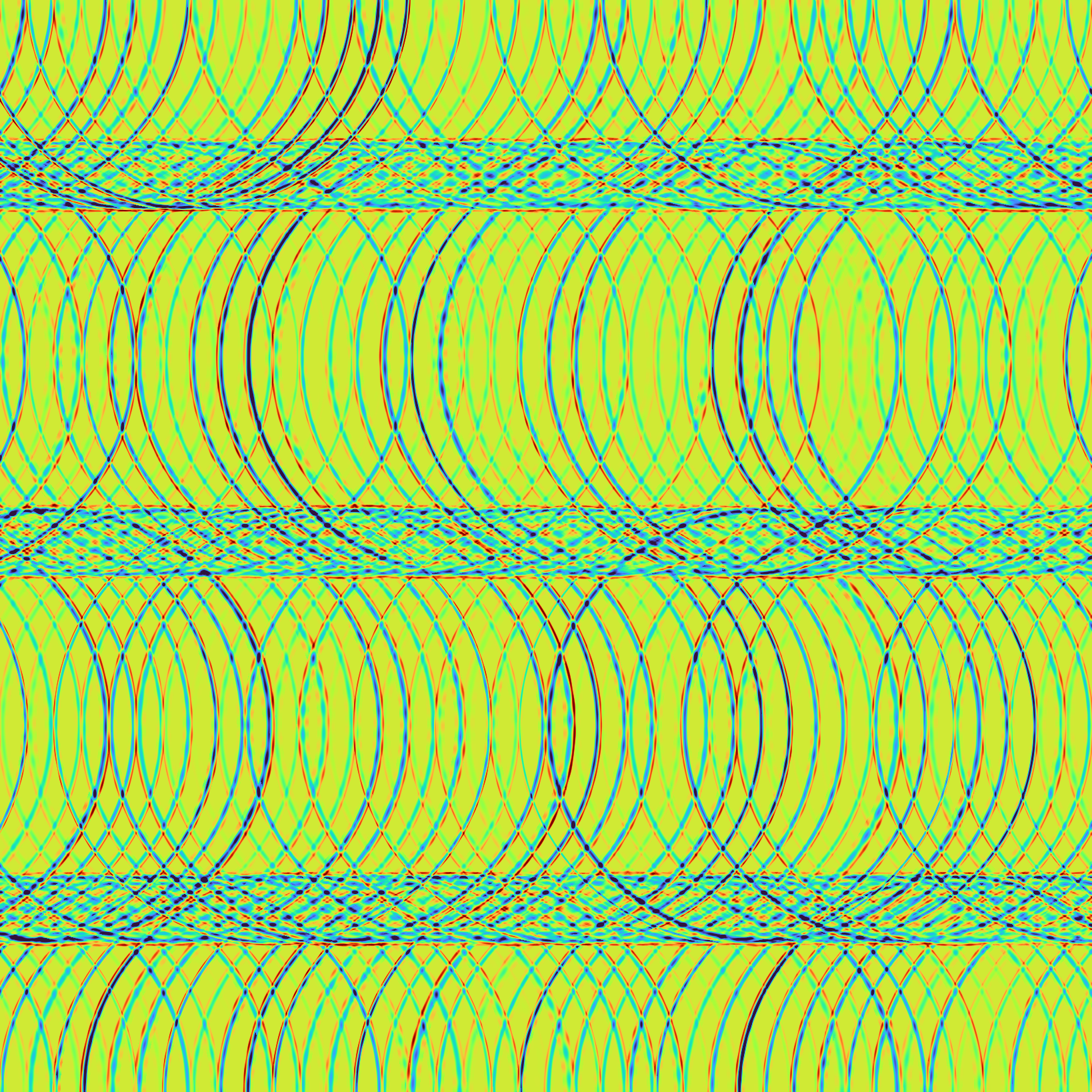}
        \caption*{Parallel milling using $d=4$mm, $\alpha=0.2$, $\delta=0.25$mm.}
    \end{subfigure}
    \phantom{\includegraphics[width=0.07\columnwidth]{images/Fig9ac.png}}\\
    \begin{subfigure}[t]{0.44\columnwidth}
        \centering
        \includegraphics[width=\columnwidth]{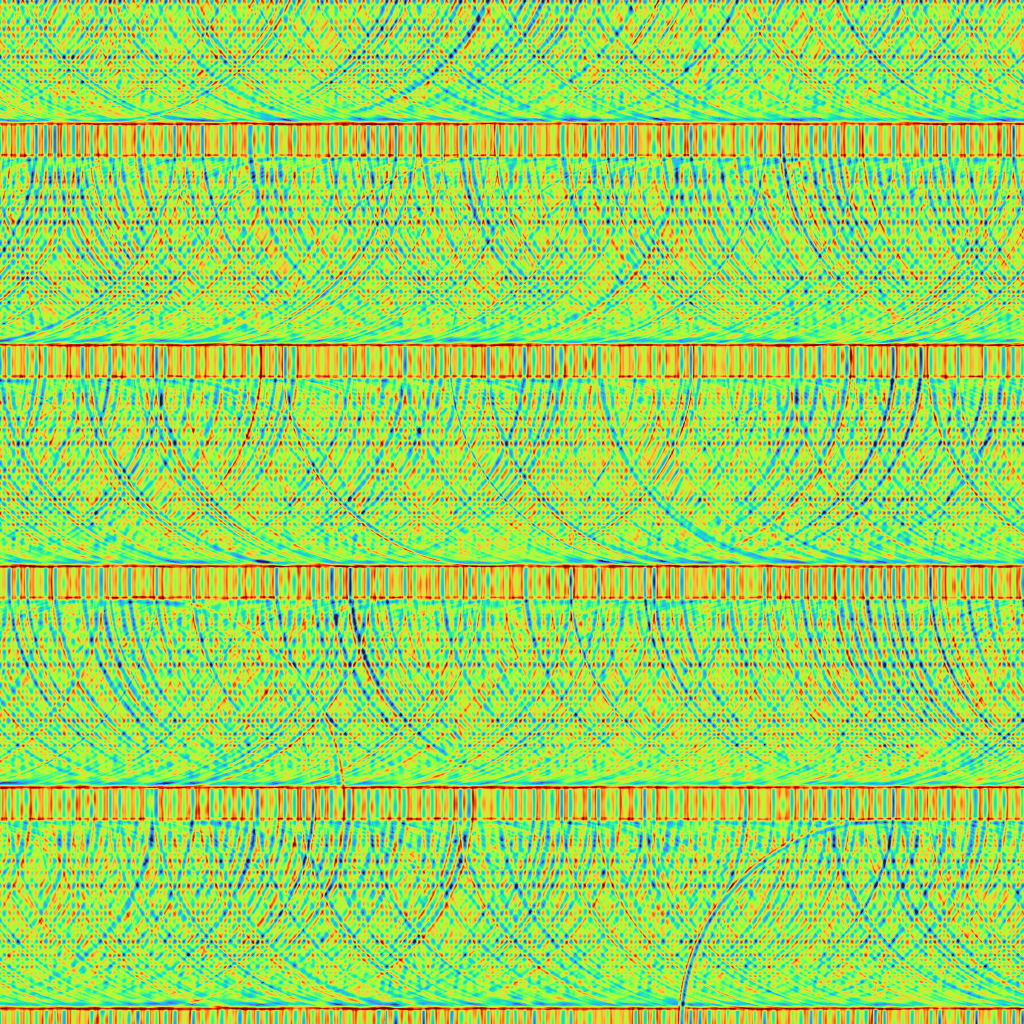}
        \caption*{Parallel milling using $d=4$mm, $\alpha=0.5$, $\delta=0.09$mm.}
    \end{subfigure}
    \begin{subfigure}[t]{0.44\columnwidth}
        \centering
        \includegraphics[width=\columnwidth]{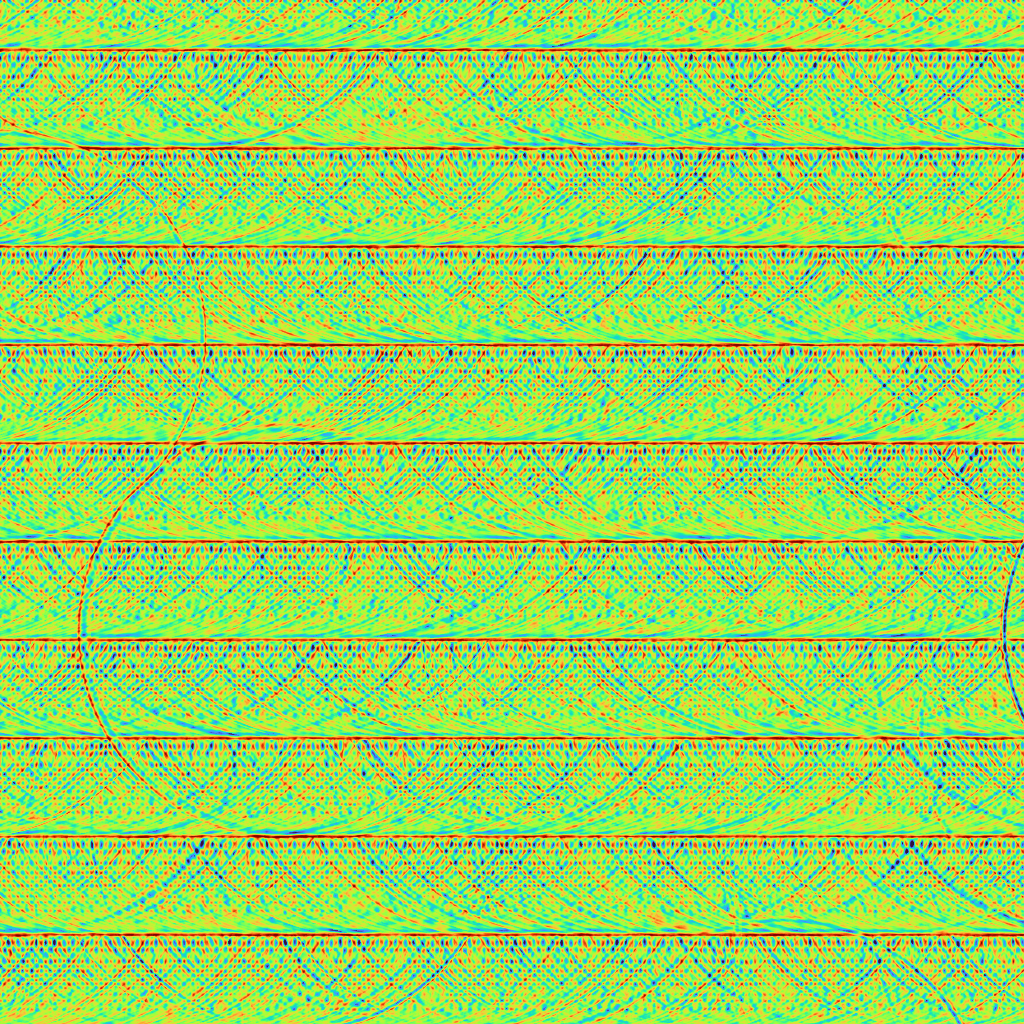}
        \caption*{Parallel milling using $d=4$mm, $\alpha=0.8$, $\delta=0.09$mm.}
    \end{subfigure}
    \phantom{\includegraphics[width=0.07\columnwidth]{images/Fig9ac}}\\
    \begin{subfigure}[t]{0.44\columnwidth}
        \centering
        \includegraphics[width=\columnwidth]{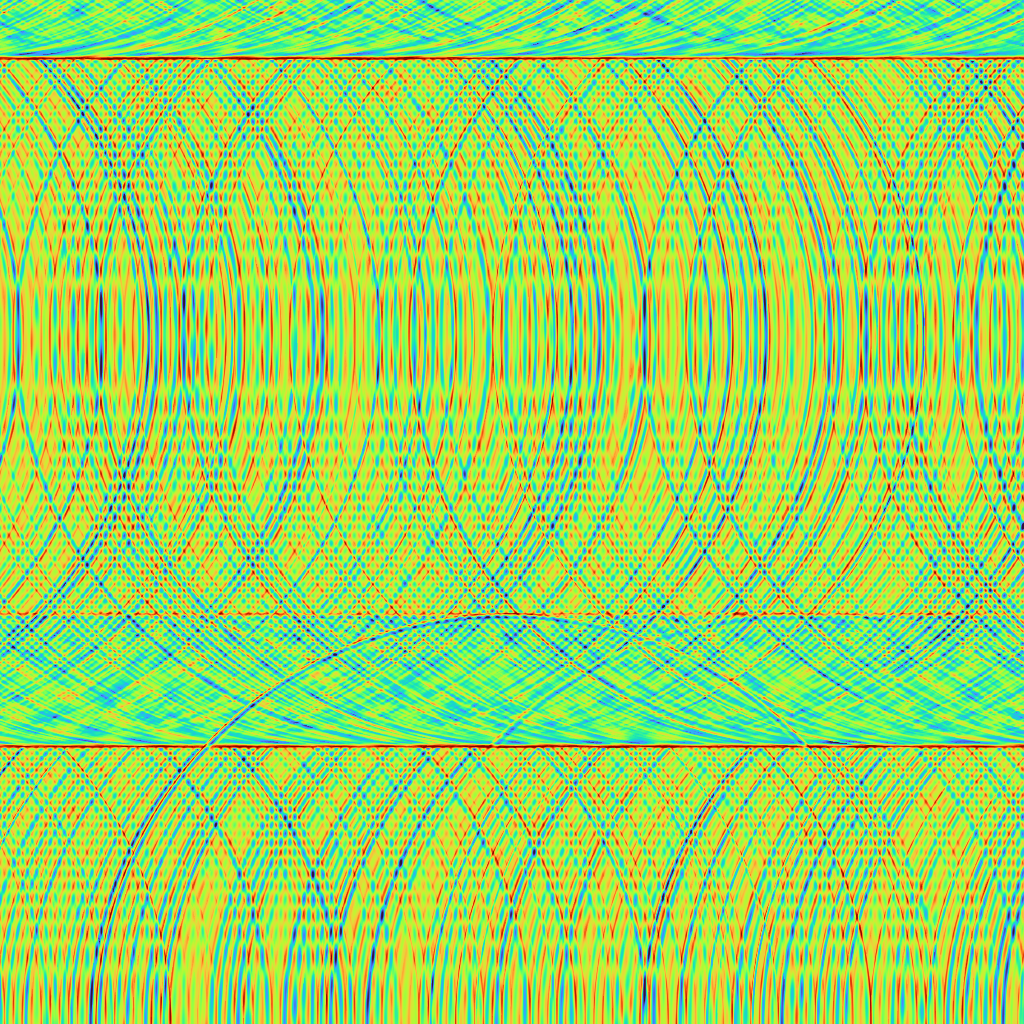}
        \caption*{Parallel milling using $d=8$mm, $\alpha=0.2$, $\delta=0.09$mm.}
    \end{subfigure}
    \begin{subfigure}[t]{0.44\columnwidth}
        \centering
        \includegraphics[width=\columnwidth]{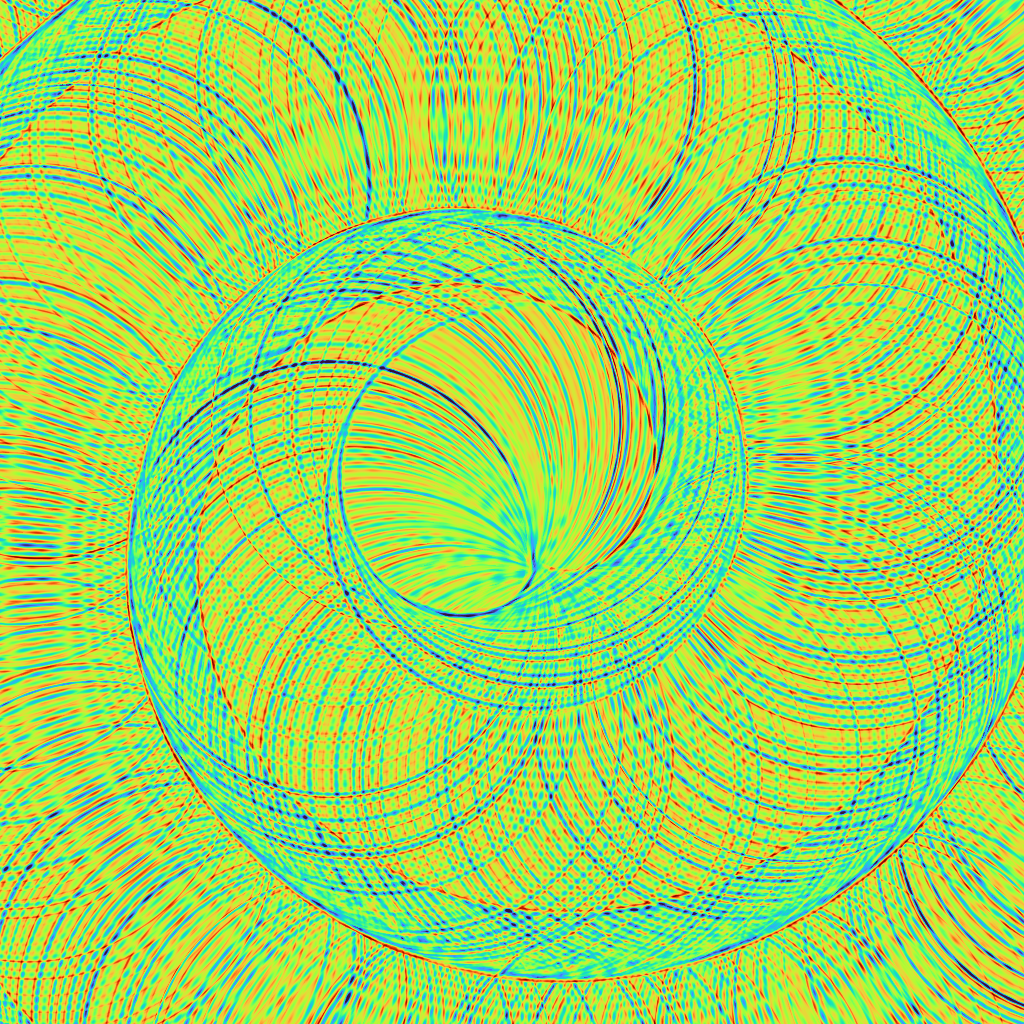}
        \caption*{Spiral milling using $d=4$mm, $\alpha=0.2$, $\delta=0.09$mm.}
    \end{subfigure}
    \includegraphics[width=0.07\columnwidth]{images/Fig9ac.png}
    \caption{Adapted simulated milled surfaces generated with different parameter configurations. Imaged region is $10\text{mm}\times10\text{mm}$.}
    \label{Fig:Milling_differences}
\end{figure}

We refer to \cite{jeziorski2024stochastic} for more details on the sub-models.
A summary of the model parameters is given in \cref{Tab:Mill_parameter}.
Some parameters are known explicitly (\cref{tab:processing_params}), while the others are estimated by visual comparison using high resolution topography measurements (\cref{fig:Overview_measurements:mill}).
Simulations of the model using distinct parameter configurations are shown in \cref{Fig:Milling_differences}.
Height values of the preliminary texture image $\mathcal{T}_\text{pre}$ are adapted so that their distribution resembles that of the corresponding measurement $\mathcal{M}$.
Mean values $\hat{\mu}_\bullet$ and sample variances $\hat{\sigma}^2_\bullet$, $\bullet\in\{\mathcal{M},\mathcal{T}_{\text{pre}}\}$ are fit by setting $\mathcal{T} = \sqrt{\nicefrac{\hat{\sigma}_\mathcal{M}}{\hat{\sigma}_{\mathcal{T}_\text{pre}}}}\cdot\left(\mathcal{T}_\text{pre}-\hat{\mu}_\mathcal{M}\right)+\hat{\mu}_{\mathcal{T}_\text{pre}}$.

\begin{table*}
	\centering
	\resizebox{\linewidth}{!}{
    	\begin{tabular}{c|cllcll}
    		
    		& Notation & Definition & Meaning & Distribution & Parameter & Values \\
    		\hline
    
            \multirow{10}{*}{\begin{turn}{90}Pattern\end{turn}}
    
            & $c_k$ & rings' center points & determined by tool path & $\mathcal{N}\left(\tilde{c}_k,\Sigma_c\right)$ & $\Sigma_c\in\mathbb{R}^{2\times 2}_{>0}$ & $\left(\Sigma_c\right)_{ii}=0.01\cdot\delta/\text{conf}$, $i=1,2$ \\[4pt]
    
            & $d$ & diameter of ring & \textbf{diameter of milling head} && $d\in\mathbb{R}_{>0}$ & $d\in\{4,8\}\text{ mm}$\\[4pt]
    
            & \multirow{2}{*}{$\alpha$} & defines $\rho$ (distance between & \textbf{amount of overlap of} && \multirow{2}{*}{$\alpha\in(0,1)$} & \multirow{2}{*}{$\alpha\in\{0.2,0.5,0.8\}$} \\
            && neighboring tool paths) & \textbf{neighboring tool paths} &&&\\[4pt]
    
            & \multirow{2}{*}{$\gamma$} & increase distance between & distance between blade && \multirow{2}{*}{$\gamma\in(0,\alpha)$} & \multirow{2}{*}{$\gamma=0.04$} \\
            && neighboring tool paths &  and outer edge of tool &&&\\[4pt]
      
            & \multirow{2}{*}{$\delta$} & distance between & depends on feed rate and && \multirow{2}{*}{$\delta\in\mathbb{R}_{>0}$} & \multirow{2}{*}{$\delta=0.09$ mm}\\
            && center points & tool's rotational speed &&& \\[4pt]
    
            & \multirow{2}{*}{$\epsilon$} & amount of rings &&& \multirow{2}{*}{$\epsilon\in[0,1]$} & \multirow{2}{*}{$\epsilon=0.01$}\\
            && with changed order &&&&\\
    
            \hline
       
    		\multirow{16}{*}{\begin{turn}{90}Appearance\end{turn}}
         
    		& \multirow{2}{*}{$w^-_k$} & width of & \multirow{2}{*}{width of cutting edge} & \multirow{2}{*}{$\mathcal{N}\left(\mu_{w^-},\sigma_{w^-}\right)$}
            & $\mu_{w^-}\in\left(0,\nicefrac{d}{2}\right)$ & $\mu_{w^-}=0.05\text{ mm}$ \\
            && indentation &&& $\sigma_{w^-}\in\mathbb{R}_{>0}$ & $\sigma_{w^-}=0.025/\text{conf}$ \\[4pt]
    
            & \multirow{2}{*}{$w^{+i}_k$} & width of & depends on & \multirow{2}{*}{$\mathcal{N}\left(\mu_{w^{+i}},\sigma_{w^{+i}}\right)$}
            & $\mu_{w^{+i}}\in\left[0,\nicefrac{d}{2}-\mu_{w^-}\right)$ & $\mu_{w^{+i}}=0.025\text{ mm}$ \\
            && inner accumulation & edges' sharpness && $\sigma_{w^{+i}}\in\mathbb{R}_{>0}$ & $\sigma_{w^{+i}}=0.1/\text{conf}$ \\[4pt]
    
            & \multirow{2}{*}{$w^{+o}_k$} & width of & depends on & \multirow{2}{*}{$\mathcal{N}\left(\mu_{w^{+o}},\sigma_{w^{+o}}\right)$}
            & $\mu_{w^{+o}}\in\mathbb{R}_{\geq 0}$ & $\mu_{w^{+o}}=0.025\text{ mm}$ \\
            && outer accumulation & edges' sharpness && $\sigma_{w^{+o}}\in\mathbb{R}_{>0}$ & $\sigma_{w^{+o}}=0.1/\text{conf}$ \\
    
            \cline{2-7}
    
            & $\phi_k$ & tilting direction & depends on tool path && $\phi_k\in(-\pi,\pi]$ & computed by $c_k$\\[2pt]
    
            & \multirow{2}{*}{$l^-_k$, $h^-_k$} & minimal/maximal scaling & \multirow{2}{*}{cutting depth with tilting} & \multirow{2}{*}{$\mathcal{N}\left(\mu_{\bullet^-},\sigma_{\bullet^-}\right)$}
            & $\mu_{\bullet^-}\in\mathbb{R}_{\geq 0}$ & $\mu_{l^-}=0.7$, $\mu_{h^-}=1$ \\
            && of indentation depth &&& $\sigma_{\bullet^-}\in\mathbb{R}_{>0}$ & $\sigma_{\bullet^-}=0.8/\text{conf}$ \\[4pt]
    
            & \multirow{2}{*}{$l^{+i}_k$, $h^{+i}_k$} & minimal/maximal scaling & depends on & \multirow{2}{*}{$\mathcal{N}\left(\mu_{\bullet^{+i}},\sigma_{\bullet^{+i}}\right)$}
            & $\mu_{\bullet^{+i}}\in\mathbb{R}_{\geq 0}$ & $\mu_{l^{+i}}=0$, $\mu_{h^{+i}}=0.2$ \\
            && of inner accumulation height & edges' sharpness && $\sigma_{\bullet^{+i}}\in\mathbb{R}_{>0}$ & $\sigma_{\bullet^{+i}}=0.5/\text{conf}$ \\[4pt]
    
            & \multirow{2}{*}{$l^{+o}_k$, $h^{+o}_k$} & minimal/maximal scaling & depends on & \multirow{2}{*}{$\mathcal{N}\left(\mu_{\bullet^{+o}},\sigma_{\bullet^{+o}}\right)$}
            & $\mu_{\bullet^{+o}}\in\mathbb{R}_{\geq 0}$ & $\mu_{l^{+o}}=0.2$, $\mu_{h^{+o}}=0.5$ \\
            && of outer accumulation height & edges' sharpness && $\sigma_{\bullet^{+o}}\in\mathbb{R}_{>0}$ & $\sigma_{\bullet^{+o}}=0.5/\text{conf}$ \\[4pt]
    
            \cline{2-7}
    
            & $\lambda_k$ & number of sine curves && $\mathcal{P}(\lambda)$ & $\lambda\in\mathbb{N}$ & $\lambda=50$ \\[4pt]
    
            & $\tau_{k_j}$ & frequency of sine curves && $\mathcal{P}(\tau)$ & $\tau\in\mathbb{N}$ & $\tau=50$ \\[4pt]
    
            & $\xi_{k_j}$ & shift of sine curves && $\mathcal{U}\left(-\pi,\pi\right)$ && \\

            \hline
    
            \multirow{4}{*}{\begin{turn}{90}\hspace{-2ex}Interaction\end{turn}}
    
            & \multirow{2}{*}{$a_k$} & minimal value && \multirow{2}{*}{$\mathcal{U}\left(a^{\text{min}},a^{\text{max}}\right)$} & $a^{\text{min}}\in[0,1]$ & $a^{\text{min}}= 0$ \\
            && convex combination &&& $a^{\text{max}}\in[a^{\text{min}},1]$ & $a^{\text{max}}= 0.3$ \\[4pt]
    
            & \multirow{2}{*}{$b_k$} & maximal value && \multirow{2}{*}{$\mathcal{U}\left(b^{\text{min}},b^{\text{max}}\right)$} & $b^{\text{min}}\in[0,1]$ & $b^{\text{min}}= 0.1$ \\
            && convex combination &&& $b^{\text{max}}\in[b^{\text{min}},1]$ & $b^{\text{max}}= 0.4$ \\
    
    	\end{tabular}
        }
	\caption{Overview of all parameters needed for the milling model including their meaning and choices thereof. Choose $\text{conf}=3$ for normally distributed parameters. Known parameters are highlighted (\cref{tab:processing_params}).}
	\label{Tab:Mill_parameter}
\end{table*}

\section{Defect Modeling}
\label{sec:defect_modeling}
In this work we use the approach of Bosnar et al. \cite{Bosnar2022DefectModelling} to generate multiple instances of defected object geometry, overview of which is provided in this chapter.

The defected object geometry is created through procedural geometry modeling of dents and scratches, thus ensuring correct light scattering.
The three main steps are (1) generation of defect positions, (2) generation of 3D geometries representing the defecting tools and (3) imprinting the defecting tools into the surface and creating geometrical masks which are used for generating defect annotations.
Defect positions are determined by sampling points from an arbitrary distribution on the inspected product geometry.
In this work we use normal distribution.
The defects are embedded into the geometry using defecting tools, i.e. negatives of dents (denting tools) and scratches (scratch tools).
Denting tools are typically spherical and parameterized for controlling their scale, rotation and shape.
Scratch tools are created by performing a random walk on the product geometry, resulting in a path which is further converted into a solid geometry.
Parameters are thickness and curviness, which control the size and shape of scratches.

The defecting tools are imprinted into the inspected product mesh using a Boolean difference operation.
Geometrical masks, used for annotation generation (see \cref{sec:methods:image_synthesis:object_shape_defecting}) are then created using the inspected product mesh and the defecting tool.
First, an intersection between the product mesh and the defecting tool mesh is computed.
The resulting mesh represents a solid filling of the defect.
The solid filling might be useful when the annotations must label missing area of the object.
In this work we label only visible defected surface and therefore perform an additional step which results in mask shell.
The defecting tool is first scaled with a factor in the range [0.9,1.0] and then intersected with the solid from the previous step.

\begin{figure*}[!h]
    \centering
        \includegraphics[width=0.1\textwidth]{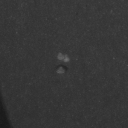}
        \includegraphics[width=0.1\textwidth]{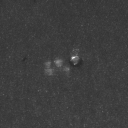}
        \includegraphics[width=0.1\textwidth]{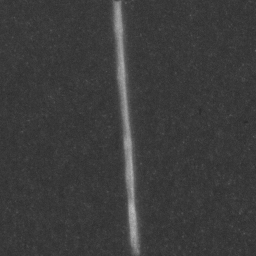}
    \hspace{0.3cm}
        \includegraphics[width=0.1\textwidth]{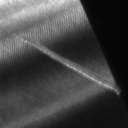}
        \includegraphics[width=0.1\textwidth]{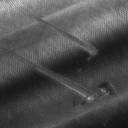}
        \includegraphics[width=0.1\textwidth]{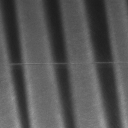}
    \hspace{0.3cm}
        \includegraphics[width=0.1\textwidth]{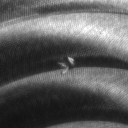}
        \includegraphics[width=0.1\textwidth]{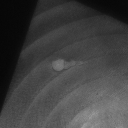}
        \includegraphics[width=0.1\textwidth]{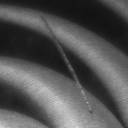}
    \\ \vspace{0.2cm}
        \includegraphics[width=0.1\textwidth]{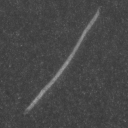}
        \includegraphics[width=0.1\textwidth]{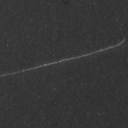}
        \includegraphics[width=0.1\textwidth]{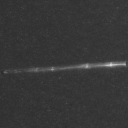}
    \hspace{0.3cm}
        \includegraphics[width=0.1\textwidth]{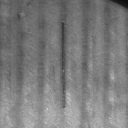}
        \includegraphics[width=0.1\textwidth]{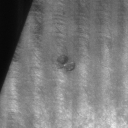}
        \includegraphics[width=0.1\textwidth]{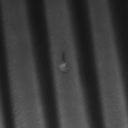}
    \hspace{0.3cm}
        \includegraphics[width=0.1\textwidth]{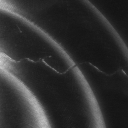}
        \includegraphics[width=0.1\textwidth]{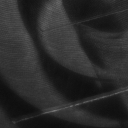}
        \includegraphics[width=0.1\textwidth]{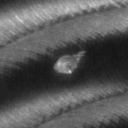}
    \\ \vspace{0.5cm}
        \includegraphics[width=0.1\textwidth]{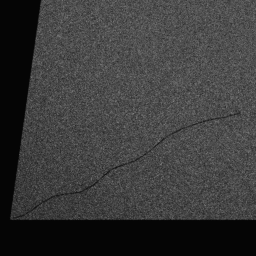}
        \includegraphics[width=0.1\textwidth]{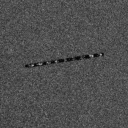}
        \includegraphics[width=0.1\textwidth]{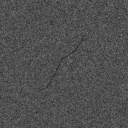}
    \hspace{0.3cm}
        \includegraphics[width=0.1\textwidth]{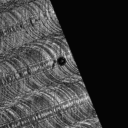}
        \includegraphics[width=0.1\textwidth]{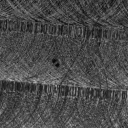}
        \includegraphics[width=0.1\textwidth]{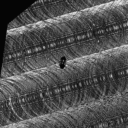}
    \hspace{0.3cm}
        \includegraphics[width=0.1\textwidth]{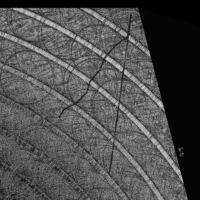}
        \includegraphics[width=0.1\textwidth]{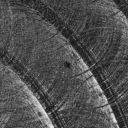}
        \includegraphics[width=0.1\textwidth]{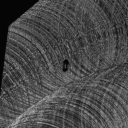}
    \\ \vspace{0.2cm}
        \includegraphics[width=0.1\textwidth]{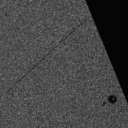}
        \includegraphics[width=0.1\textwidth]{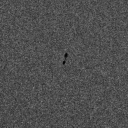}
        \includegraphics[width=0.1\textwidth]{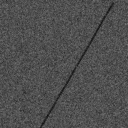}
    \hspace{0.3cm}
         \includegraphics[width=0.1\textwidth]{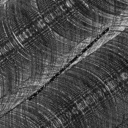}
        \includegraphics[width=0.1\textwidth]{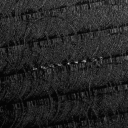}
        \includegraphics[width=0.1\textwidth]{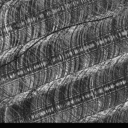}
    \hspace{0.3cm}
         \includegraphics[width=0.1\textwidth]{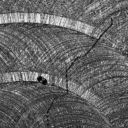}
        \includegraphics[width=0.1\textwidth]{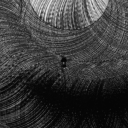}
        \includegraphics[width=0.1\textwidth]{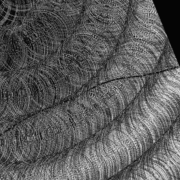}
    \caption{ \label{fig:defect_crops_real} Real (top) and synthetic (bottom) crops of defects on different surfaces. Left: sandblasted. Middle: parallel milling. Right: spiral milling.}
\end{figure*}

The defect parameters can be chosen to match the shape, size and positions of real defects observed in the measured defect topographies.
However, more importantly, the parameter range can also be extended to produce defects beyond those observed in the available samples, i.e., rare occurences or edge-cases.
This enables creation of an arbitrary number of defected 3D objects containing a wide variety of possible defects (see \cref{fig:defect_crops_real}) for which pixel-precise annotations are generated.

\section{Dual Dataset}
\label{sec:dual_dataset}
Publicly available datasets for inspection of metal objects mostly focus on defect recognition over flat surfaces and single-view inspection setups \cite{bo2023steeldefectreview}.
With it, they reduce the expected appearance diversity of defects and surfaces in the image, making the defect recognition task easier.
However, such restriction is not possible for inspection of more complex objects where multi-view imaging is required and visibility of the defects varies between different views.
The aforementioned datasets contain objects such as hot rolled steel \cite{kechen2013neu,guithong2019eneu,severstal,lv2020gc10det}, slightly curved metal pads \cite{tabernik2020ksdd,bozic2021ksdd2}, rails \cite{zhang2021rail5k} and pipes \cite{tianchi}.
Datasets of smaller metal objects with complex geometry are also present, however also under single-view inspection setups \cite{bergmann2019mvtecad,mishra2021btad}.
Multi-view inspection setups inspect the same surface multiple times while varying at least the direction of the incoming light source or the direction of the camera to reduce the chance of missing defects due to low visibility.
Examples in recent datasets include single-axis rotation \cite{schlagenhauf2021bsdata}, multiple illumination directions \cite{Honzatko2021csem_misd} and multiple view directions \cite{fulir2023syntheticclutch}.

Synthetic datasets for inspection exist in form of generic stochastic textures \cite{matthias2007dagm}, outdoor maintenance scenes \cite{bao2022miad}, multi-object pose estimation task \cite{roovere2022dimo}, industrial part recognition \cite{Zhu2023partclass} and surface defect recognition \cite{fulir2023syntheticclutch}.

We present a dual dataset consisting of a real and synthetic part, following the multi-view inspection setup described in \cite{fulir2023syntheticclutch}.
Our dataset focuses on recognition difficulties arising from complex reflectance of different surface finish patterns which may cause defect curtaining.

\subsection{Real Dataset}
\label{sec:dual_dataset:real_acquisition}
\begin{figure}
    \centering
    \includegraphics[width=0.99\linewidth]{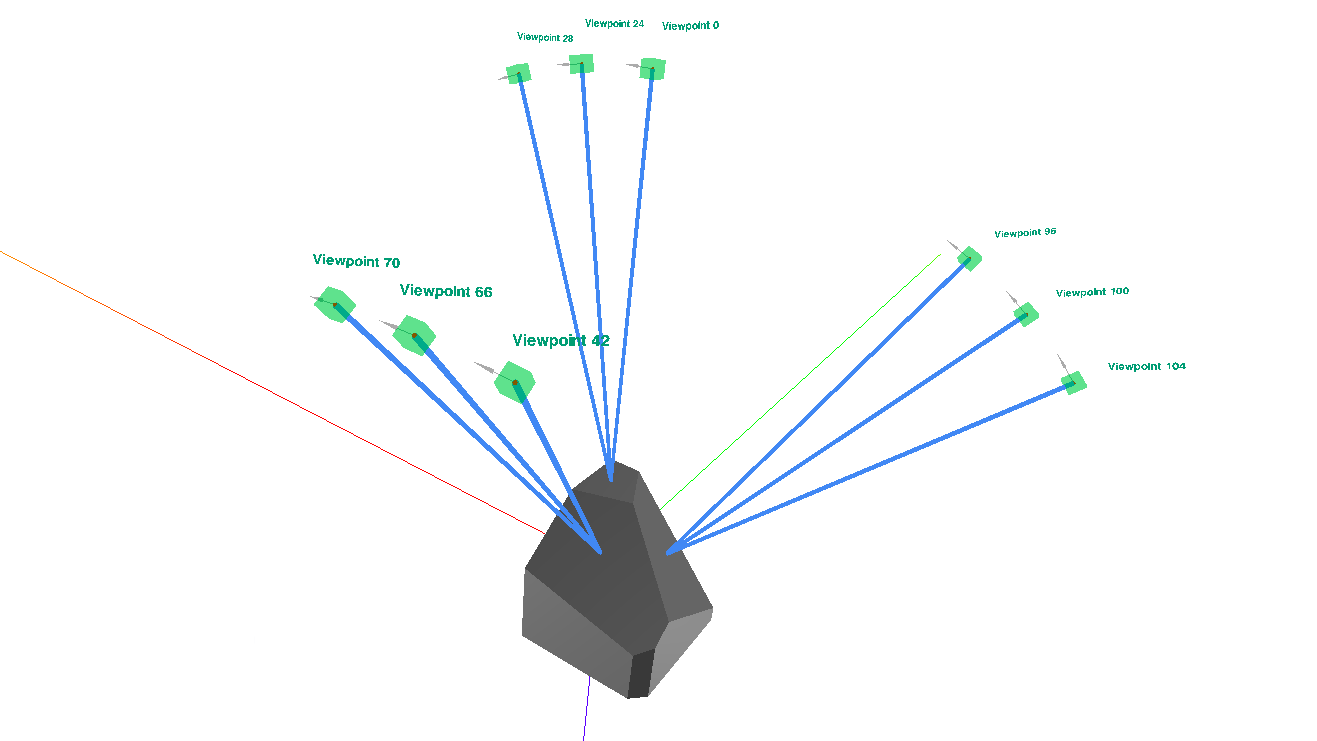}
    \caption{Visualization of viewpoints used for inspection. The blue line shows the optical axis of the camera and has the length of focusing distance.}
    \label{fig:vpoi_viz}
\end{figure}

The real dataset was acquired using a setup consisting of a robot manipulator, a matrix grayscale camera, a diffuse ring light and an acquisition table.
A thick black curtain removes any influence of the environment illumination or reflection from within the acquisition box.
The manipulator is used to position the camera and illumination onto predefined viewpoints.
The acquisition table is a flat surface covered in diffuse black velvet and the acquired object is manually placed on top.
The acquisition viewpoints are defined using V-POI \cite{gospodnetic2020surface} as an inspection plan, relative to the 3D model of the acquired object.
Our acquisition plan consists of viewpoints focused on the middle of each inspected object face (faces A, B, and C), at angles $0$, $10$ and $20$ degrees from the respective surface normal as visualized in \cref{fig:vpoi_viz}. 

To minimize the placement error between the expected and actual position of the object, the operator must manually perform \textit{object positioning}. 
This routine uses a predefined reference viewpoint - in our case the viewpoint perpendicular to the C-face - and a synthetic image of the object taken from it.
The physical camera is placed into the position defined by the reference viewpoint and the reference synthetic image is overlayed.
The object is then aligned manually to make the observed image match the reference image.
For this procedure to be appropriate, either the synthetic image must contain image distortion introduced by the lens or the real image must be undistorted. 

The real dataset consists of images of the test bodies defined in \cref{sec:method:test_body_design}, acquired using the setup above and manually annotated using LabelMe \cite{WadaLabelMe}.
The dataset contains 30 objects, 20 of which are defective, and 9 viewpoints per object.
Overall this amounts to 270 annotated images, 180 of which display defects.
The dataset was split into train-val-test sets.
The test set contains the 10 non-defective objects and 10 defective objects.
Train and validation sets contain the remaining 10 defective objects, where only 1 viewpoint per object is used for the validation set.

\subsection{Synthetic dataset}
\label{sec:dual_dataset:generation}

The synthetic dataset generation environment described in \cref{sec:method:synth_datagen} was developed to match the physical acquisition setup.
The dataset generator is an automated procedure which uses object, texture and defect specifications for generating annotated synthetic data in multiple stages.
The first stage of the dataset generator is geometric defecting as described in \cref{sec:defect_modeling}, which generates and applies defects to the original object geometry. 
We generate $30$ defective synthetic object instances through stochastic sampling of defect locations and specifications defined in \cref{tab:defect_specs}.
The defect locations were sampled independently over the inspected surfaces using uniform barycentric face sampling.
Finally, we group our generated defects into the classes 'dent' and 'scratch'.

\begin{table*}[t]
    \centering
    \begin{tabular}{lrrrr}
        
        Parameter & Small dent & Big dent & Flat scratch & Curvy scratch \\
        \hline
        Quantity & 5 & 3 & 2 & 2 \\
        Diameter & [0.02, 0.2] & [0.2, 1.0] & [0.02, 0.2] & [0.02, 0.1] \\
        Elongation & [1, 2] & [1, 4] & - & - \\
        Depth & [0.05, 0.2] & [0.2, 1.0] & - & - \\
        Path length & - & - & [5, 20] & [10, 20] \\
        Step size & - & - & 0.1 & 1.0 \\
        
    \end{tabular}
    \caption{Defect quantities and specification ranges obtained from approximate measurements of their correspondents in physical samples, with gentle increase in ranges to model expected unobserved defects.}
    \label{tab:defect_specs}
\end{table*}

The second stage produces texture images of the object surface using the methods introduced in \cref{sec:texture_modeling}.
For each generated texture instance, the texture parameters are sampled at random from the values listed in \cref{tab:texture_vars}, thus producing controlled surface appearance variation.
The ranges were chosen empirically by observing differences in physical samples and to model plausible variation in textures, such as milling at different translational speeds or amplitudes of parallel and orthogonal tool vibrations.
For each of the three surface finishing methods, we generate $5$ different texture instances.

\begin{table*}[t]
    \centering
    \begin{tabular}{ll}
        Parameter & Set of values \\
        \hline
        Ring center points noise ($\Sigma_c$) & $\left\{\textbf{0.01}, 0.05, 0.1, 0.2, 0.4, 0.6, 0.8\right\} \cdot \nicefrac{\delta}{\text{conf}}$ \\
        Ring distance ($\delta$) & $\left\{0.7, 0.75, 0.8, 0.85, \textbf{0.9}, 0.95, 1.0, 1.05, 1.1\right\}$ \\
        Ring flip probability ($\epsilon$) & $\left\{\textbf{0.01}, 0.05, 0.1, 0.2, 0.3, 0.4, 0.5\right\}$ \\
        Ring width noise ($\sigma_{\omega^-}$) & $\left\{0.015, 0.02, \textbf{0.025}, 0.03, 0.035\right\} \cdot \nicefrac{1}{\text{conf}}$ \\
        Ring depth noise ($\sigma_{l^-},\sigma_{h^-}$) & $\left\{0.4, 0.6, \textbf{0.8}, 1.0, 1.2\right\} \cdot \nicefrac{1}{\text{conf}}$ \\
        Cosine curves number ($\lambda$) & $\left\{30, \textbf{50}, 70\right\}$\\
        
    \end{tabular}
    \caption{Texture parameter ranges used to modify the default values (highlighted) given in \cref{Tab:Mill_parameter}. The standard deviations were chosen by the $3 \sigma$ rule such that the desired ranges of the random variables are obtained. Hence, we set $\text{conf}=3$.}
    \label{tab:texture_vars}
\end{table*}

The third stage combines the pre-generated synthetic objects and textures to produce annotated inspection images, as described in \cref{sec:method:synth_datagen}.
A set of synthetic image-mask pairs are rendered for every synthetic object, following the same pre-defined inspection plan.
Before rendering, each object surface gets assigned one of the pre-generated textures at random.
Additionally, each time a texture is assigned it is randomly rotated by an angle in the range $[-15^{\circ},15^{\circ}]$ and translated by $5$, $3$ or $1$ pixels for faces A, B and C, respectively.
To model different surface oxidation levels, for each synthetic object we uniformly sample the material roughness parameter from the range $[0.05,0.3]$.
To produce a balanced dataset, we generate the same amount of image-mask pairs by re-using the correct object geometry instead of the defective ones.

Overall, the synthetic dataset consists of $30$ correct and $30$ defective object instances for each of the $10$ physical objects.
To remove the bias arising from the location and choice of defects when comparing results between textures, we re-use the same $30$ defective object geometries across all of the physical objects.
Each object is inspected using the same $9$ viewpoints used in the pre-defined inspection plan, totaling in $5400$ images, $2700$ of which contain defects.
The images were rendered using $256$ SPP and $8$ light bounces at the resolution $1224\times 1025$.
The train-val-test set was produced ensuring label balance in the ratio $70:10:20$ by splitting along the dimension of object instances.

\begin{figure*}
    \centering
    \begin{subfigure}[t]{0.195\textwidth}
        \centering
        \includegraphics[width=\textwidth]{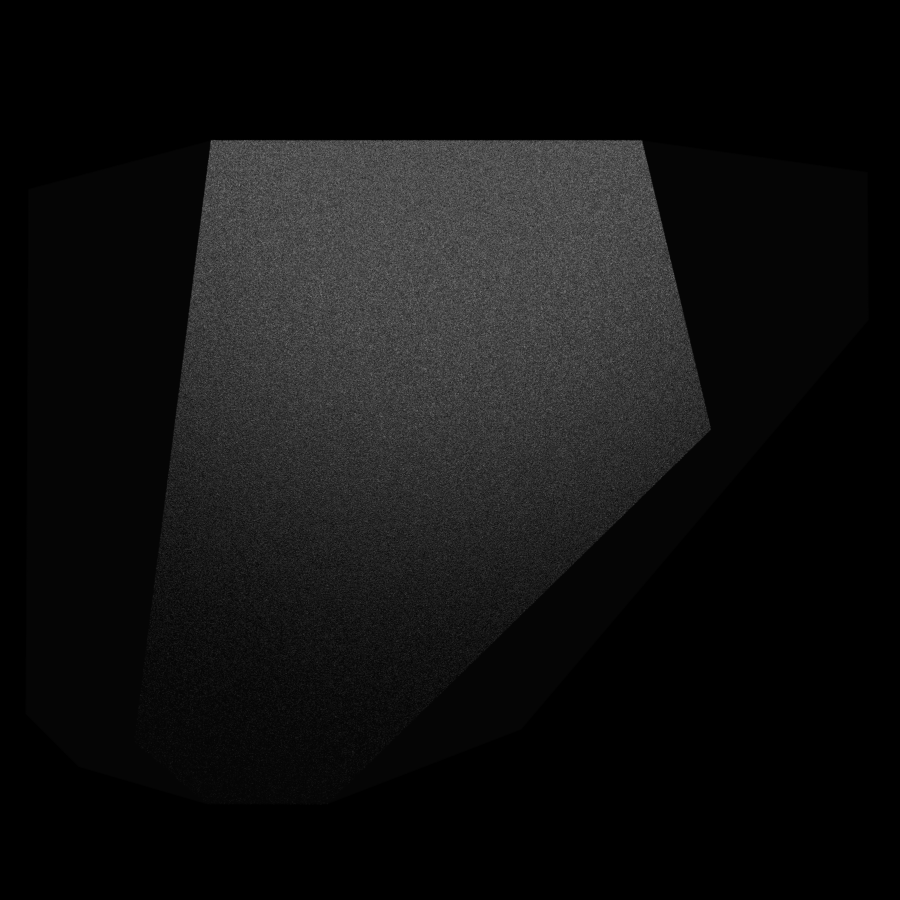} \\[2pt]
        \includegraphics[width=\textwidth]{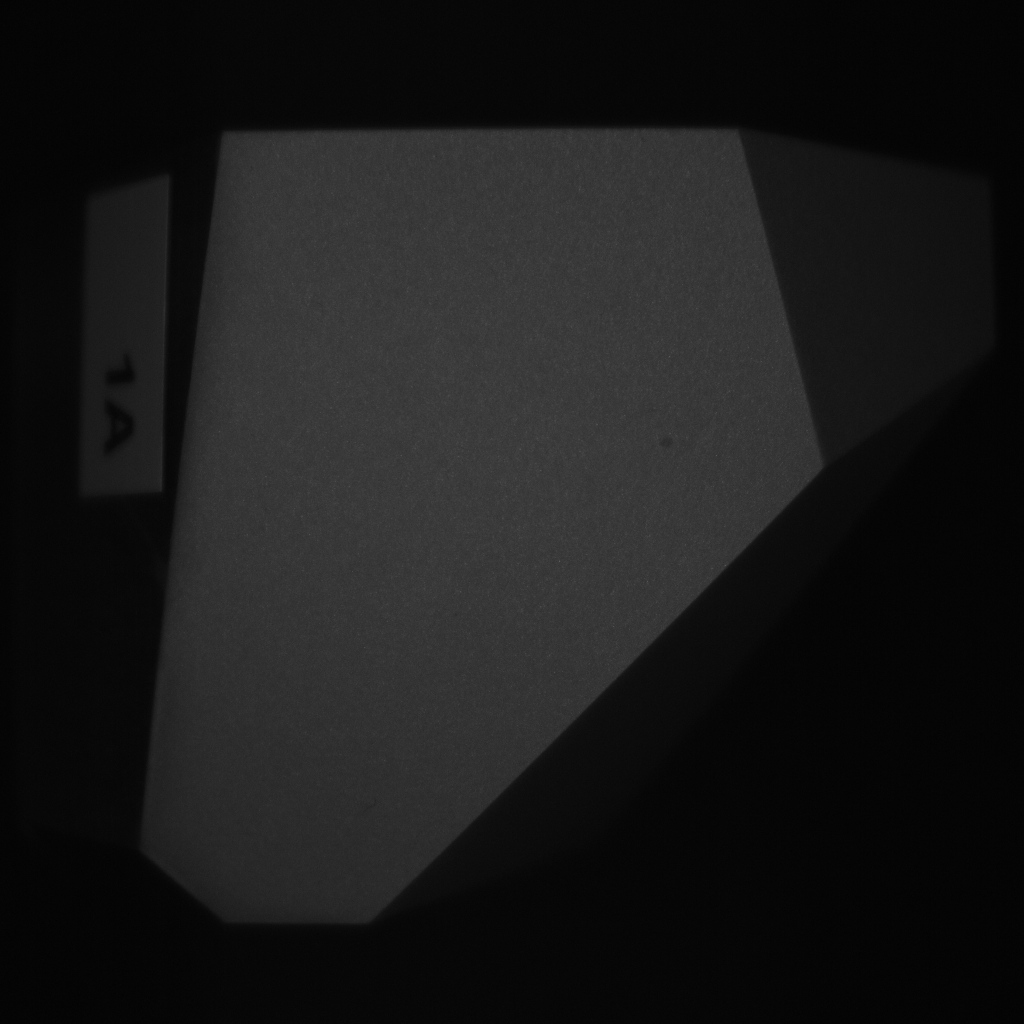}
        \caption*{Sandblasting 2.5 bar.}
    \end{subfigure}
    \begin{subfigure}[t]{0.195\textwidth}
        \centering
        \includegraphics[width=\textwidth]{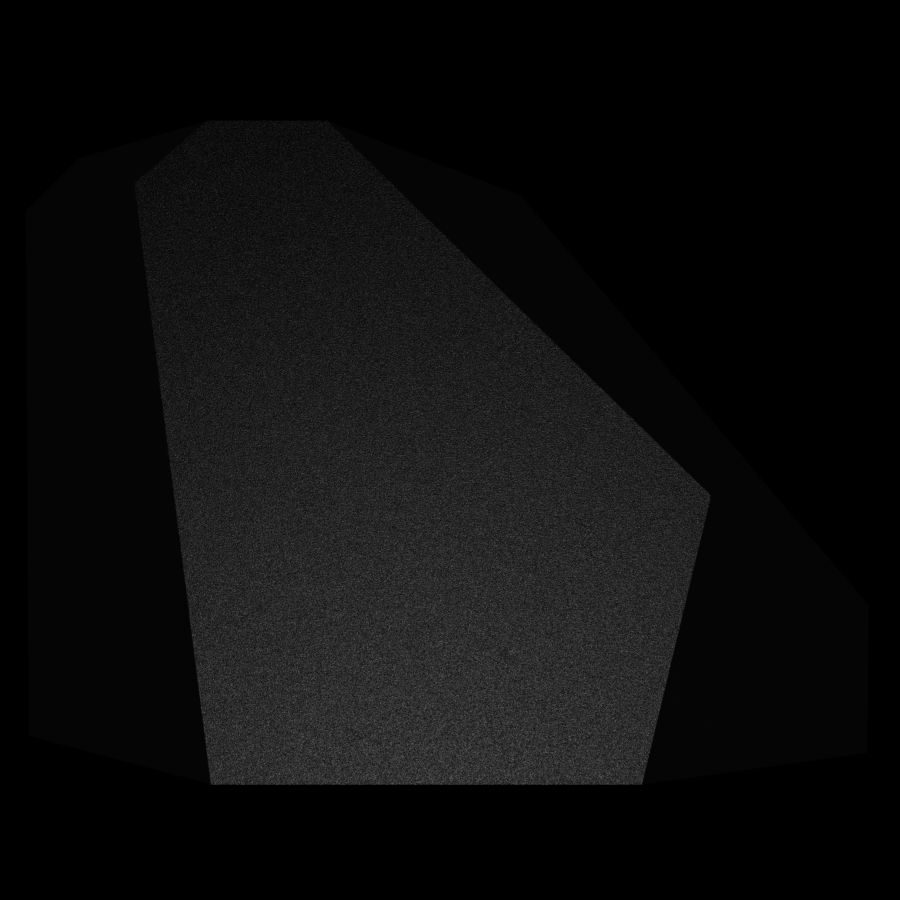} \\[2pt]
        \includegraphics[width=\textwidth]{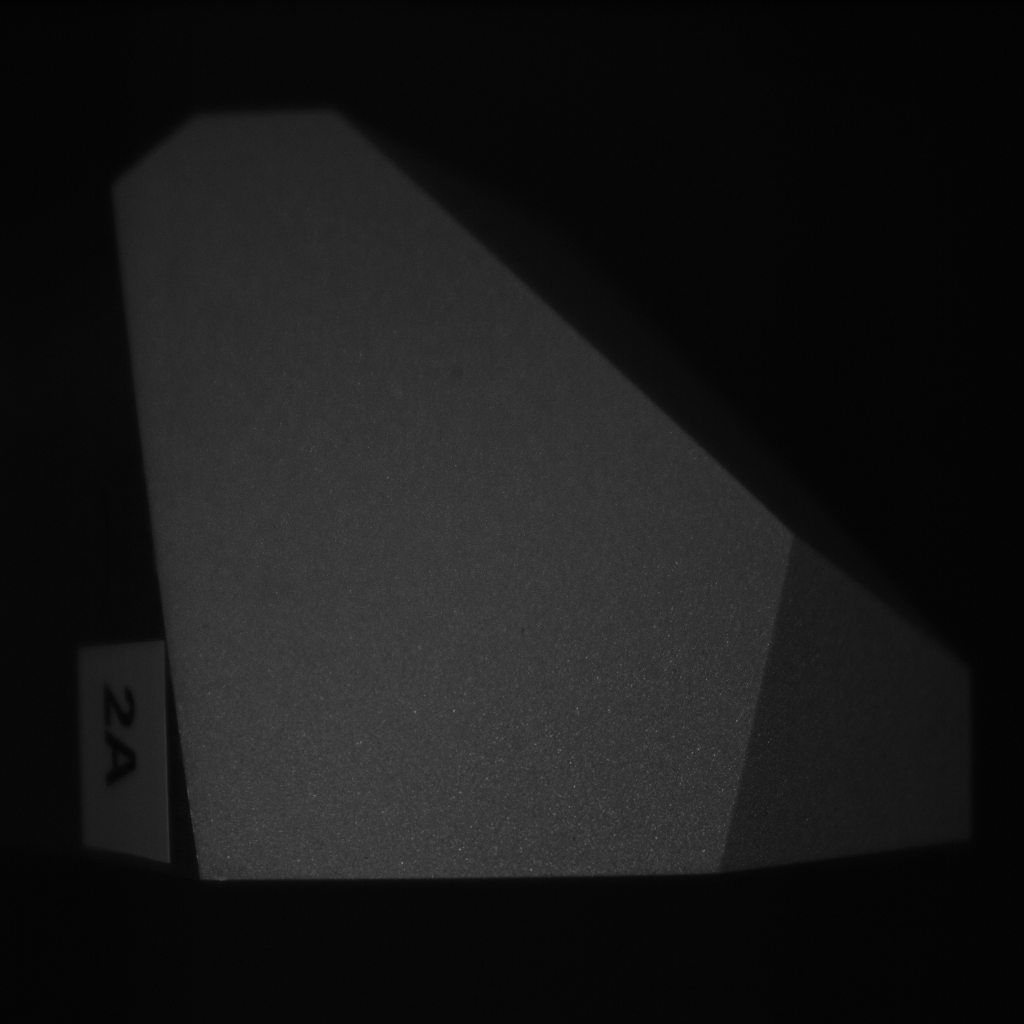}
        \caption*{Sandblasting 6 bar.}
    \end{subfigure}
    \begin{subfigure}[t]{0.195\textwidth}
        \centering
        \includegraphics[width=\textwidth]{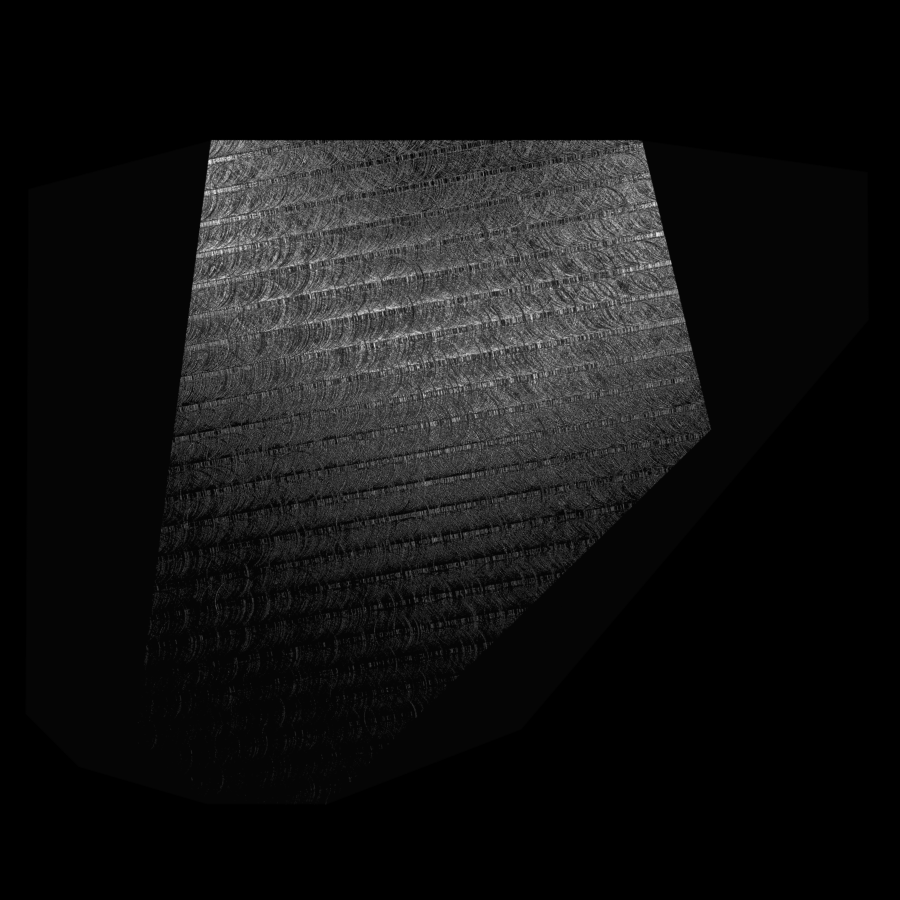} \\[2pt]
        \includegraphics[width=\textwidth]{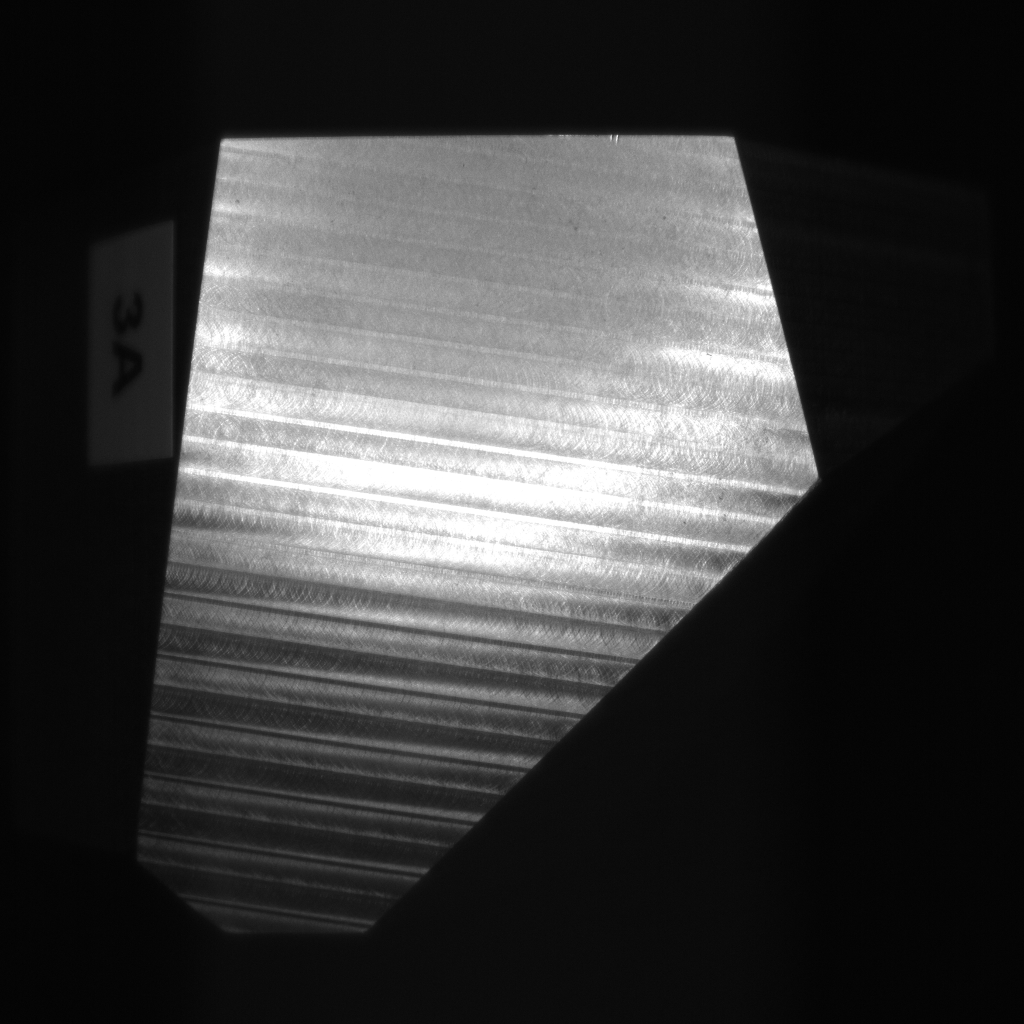}
        \caption*{Parallel milling, head diameter 4 mm, radial depth of cut 0.5.}
    \end{subfigure}
    \begin{subfigure}[t]{0.195\textwidth}
        \centering
        \includegraphics[width=\textwidth]{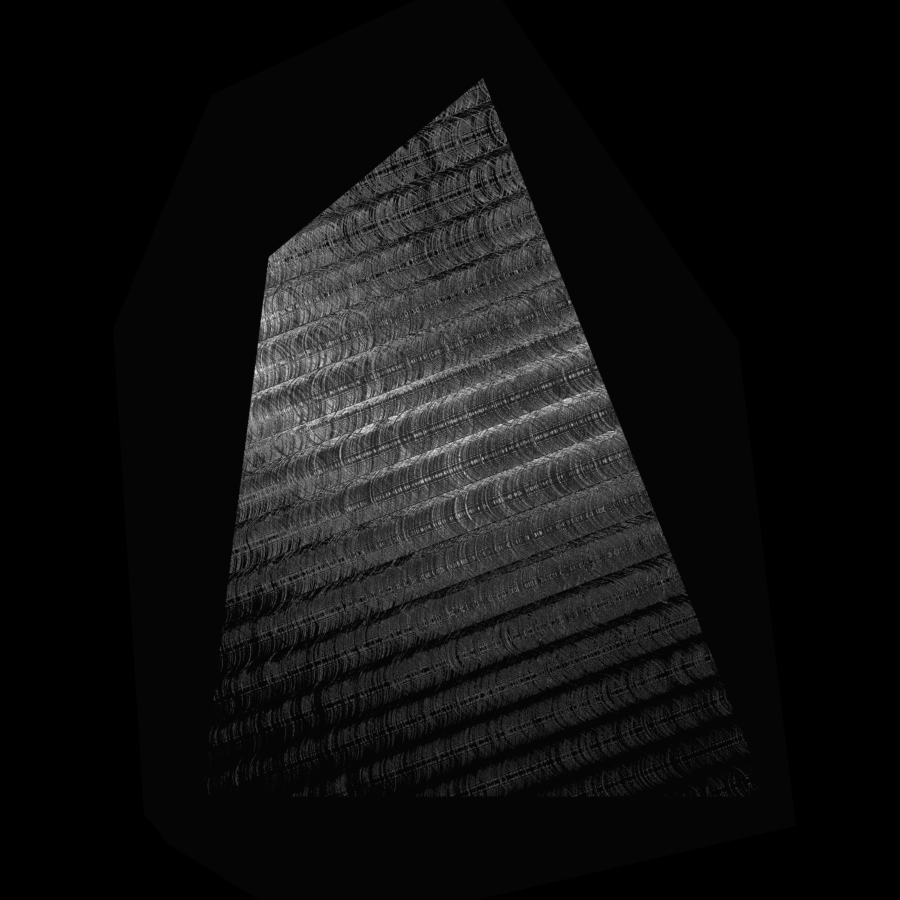} \\[2pt]
        \includegraphics[width=\textwidth]{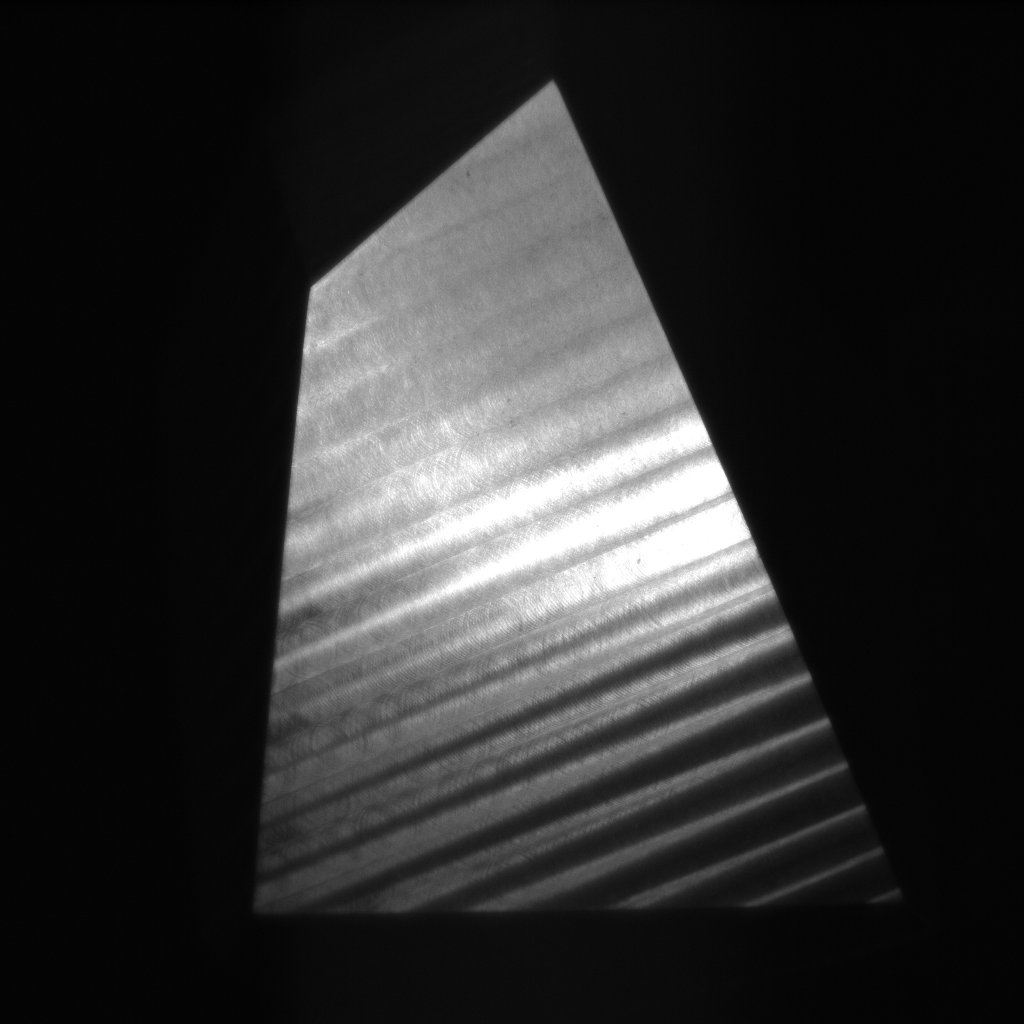}
        \caption*{Parallel milling, head diameter 4 mm, radial depth of cut 0.8.}
    \end{subfigure}
    \begin{subfigure}[t]{0.195\textwidth}
        \centering
        \includegraphics[width=\textwidth]{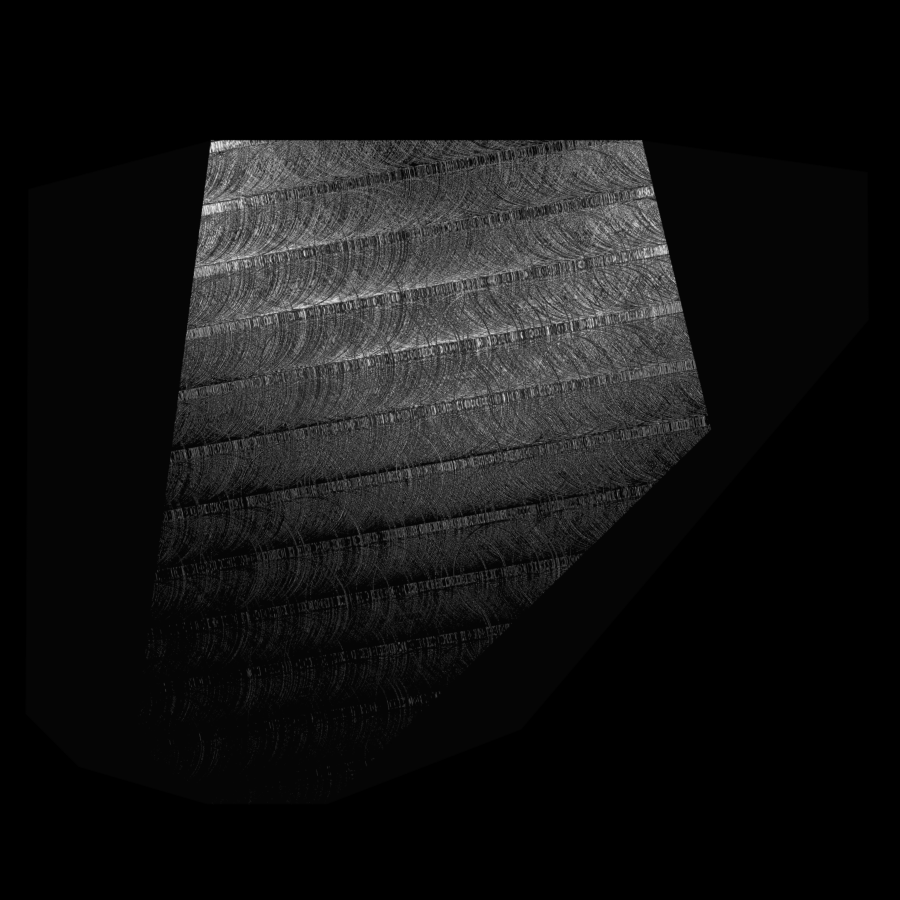} \\[2pt]
        \includegraphics[width=\textwidth]{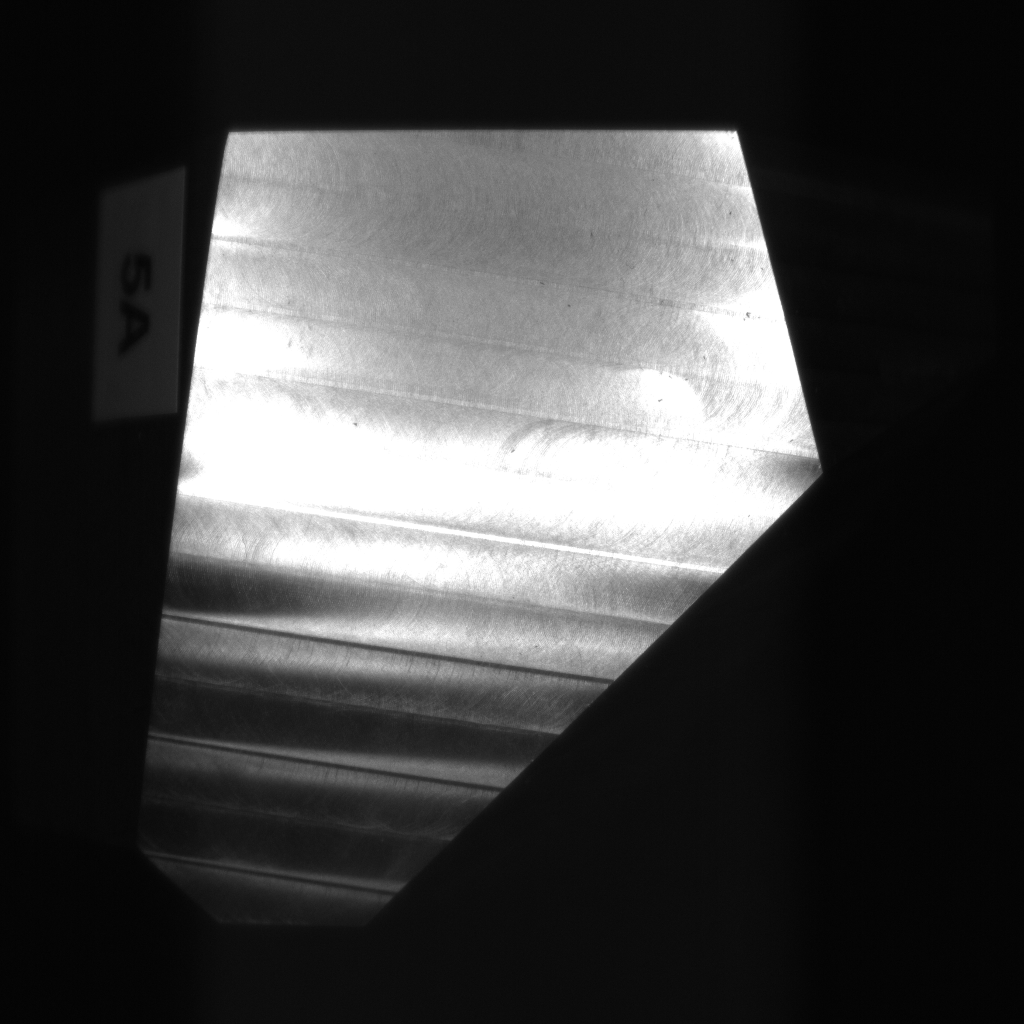}
        \caption*{Parallel milling, head diameter 8 mm, radial depth of cut 0.5.}
    \end{subfigure}

    \begin{subfigure}[t]{0.195\textwidth}
        \centering
        \includegraphics[width=\textwidth]{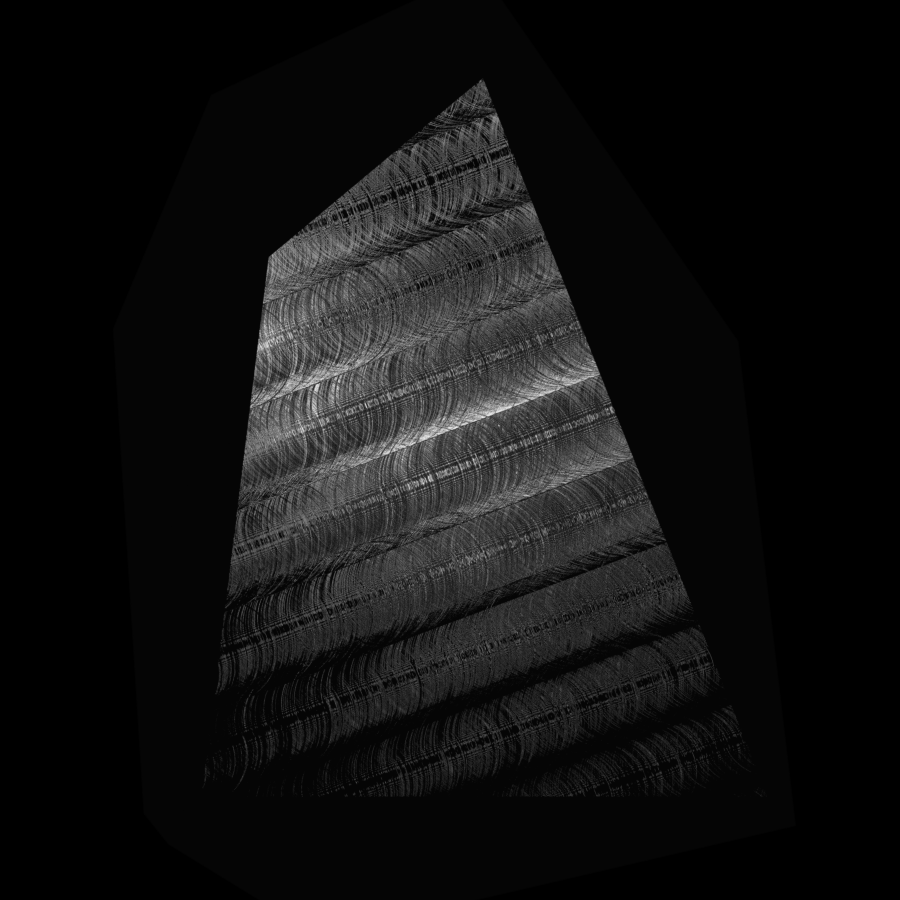} \\[2pt]
        \includegraphics[width=\textwidth]{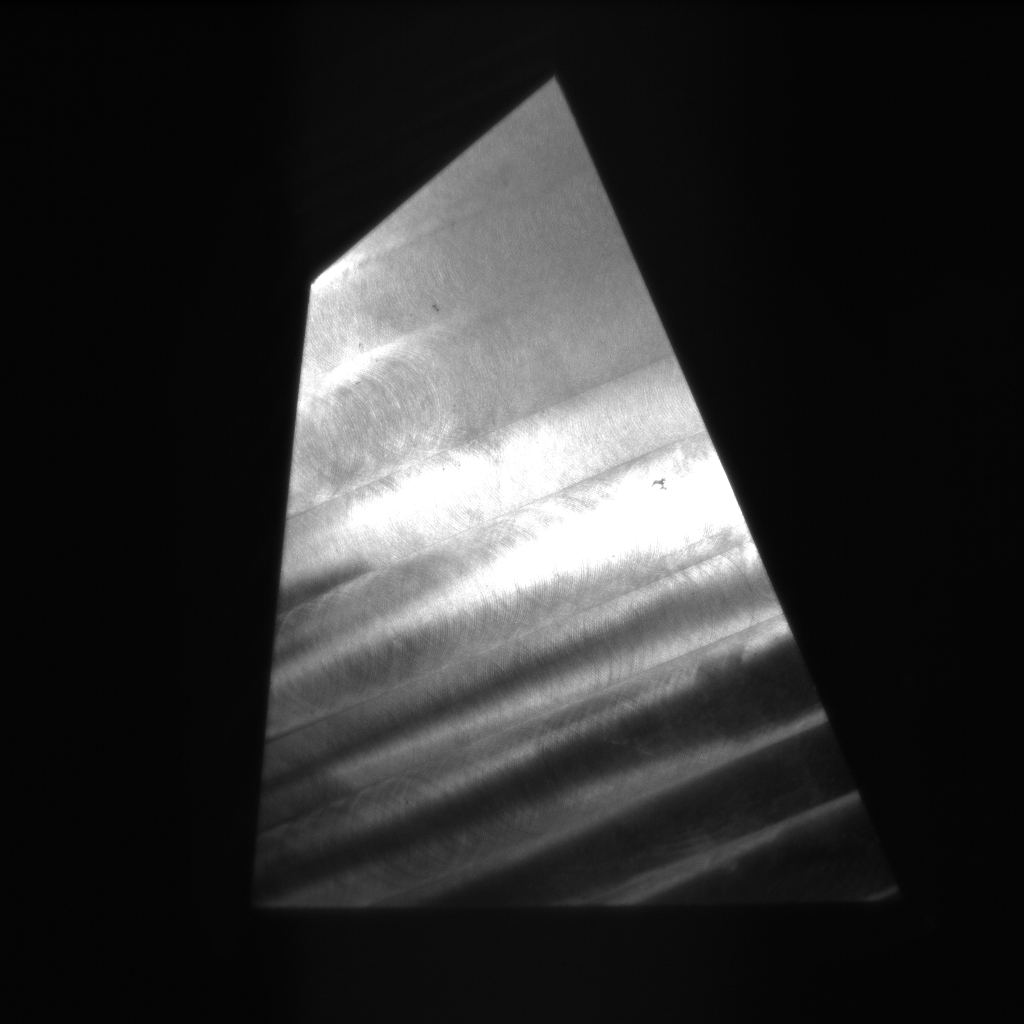}
        \caption*{Parallel milling, head diameter 8 mm, radial depth of cut 0.8.}
    \end{subfigure}
    \begin{subfigure}[t]{0.195\textwidth}
        \centering
        \includegraphics[width=\textwidth]{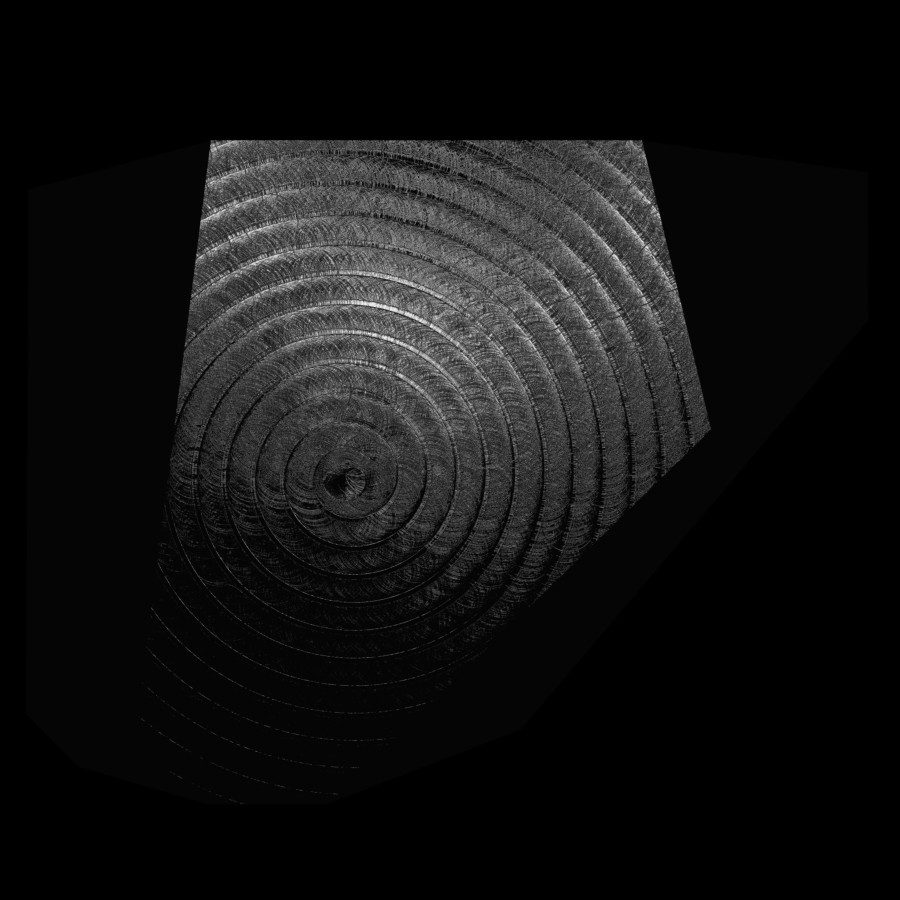} \\[2pt]
        \includegraphics[width=\textwidth]{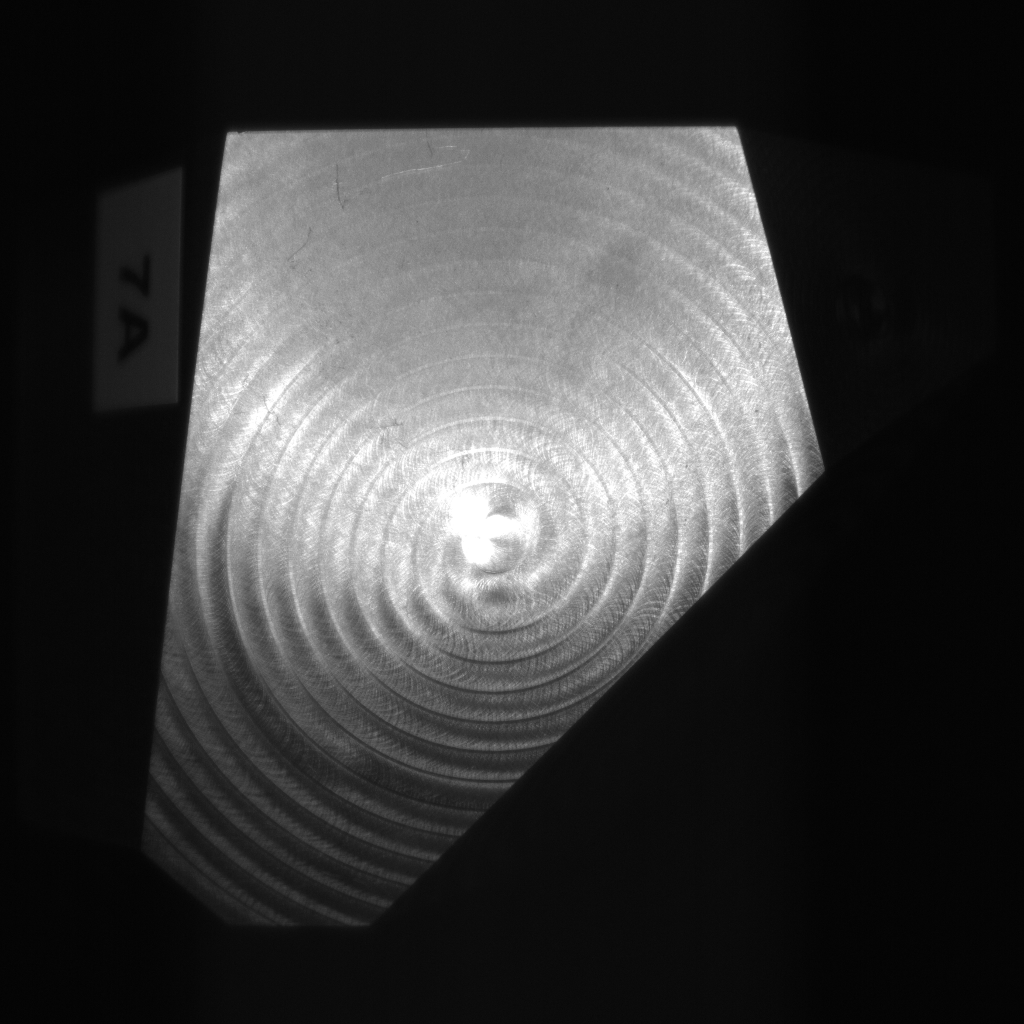}
        \caption*{Spiral milling, head diameter 4 mm, radial depth of cut 0.5.}
    \end{subfigure}
    \begin{subfigure}[t]{0.195\textwidth}
        \centering
        \includegraphics[width=\textwidth]{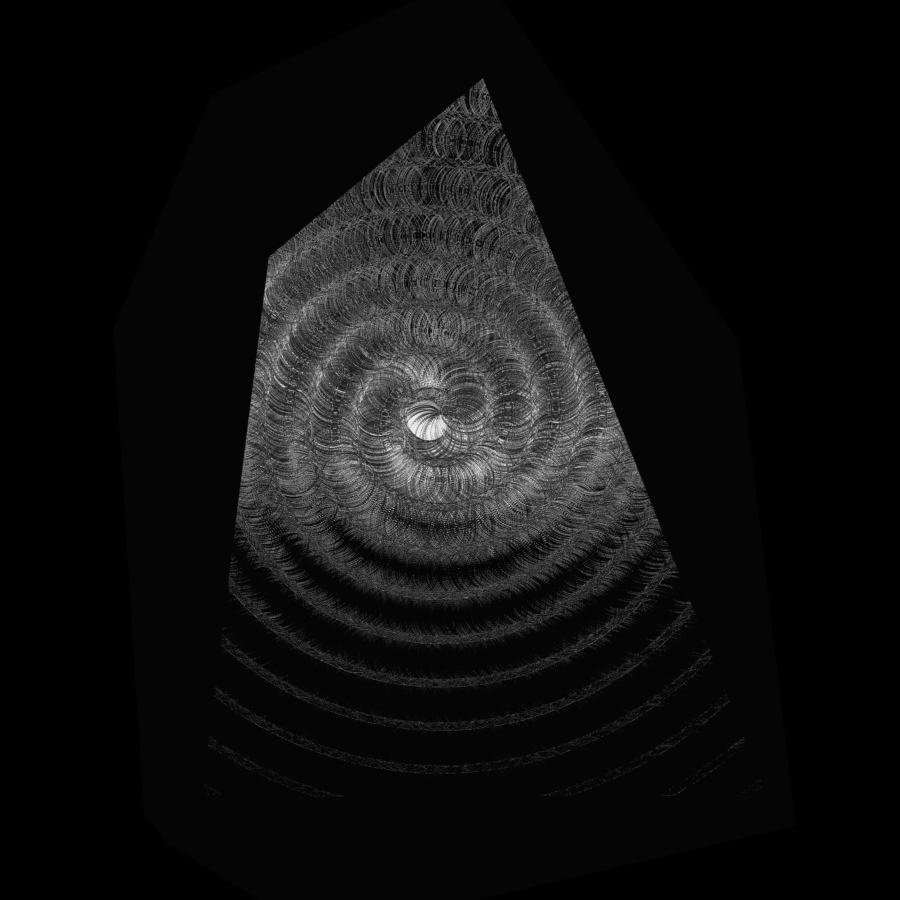} \\[2pt]
        \includegraphics[width=\textwidth]{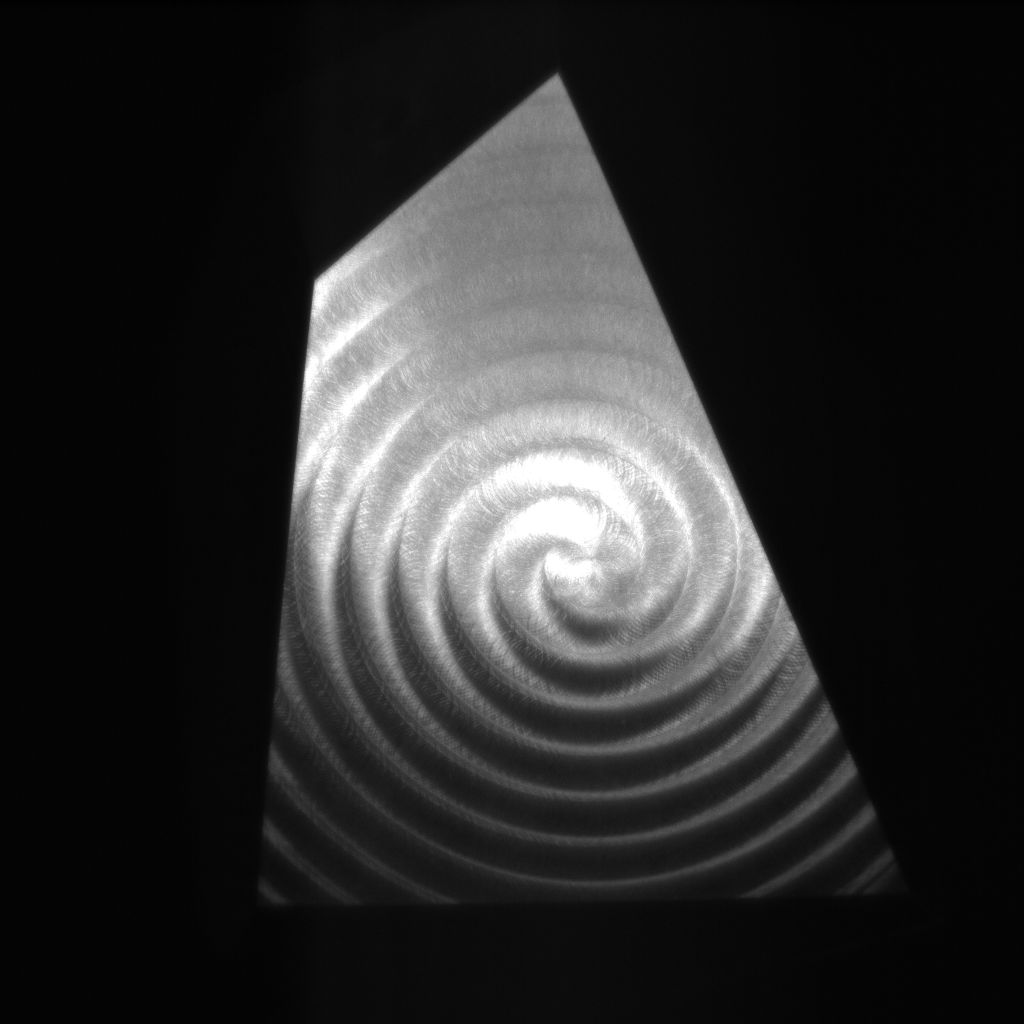}
        \caption*{Spiral milling, head diameter 4 mm, radial depth of cut 0.8.}
    \end{subfigure}
    \begin{subfigure}[t]{0.195\textwidth}
        \centering
        \includegraphics[width=\textwidth]{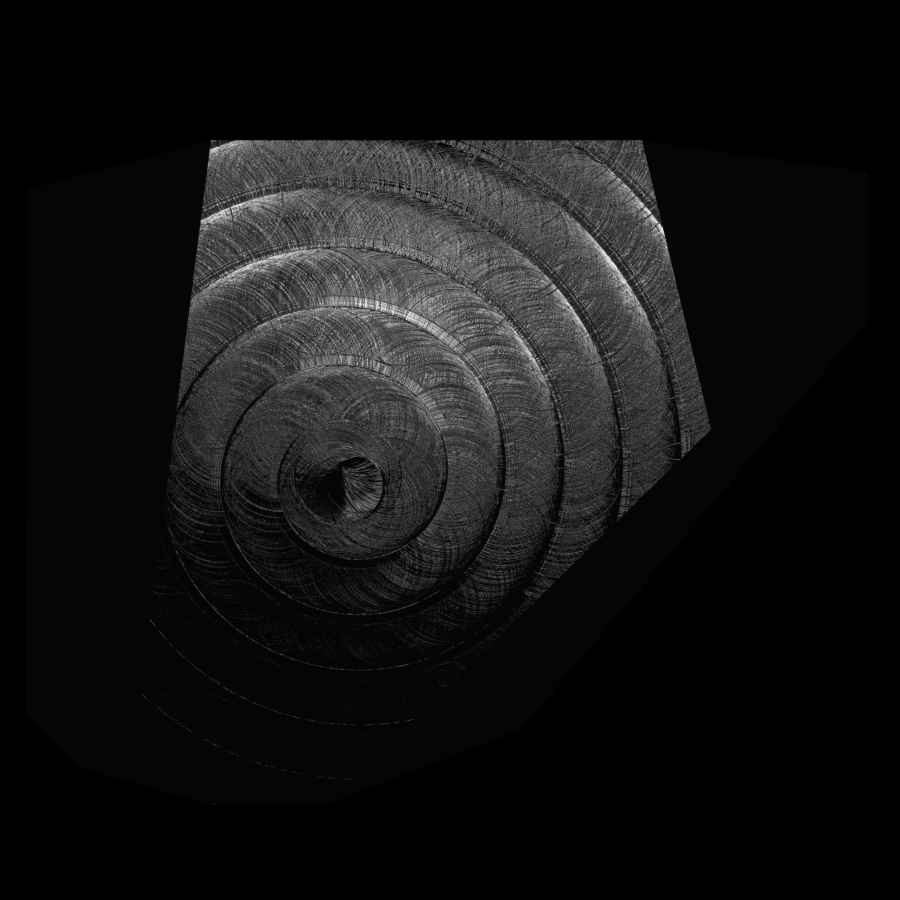} \\[2pt]
        \includegraphics[width=\textwidth]{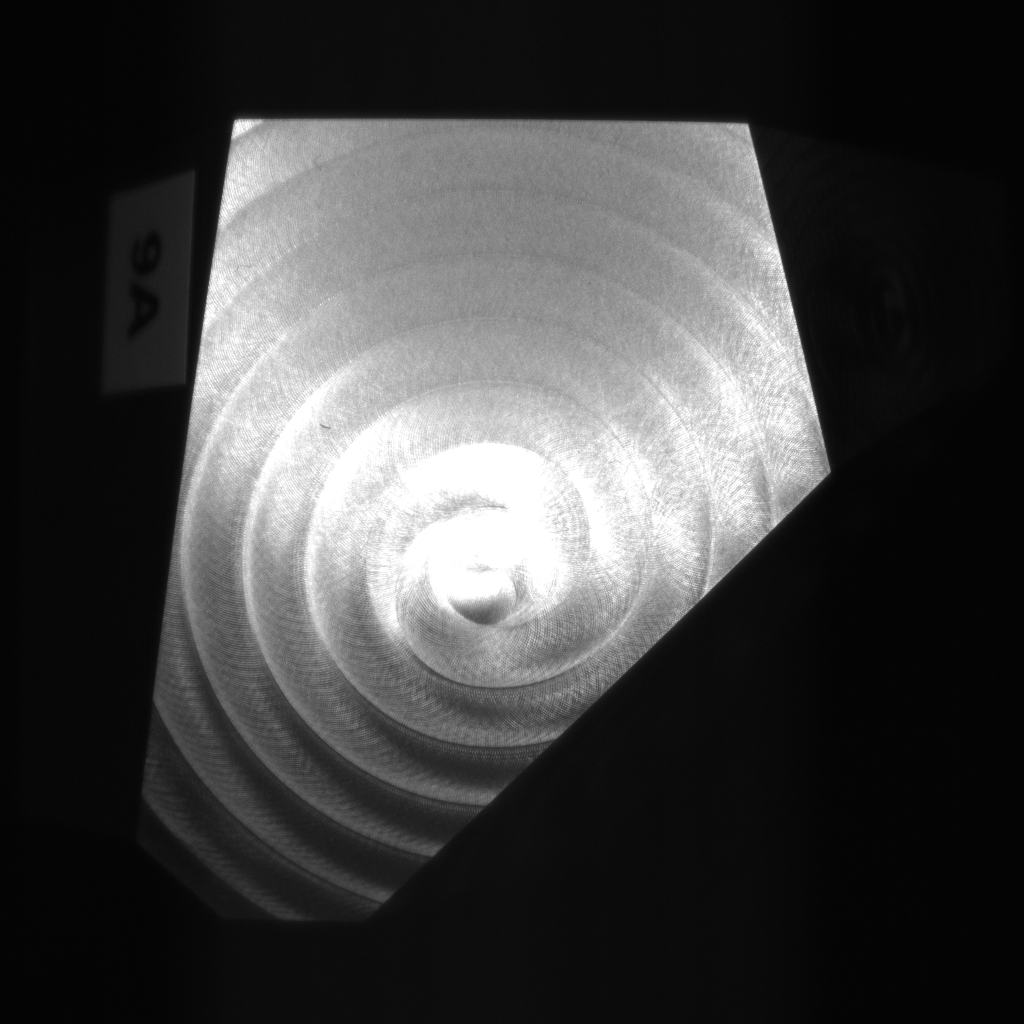}
        \caption*{Spiral milling, head diameter 8 mm, radial depth of cut 0.5.}
    \end{subfigure}
    \begin{subfigure}[t]{0.195\textwidth}
        \centering
        \includegraphics[width=\textwidth]{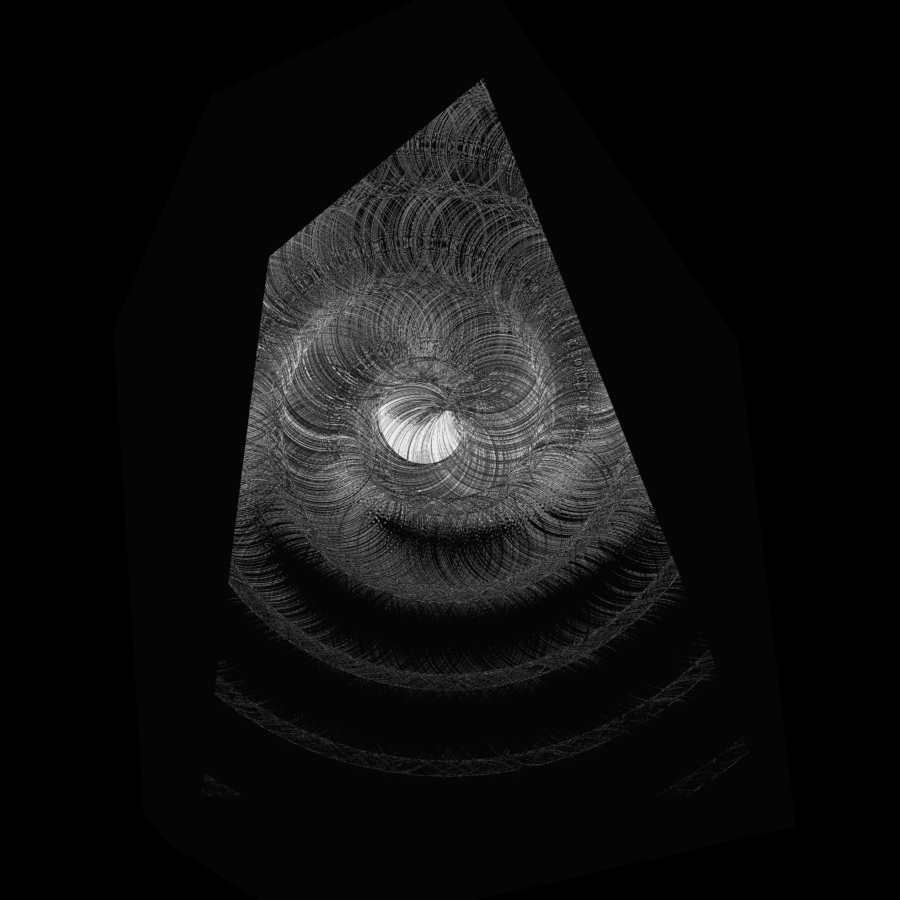} \\[2pt]
        \includegraphics[width=\textwidth]{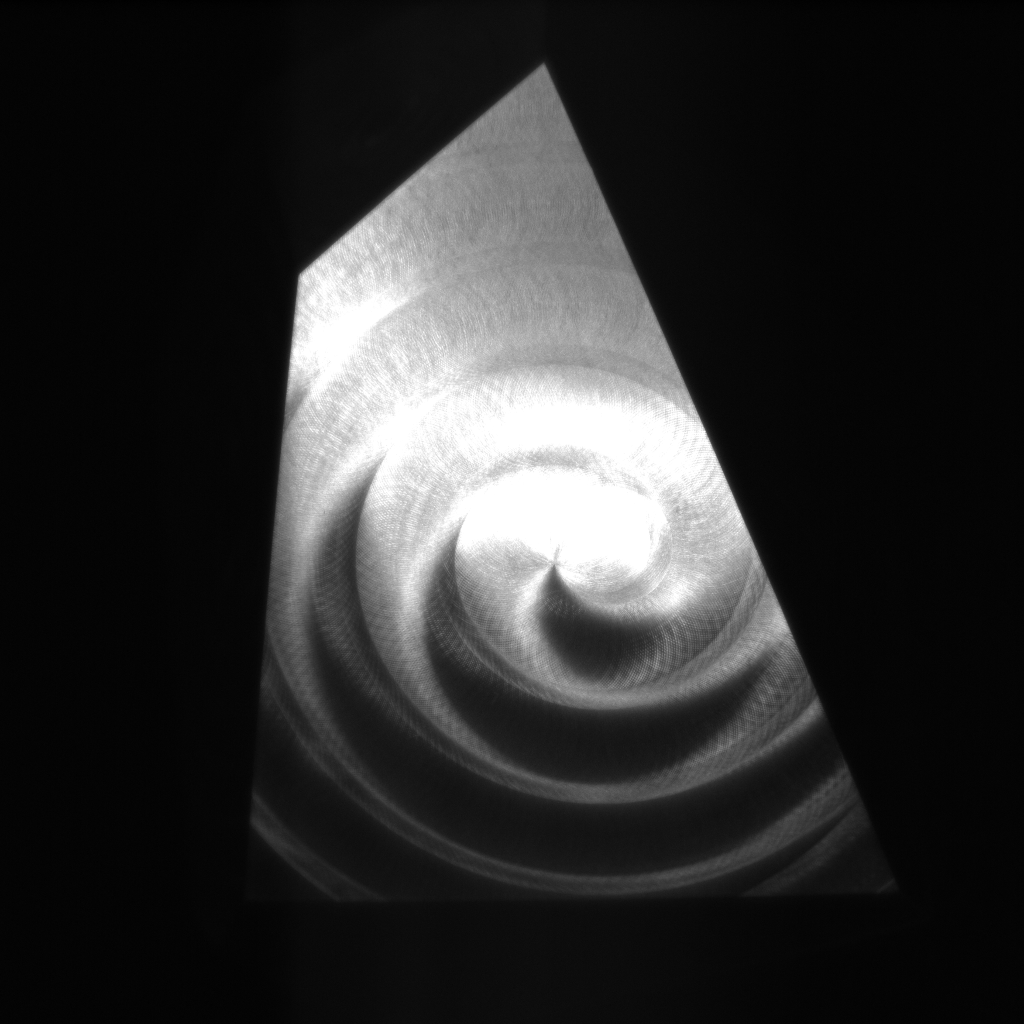}
        \caption*{Spiral milling, head diameter 8 mm, radial depth of cut 0.8.}
    \end{subfigure}
    \caption{Non-defective samples from the dual dataset. First and third row: synthesized images. Second and fourth row: real images of the same faces. The viewpoint is the 20-degree angle from a perpendicular view.}
    \label{fig:image_synthesis:real_synth_comp_set1}
\end{figure*}

A selection of images from our dataset is shown in \cref{fig:image_synthesis:real_synth_comp_set1}.

\section{Pipeline Evaluation}
\label{sec:pipeline_evaluation}
To evaluate the quality of the pipeline we have to evaluate similarity of the data it produces to the real data acquired through inspection.
Various metrics were proposed in recent works to estimate the quality of a dataset \cite{survey_data_quality} through general dataset factors such as annotation consistency or class balance.
However, these metrics usually provide limited insight into the data coverage of the dataset and ignore domain and task specific factors which could describe biases taken when collecting the dataset.
Another approach of comparing the quality of synthetic datasets to the real ones is by measuring their utility for the given task through the model performance changes when using the synthetic data \cite{Zhu2023partclass,fulir2023syntheticclutch,schmedemann2022procedural,Wood2021domainrand,tremblay2018domainrandom,gaidon2016virtualkitti,Tsirikoglou2020Survey}.
This however includes the inductive biases introduced by the model and training procedure selection making it more difficult to estimate the quality of the dataset itself.
We thus perform a more direct comparison of datasets through image and data distribution similarity.
When modeling a synthetic dataset, assumptions about the real world are made to increase the feasibility and economy of generating such datasets over the collection of more real data.
To organize our evaluations, we propose a distinction between \textit{a priori} quality, evaluating the accuracy of assumptions made before the data generation, and \textit{a posteriori} quality, evaluating the accuracy of the generated synthetic data.
To the best of the authors' knowledge, this is a first evaluation of synthetic data quality which goes beyond the comparison of recognition model performances.

\subsection{A priori quality}
\label{sec:quality:apriori_similarities}
The \textit{a priori} quality of a synthetic dataset quantifies the realism and precision of the environmental models used to generate the images compared to the target domain.
It is reported through assumptions used to define parameters for the data generation and can be described prior to generating the final synthetic dataset.
In our application, assumptions made to generate the dataset are tightly coupled to the physical processes used to manufacture and inspect the object.
Using the same CAD model as used in manufacturing of the object, we guarantee the produced images will represent the correct geometry and restrict the data to the object of interest (\cref{sec:metods:image_synthesis:texture_application}).
Using textures modeled to match the real surface measurements we restrict the recognition task to the nominal patterns present in the real objects (\cref{sec:texture_modeling}). 
Where possible, texture parameters are directly derived from processing parameters. All parameters
 are sampled around the nominal values to increase model robustness towards variations in the surface texture (\cref{sec:dual_dataset:generation}).
Defect randomization contains ranges of parameters such as expected defect dimensions and shapes that cover defects present in the real objects, but also defects outside the observed range for better generalization (\cref{sec:dual_dataset:generation}).
By using physics based light simulation we ensure that the images are produced in a realistic way, holding the assumptions used for the material models and the rendering processes defined in \cref{sec:methods:image_synthesis:rendering}.
The simulated acquisition hardware (camera and light) uses the real acquisition hardware parameters to resemble the real inspection environment.

\subsection{A posteriori quality}
The \textit{a posteriori} quality of a synthetic dataset describes the similarity of the generated data to the data collected in the target environment.
This type of quality estimation relies on the established concepts of domain shifts actively used for methods of domain adaptation \cite{Farahani2020domainadasurveybrief}.
We categorize these qualities into domain similarity and task similarity.

\textbf{Domain similarity} measures the similarity between the distributions of real and synthetic images to quantify the covariate shift.
In our case we measure the distance between generated images and the collected real images acquired using different types of metrics. 

Value-wise metrics compare the global distributions of intensities between images.
Distributions of image intensities can be summarized using histograms and compared using a suitable metric \cite{hist_metrics2010Ma}.

Pattern-wise similarity compares the local patterns between the images.
This type of comparison is affected by the scene structures, e.g. patterns of the surface finish, the light source shape and the material reflectance function.
Handcrafted image descriptors can be used to measure similarity of specific aspects of patterns in images, such as: pixel-wise closeness and alignment (e.g. mean absolute error, root mean squared error, cosine similarity) or statistical local similarity (e.g. SSIM \cite{Zhou2004SSIM}, peak signal-to-noise ratio, Adapted Rand \cite{Carreras2015adarad}, GLCM \cite{Haralick1973GLCM}).
Learned feature extractors alleviate the need to manually select specific features in images and instead rely on a very diverse dataset and deep learning to train a general purpose feature-space for pattern and shape comparison (e.g. LPIPS \cite{Zhang2018LPIPS}, FID \cite{Heusel2017FID}).
However, due to inherent differences between natural (e.g. dogs, traffic, people) and technical images (e.g. surface inspection, manufacturing line, product appearance), the generality is still bounded to imaging context.
To expand that context, one would need a large amount of representative technical images, which is not publicly available.

We evaluate the following metrics: Wasserstein distance between image histograms (HistWD), the mean absolute error (MAE), SSIM, and LPIPS using ImageNet pretrained AlexNet \cite{Krizhevsky2012alexnet}.
We limited our study to a single representative metric from each aforementioned subgroup with regards to their popularity in image quality estimation, interpretability and implementation availability.
Our experiments will prove sufficient to discuss the influence and shortcomings of different metrics (\cref{sec:discuss_domain_similarity}).
To make the results easier to read, the metrics were normalized to the range $[0,1]$ and flipped when necessary to report the degree of similarity (see \cref{tab:image_similarities}).
Our reasoning is that metrics which report the highest differences in the dataset (excluding trivial metrics) account best for small variations between the images in our datasets.

To compare our synthetic images to real ones we need to remove the influence of background noise and minor offsets, present from the imprecision of the acquisition system and background scene structures.
We first mask out all the regions that do not contain the observed object face, thus removing the influence of sharp edges between the dark background and the object.
For each viewpoint we construct a common meta-mask from the intersection of face masks corresponding to that viewpoint.
To reduce the difference in intensity levels, we manually estimate parameters of a linear transformation to bring the histograms between matching real and synthetic images closer together.
The minimal grey value in real images is typically larger than 0 while synthetic images also show grey values down to 0.   
Thus, for each texture type, we add the mean minimal grey value observed in the real images to all synthetic images. 

Next, we estimate a multiplication factor that jointly compensates for the illumination intensity and material absorption difference between synthetic and real data.
This method is justified because the factor can be extracted from the integral of the rendering equation \cite{kajiya1986rendering}, assuming that the inter-reflectance of the surface meso-structures is negligible. The assumption holds for our case since we use normal-mapping for meso-structure, which does not produce inter-reflectance.
The bias value is added to all of the pixels to compensate for the ambient illumination caused by the low-intensity inter-reflections of the inspected object and the inspection environment.
It can be observed in the image histograms as a gap of zeros between the illumination value $0$ and the first next intensity value in the histogram.
We estimate the bias by calculating the average gap size.
The estimated factors and biases are: $(0.632, 34)$ for sandblasted, $(1.481, 39)$ for parallel and $(1.526, 33)$ for spiral milling.
Finally, since the synthetic pinhole camera model does not include blurring and imperfections of the real camera, we artificially blur the synthetic images using defocus blur \cite{Hendrycks2019defocusblur} with radius $1.0$.

The synthetic data does not model an exact 1:1 replica of the texture found on a single real object, but stochastically varying texture appearance.
Therefore, we measure the distance to the closest synthetic sample, similarly to a one-sided Chamfer distance.
When evaluating the metrics, for each viewpoint we evaluate the metric between the real image and all available synthetic object instances and select the value representing the highest similarity (lowest distance).
In \cref{tab:image_similarities} for each texture type we report the metric value averaged across all viewpoints and objects sharing the same texture type.

\begin{table}[]
    \centering
    \begin{tabular}{lcccc}
        
        Method & Sandblasted & Parallel & Spiral & All \\
        \hline
        1-HistWD          & 0.980 & 0.932 & 0.942 & 0.946 \\
        1-MAE              & 0.807 & 0.681 & 0.655 & 0.696 \\
        SSIM                 & 0.896 & 0.561 & 0.585 & 0.638 \\
        1-LPIPS            & 0.916 & 0.660 & 0.686 & 0.722 \\
        
    \end{tabular}
    \caption{Domain similarities averaged within and across texture types, bounded to interval $[0,1]$. Histogram WD, MAE and LPIPS were inverted to be interpreted as the degree of similarity.}
    \label{tab:image_similarities}
\end{table}

\textbf{Task similarity} is affected by concept shift and estimates how well the task function, mapping inputs to labels, obtained on synthetic data transfers to the target dataset.
In our case we measure how well an image semantic segmentation model trained on synthetic data generalizes to the real data.
In the field of domain adaptation \cite{Farahani2020domainadasurveybrief}, a common approach is to report the difference in task metrics. 
As discussed in \cite{Torralba2011dataset_bias}, this approach measures the difference in biases present in the two datasets, which can be interpreted as domain shifts when working with datasets acquired from different environments.

We train a UNet \cite{Ronneberger2015unet} with a ResNet-50 \cite{He2016resnet} backbone from $5$ random parameter initializations per experiment and report the top obtained result, to reduce the bias of weight initialization. The model is trained to perform semantic segmentation through pixel classification into the classes background, dent, and scratch.
Since the non-background classes are highly underrepresented, we observed the need to use class weighting which we empirically found to be $(1,1.5,1.5)$ for the real and $(1,2,2)$ for the synthetic dataset.
We optimize the models using the AdamW optimizer \cite{Loshchilov2017adamw} with learning rate $10^{-4}$ and $L_2$ weight regularization of $10^{-5}$, for a maximum of $1000$ epochs with early stopping at validation loss convergence upon $5$ consecutive validations without at least $10^{-2}$ relative improvement.
For model fine-tuning we use the learning rate $10^{-5}$.
For memory efficiency, during training we extract random image crops of size $256\times256$ and during evaluation we split the images into a $3\times3$ grid of patches of size $416\times352$.
All images were zero padded to ensure dimension divisibility correctness when evaluating the UNet and image patching.
To speed up model training, we center the images linearly between $[-1,1]$.

Since rendering images is a memory and compute expensive operation we offload as much of the domain alignment as possible to online post-processing.
To simulate the non-zero background, we artificially add Gaussian noise of amplitudes in the interval $[0,5]$ to the background around the object.
To simulate the vertical blooming effect caused by photon leakage within the saturated regions of the CCD sensor array, we extract the maximum value of each image column, mask the column using a threshold $0.95$, blur it using a box filter of size $64$, multiply by $0.02$ and add it to the original image with clipping.
Manually annotated real images contain imprecision in masks as they often cover the area around the rim of the defect.
We emulate this by dilating the synthetic masks by $1$ pixel.
The synthetic images contain masks in regions that have mostly uniform value and should be considered invisible.
We pre-filter the masks to remove defects with insufficient visibility.
We empirically estimate the visibility as the difference between the mean intensity of the defect and the mean intensity of the surface ring around the defect, calculated as difference between the defect mask which is dilated $2$ times and the original defect mask.
Defects are considered visible if the difference in means is over $0.05$.

Data augmentation further increases the diversity of the dataset and helps regularizing the recognition model to consider the important features.
To simulate the offsets of the camera we use random rotation of amplitude $15^{\circ}$.
Illumination strength and object material diffusion variation can be jointly controlled using random exposure from the interval $[-0.5,1.5]$.
To simulate the blurred background structures and impurities in the lens or the light carrying medium, we add Gaussian noise of amplitude in the range $[0,5]$ to the entire image and then blur it using defocus blur \cite{Hendrycks2019defocusblur} of kernel sizes in the interval $[1,3]$.
Gaussian noise of amplitude in $[0,5]$ is added once more to simulate the static noise of the real images.
Finally, images can be randomly flipped horizontally to increase the texture-structure pair diversity.
All of the augmentations are performed with probability of $50\%$.
Finally, we perform intensity biased random cropping \cite{fulir2023syntheticclutch} using threshold $15$ to ensure production of crops that contain more visible structures in the image.
 
The results of image segmentation are collected in \cref{tab:prediction_similarities}.
When computing the mean score, the contribution of the background class to the mean value was ignored as it is the majority class that would skew the results away from the defect classes.
To evaluate the concept shift, we compare common semantic segmentation metrics of the model generalization when trained on the real subset to the synthetic subset.

\begin{table}[t]
    \centering
    \resizebox{\linewidth}{!}{
    \begin{tabular}{lrcccc}
        
        Texture & Domains & mP & mR & mF1 & mIoU \\
        \hline
        \multirow{4}{*}{Sandblasted}
        & Sy $\rightarrow$ Sy         & 57.0 & 53.1 & 54.7 & 37.7 \\ 
        & Sy $\rightarrow$ Re         & 34.0 & 64.3 & 41.7 & 26.3 \\
        & Sy+ft $\rightarrow$ Re      & 60.1 & 71.9 & 65.5 & 49.6 \\
        & Re $\rightarrow$ Re         & 57.9 & 61.9 & 59.1 & 42.6 \\
        \hline
        \multirow{4}{*}{Parallel}
        & Sy $\rightarrow$ Sy         & 57.1 & 39.6 & 45.6 & 29.6 \\ 
        & Sy $\rightarrow$ Re         & 26.5 & 23.0 & 23.5 & 13.8 \\
        & Sy+ft $\rightarrow$ Re      & 52.6 & 33.4 & 38.3 & 23.9 \\
        & Re $\rightarrow$ Re         & 50.3 & 34.5 & 40.8 & 25.8 \\
        \hline
        \multirow{4}{*}{Spiral}
        & Sy $\rightarrow$ Sy         & 59.5 & 39.7 & 47.6 & 31.3 \\ 
        & Sy $\rightarrow$ Re         & 35.2 & 22.1 & 26.4 & 15.5 \\
        & Sy+ft $\rightarrow$ Re      & 49.5 & 40.1 & 43.9 & 28.4 \\
        & Re $\rightarrow$ Re         & 50.6 & 43.1 & 46.2 & 30.7 \\
        \hline
        \multirow{4}{*}{All}
        & Sy $\rightarrow$ Sy         & 59.4 & 46.0 & 51.5 & 34.7 \\ 
        & Sy $\rightarrow$ Re         & 32.2 & 31.3 & 31.6 & 18.9 \\
        & Sy+ft $\rightarrow$ Re      & 59.0 & 49.1 & 53.1 & 36.1 \\
        & Re $\rightarrow$ Re         & 59.0 & 48.2 & 52.6 & 35.8 \\
    \end{tabular}}
    \caption{Task similarities between texture groups. The domains represent the experiment \textit{training $\rightarrow$ testing} domains, with the real (Re) and synthetic (Sy) domains. Fine-tuning (ft) is performed using real data after training on synthetic data.}
    \label{tab:prediction_similarities}
\end{table}

\section{Discussion}
\subsection{Pipeline controlability and simplicity}
The pipeline is very versatile and allows generation of a wide variety of texture variations needed for domain randomization.
Parametrization of the modules allows defining elements using physical measurements.
The modularity of the generation process allows fast iterations during initial dataset development phases and extensibility to new textures and defects.
The rendering is the most computationally demanding component, but it can be easily parallelized.

The large number of parameters make it difficult to study the influence of different parameter sampling schemes on the recognition performance.
However, this pipeline opens the opportunity to design a higher level of abstraction for dataset design through modeling of lower dimensional constraints over larger sets of parameters.
While this study is out of scope of this paper, we strongly encourage it to be performed in the future.

\subsection{Domain similarity}
\label{sec:discuss_domain_similarity}
Measuring texture similarity is a difficult problem due to surface and acquisition imperfections present in the real data.
Camera blurring, surface imperfections and other effects move away from the nominal texture without degrading its performance, thus producing falsely bad results.
In addition, it is difficult to generate a nominal image that perfectly aligns to the acquired image due camera-object and texture-object alignments.
In this work we instead rely on methods that compare global or local pixel statistics of textures to remove alignment.
From results in \cref{tab:image_similarities} we observe that the sandblasted textures have the highest degree of similarity compared to the parallel and spiral milling textures.
This result is related to the simplicity of modeling the sandblasted texture as the texture appears as mostly uniform noise.
More structured textures such as those of milling are more difficult to replicate as they have significantly more parameters that interplay to produce unique patterns.

Comparing the values of metrics is difficult, even though they are bounded between $[0,1]$ and might measure similar properties of images, they do not share the same linearity of response to changes in inputs.
A separate study is required to compare which physical properties different metrics attempt measuring over additional datasets.
Since such a study targets a different task, it requires different definitions of dataset parameters which is out of scope of this study.

\subsection{Task similarity and recognition performance}
The defect recognition results in \cref{tab:prediction_similarities} show very similar performance in all experiments, except when the model trained on the synthetic domain is evaluated directly on the real domain.
The similarity of results, even after fine-tuning, indicates that the models are saturated in their capacity to solve the task.
The low generalization to the real domain indicates that there exists a misalignment between the synthetic and real data distributions, both in the form of covariate shift (as observed by similarity metrics) and concept shift.
Results of models trained on all textures do not surpass the performance on the highest performing, sandblasted texture, leading us to the conclusion that training a model on multiple textures at once does not give any benefit over training a separate model for each texture.

When inspecting the model predictions over the datasets we notice some patterns.
All models tend to predict masks that are slightly wider than the label.
Most false positives come from the milled textures, in the lines where two milling path circles overlap and along the exit lines.
In sandblasted textures, tiny glints from the surface are often falsely predicted as defected.
False positives are also present in minor deviations in object shape such as edge scuffing, beveling, deeper sandblasting holes, etched signature, on paper label sticker and where the milled pattern has high contrast.
Models also tend to predict defects on out-of-focus faces that the annotator left as unannotated due to heavy blurring or unrecognazibility.
When predicting scratches, the models prefer areas where the texture has lower contrast and defect stand out.
Some scratches that are not close to being horizontal or vertical are not predicted, even though the visibility is similar to the predicted ones, meaning the recognition models need more filters in lower levels or more explicit rotation invariance.
Overall, it seems a higher resolution might help the model distinguish between defect and texture patterns.

\subsection{Influence of domain similarity on task performance}
The largest drop in recognition performance is on milled textures, which is consistent with their lower domain similarity compared to sandblasted textures.
Furthermore, SSIM and LPIPS seem to be linearly correlated to the best obtained recognition performance.
If this were true, we could use these metrics when iteratively improving the synthetic data to reduce the domain gap at its source, the data simulation or generation process.
This forms the questions of weather these metrics are indeed predictive of the domain gap that impedes the recognition performance, which other metrics have a similar property and why MAE and histogram comparison do not seem to be.
These questions are vastly out of scope of this paper and we leave their treatment for a future study.

\section{Conclusion}
This work presents a rule-based solution for generating synthetic data of various surface finishes for metal objects.
The solution consists of surface texture models controlled by parameters measured in examples, application of specification based defects and rendering to produce a synthetic images and pixel-wise annotations.
An approach for synthetic image and distribution quality estimation is proposed.
Utility of generated data is measured on the defect segmentation task.
Results of quality estimation shows correlation to results of the segmentation task.
The utility of the synthetic data requires further research due to the need for modeling of significant imperfections in the image acquisition system and object geometry.

\section*{Acknowledgments}
This work was funded by the German Federal Ministry of Education and Research (BMBF) [grant number 01IS21058 (SynosIs)].

\bibliographystyle{ieeetr}
\bibliography{references}

\end{document}